\documentclass[runningheads]{llncs}
\usepackage{graphicx}
\usepackage{comment}
\usepackage{amsmath,amssymb} 
\usepackage{color}
\usepackage{boxedminipage}
\usepackage{tikz}
\usepackage{pgfplots}
\usepackage{float}
\usepackage{rotating}
\usepackage{booktabs}
\usepackage{subfigure}
\usepackage{multirow}

\begin{document}
\pagestyle{headings}
\mainmatter
\def\ECCVSubNumber{6068}  

\title{Quantized Neural Networks: Characterization and Holistic Optimization} 

\titlerunning{Quantized Neural Networks: Characterization and Holistic Optimization}
%
\author{Yoonho Boo \and Sungho Shin \and Wonyong Sung}

%
\authorrunning{Y. Boo et al.}
%
\institute{
Department of Electrical and Computer Engineering \\
Seoul National University \\
\email{dnsgh337@snu.ac.kr; sungho.develop@gmail.com; wysung@snu.ac.kr}
}
\maketitle

\begin{abstract}
Quantized deep neural networks (QDNNs) are necessary for low-power, high throughput, and embedded applications. Previous studies mostly focused on developing optimization methods for the quantization of given models. However, quantization sensitivity depends on the model architecture. Therefore, the model selection needs to be a part of the QDNN design process. Also, the characteristics of weight and activation quantization are quite different. This study proposes a holistic approach for the optimization of QDNNs, which contains QDNN training methods as well as quantization-friendly architecture design. Synthesized data is used to visualize the effects of weight and activation quantization. The results indicate that deeper models are more prone to activation quantization, while wider models improve the resiliency to both weight and activation quantization. This study can provide insight into better optimization of QDNNs.


\end{abstract}

\section{Introduction}
\label{sec:intro}

Deep neural network (DNN) applications frequently demand extremely large models for an improved performance, which consumes a large amount of computation power not only for training but also for inference ~\cite{zoph2018learning,devlin2018bert}. Thus, it is necessary to reduce their complexity for implementation on embedded devices. Various DNN compression methods have recently been devised to reduce the computational cost, power consumption, and storage space. Network quantization is one well-known method for substituting 32-bit floating-point weights with low bit-width numbers that usually employ one to four bits. Specifically, the performance of a quantized DNN (QDNN) is mostly maintained when retraining is applied after weight quantization~\cite{hwang2014fixed,courbariaux2015binaryconnect}. Meanwhile, activation quantization has also been studied to reduce the computational cost and working memory footprint~\cite{rastegari2016xnor,zhou2016dorefa}. Therefore, activation quantization is particularly effective for DNN models with a large hidden-state dimension, such as convolutional neural networks (CNNs). Most previous studies on QDNN optimization have focused on the quantization number formats and training methods. The goal of these previous studies has been reducing the performance gap between the floating-point and quantized models. However, not all networks can be quantized in the same manner. Some networks are more robust to weight quantization, whereas some others are not~\cite{sung2015resiliency}. Optimizing a QDNN requires understanding the characteristics of such quantization errors.

In this paper, we visualize the characteristics of the quantization errors and their effects on the performance of QDNNs when the model architecture and sizes are different. We use synthetic data and DNN models for error characteristic visualization. Based on the analysis results, we adopt two simple training methods to compensate weight and activation quantization errors; fine-tuning with cyclic learning rate scheduling for improved generalization and applying regularization term that reduces noise amplification through propagation. Experiments are conducted using CIFAR-10, ImageNet, and PASCAL VOC 2012 semantic image segmentation benchmark. The contributions of this paper are as follows: 


\begin{itemize}
    \item We visualize the errors from the weight and activation quantization. The results indicate that the effect of the weight quantization error reduces the generalization capability whereas activation quantization error induces noise.
    \item We show that increasing the width of a DNN model helps to mitigate the quantization effects of both the weight and activation whereas increasing the depth only decreases the weight quantization error.  
    \item We reduce the weight and activation quantization errors by employing training methods that improve the generalization capability and a regularization term that increases the noise robustness, respectively.
    \item Our work is a holistic approach for the optimization of QDNN by examining the quantization effects of weights and activations, and also the architectural change.
\end{itemize}

\section{Backgrounds}
\label{sec:related_works}


\subsection{Related Works on Network Quantization}

Most DNN models are trained using 32-bit floating-point numbers. Apparently, DNN models do not demand 32-bit precision. Many quantization methods have been developed, some of which use an extremely small bit-width for a weight representation, such as 1-bit binary~\cite{courbariaux2015binaryconnect,rastegari2016xnor} or 2-bit ternary~\cite{fengfu2016ternary,zhu2016trained}. The signal-to-quantization-noise ratio (SQNR) of several weight quantizers was also compared \cite{lin2016fixed}. Quantization noise has been measured to find a better training scheme \cite{hou2018loss} or optimal quantization precision \cite{sakr2018minimum}. Activation quantization has also been developed to lower the computational costs~\cite{choi2018pact}. An efficient QDNN implementation on embedded systems has also been studied~\cite{gupta2015deep,andri2016yodann}. The weight quantization effects usually depend on the model size; small DNN models tend to show considerable performance degradation after quantization~\cite{sung2015resiliency}. In particular, increasing the number of parameters in CNNs reduces the quantization sensitivity \cite{mishra2017wrpn}. However, considering the purpose of model compression, the number of parameters needs to be constrained. A recent study showed that weight quantization up to certain bits does not reduce the memorization capacity \cite{boo2019memorization}. During the last several years, residual connections have been developed mainly for improved training of neural networks~\cite{he2016deep}. Architectural modifications of increasing the width or moving the location of activation and batch normalization have been studied~\cite{zagoruyko2016wide,he2016identity}. These architectural changes also affect the quantization sensitivity. 

The activation quantization has not been discussed as much as weight quantization, and most studies have not distinguished the effects of activation and weight quantization~\cite{rastegari2016xnor,jung2019learning}. It has been observed that activation usually demands more bits than weights~\cite{zhou2016dorefa}. The different quantization approaches for the weight and activation are applied in \cite{park2017weighted} because the latter was not suitable for cluster-based quantization. Many studies have shown that a DNN can be vulnerable to noise. Even with an extremely small amount of noise, the inference of a DNN can easily be manipulated \cite{szegedy2013intriguing,kurakin2018adversarial}. Previous studies have shown that quantizing the input makes it robust to adversarial attacks by reducing the amount of noise \cite{xu2017feature}. Several studies have shown that a QDNN can help defend from adversarial attacks \cite{rakin2018defend,galloway2017attacking}. However, QDNNs become more vulnerable to adversarial attacks than floating-point models when the noise exceeds a certain level \cite{lin2019defensive}. The adversarial noise becomes larger at the deeper layers \cite{liao2018defense}.

\subsection{Revisit of QDNN Optimization}

The process of uniform quantization for a DNN involves the following two steps, namely, clipping and quantization:
\begin{align}
\hat{x} = \textit{Clip }(x, \alpha, \beta), \quad Q(x) = \Delta \lfloor \frac{\hat{x}}{\Delta} + 0.5 \rfloor. \label{eq:quant}
\end{align}
The parameters of the DNNs are signed values so that the clip value $\beta = -\alpha = \Delta (2^{n-1}-1)$ where $n$ is the number of bits used to represent each parameter. Parameter quantization is mainly applied to the weights. For the sake of simple structure, all fixed-point weights in a layer share one scale factor $\Delta$. The activation quantization is used to lower the computational cost and the size of the working memory for inference. When using the ReLU activation, the hidden vectors are represented with unsigned values and $ \alpha $ and $ \beta $ becomes 0 and $ \Delta ( 2^{n} -1 ) $, respectively. Low-precision quantized networks require training in a fixed-point domain to improve the performance as follows:
\begin{gather}
W_{t}^q = Q(W_t) \\
E_t = f(x_t, y_t, W_{t}^q) \\
W_{t+1} = W_t - \alpha \frac{\partial E_t}{\partial W_{t}^q}, \label{eq:quant_retrain}
\end{gather}
where $E_t$ is the loss computed through the model, $f(\cdot)$, at $t$-th iteration. Forward and backward propagations are conducted using quantized weights and activation. However, the computed gradients need to be added to the floating-point weights because those gradients are relatively small compared to the step size $\Delta$ \cite{courbariaux2015binaryconnect}. It is known that QDNNs perform better when retrained from a pretrained model at floating-point than trained from the scratch \cite{zhou2016dorefa,jung2019learning}.


Most QDNN studies quantize the weights or activation according to Equation~\eqref{eq:quant}, although the processes of obtaining $ \Delta $, $ \alpha $, and $ \beta $ are different \cite{hwang2014fixed,choi2018pact}. However, the effect of each quantization on the inference is quite different. The quantization errors are $\epsilon_W = W$ $-$ $Q(W)$ and $\epsilon_a = a$ $-$ $Q(a)$ where  $\epsilon_W$ and $\epsilon_a$ are errors owing to the quantization of the weight and activation, respectively. Because the trained weights have fixed values during inferences, $\epsilon_W$ is a constant error. In other words, weight quantization can be modeled as the process of distorting the weights of the DNNs. This changes the direction of the input-prediction mapping function of the DNNs and thus causes distorted results at the inference. By contrast, $\epsilon_a$ is an error that depends on the input applied during the inference process. Depending on the remainder of the hidden vector divided by $\Delta$, the direction or magnitude of the error may change. That is, $\epsilon_a$ induces noise with a maximum magnitude of $ \frac{\Delta}{2}$. 

In the rest of this paper, we analyze how such differences in quantization errors affect the performance of QDNNs under various model architectures. The weight quantization method in \cite{hwang2014fixed} and the PACT activation quantization \cite{choi2018pact} are adopted for our experiments. QDNNs are retrained from pretrained floating-point models.

\section{Visualization of Quantization Errors using Synthetic Dataset}
\label{sec:analysis}

\subsection{Synthetic Dataset Generation}

Most DNNs and their training samples used in real tasks have extremely high dimensions, and it is therefore very difficult to discern the effects of quantization. For a visualization analysis of QDNNs, we synthesized 2D inputs whose elements consist of $x$ and $y$ axes. The training dataset is composed of inputs $S \in \mathbb{R}^2 $ and $C \in (0,1) $, which are used to train FCDNNs for binary classification. The training dataset is synthesized through two steps. First, the core samples, $s_{c}$, mapped to a label, $c$, are generated using the following equation:
\begin{align}
s_{0} \in &\bigg\{(x,y) | \begin{cases}x^2 + y^2 = (2i+1)^2r^2, &y > 0\\ (x-r)^2 + y^2 = (2i+2)^2r^2, &y \leq 0\end{cases} \bigg\}, \\
s_{1} \in &\bigg\{(x,y) | \begin{cases}x^2 + y^2 = (2i+2)^2r^2, &y > 0\\ (x-r)^2 + y^2 = (2i+1)^2r^2, &y \leq 0\end{cases} \bigg\}.
\end{align}
In our experiments, $i$ is within $\{0,1,2,3,4\}$ and $r$ is $0.1$. For each $i$, two semicircles correspond to one label. In each semicircle, we sample 100 points by increasing the angle linearly. Therefore, the total number of core samples is 2,000. Next, we generate subsamples by adding Gaussian noise to each core sample as follows:

\begin{align}
s' = s + \epsilon, \quad \epsilon \sim \mathcal{N}(0, \frac{1}{3}r\mathbf{I}).
\end{align}

\begin{figure}[t]
    \centering
    \subfigure[Dataset]{\includegraphics[width=0.22\linewidth]{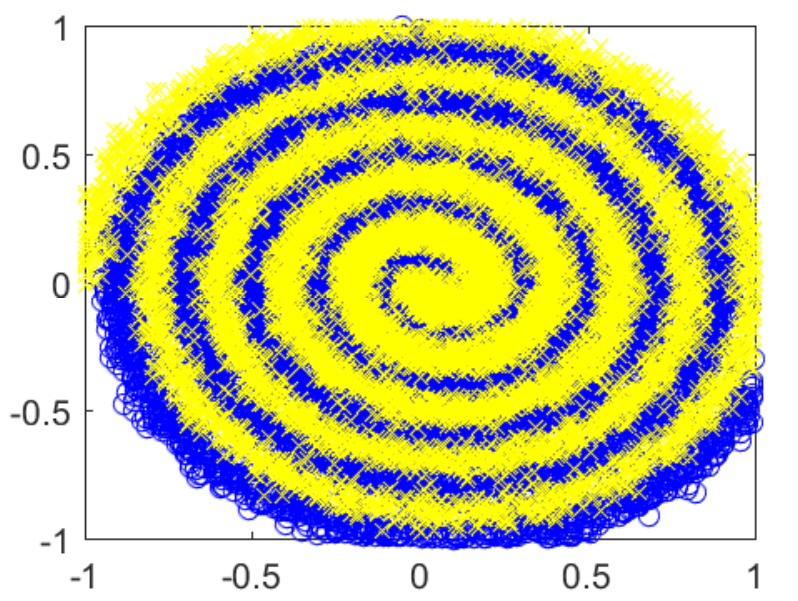}}
    \subfigure[128-3 (69.8\%)]{\includegraphics[width=0.21\linewidth]{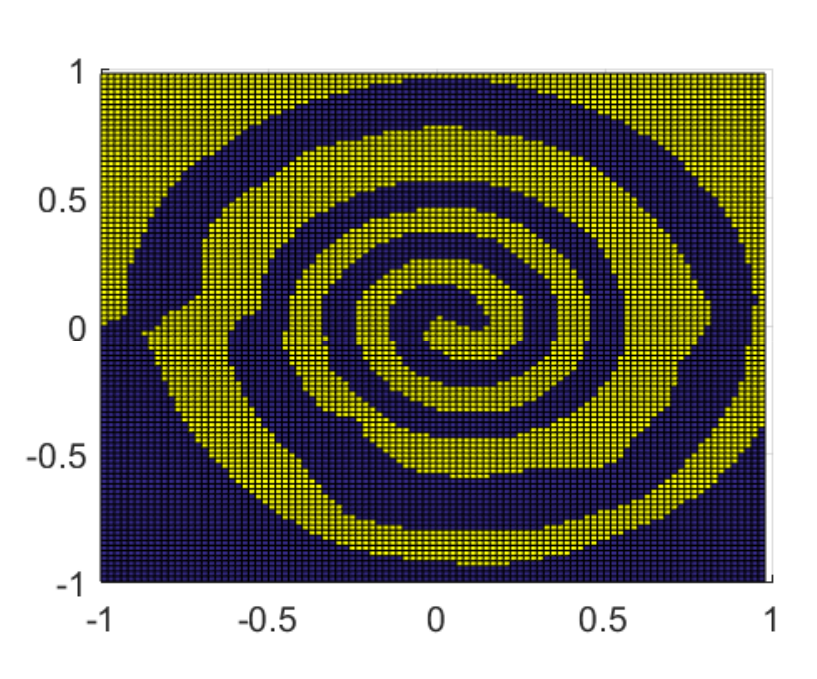}}
    \subfigure[128-4 (99.1\%)]{\includegraphics[width=0.22\linewidth]{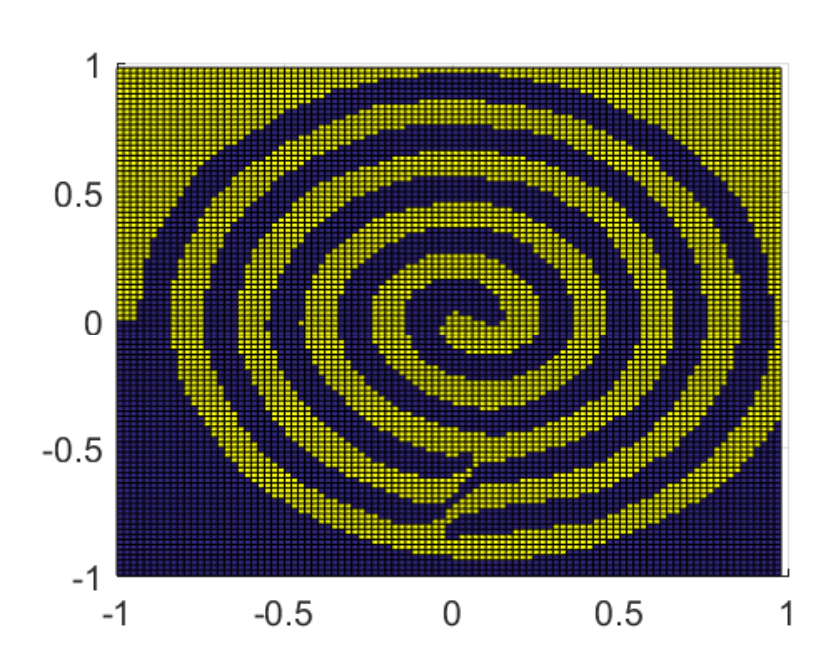}}
    \subfigure[256-3 (99.2\%)]{\includegraphics[width=0.21\linewidth]{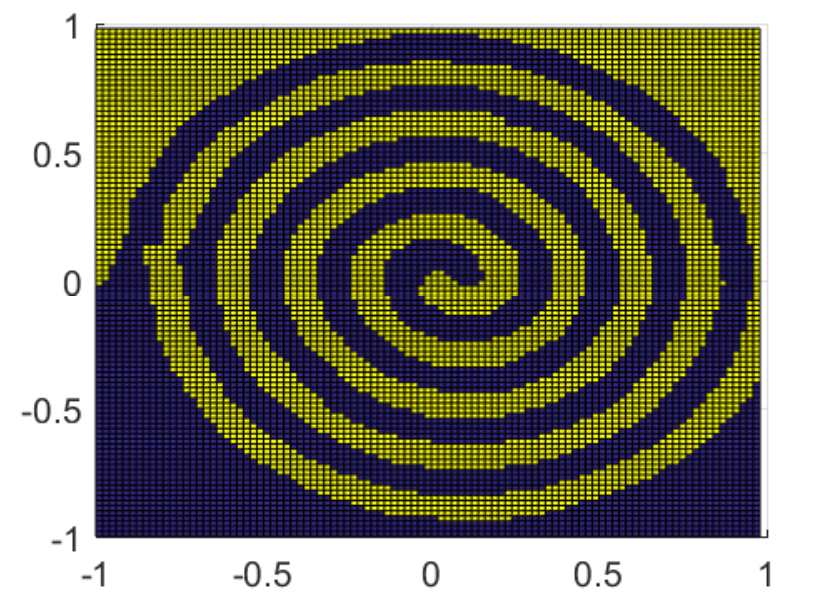}}
    \caption{Illustrations of the dataset and prediction results from FCDNNs. (a) The synthetic dataset used to train FCDNNs. (b) An example of the evaluation results when the model is too small to learn to the data distribution. (c, d) Examples of large models. The values in parentheses are the accuracy for the correct answer.}
    \label{fig:toy_example}
\end{figure}
\noindent Nine subsamples for each core sample are used, and the samples are mapped to the same labels. As a result, the total number of datasets used for training is 10,000 for each label. The distribution of the generated dataset is shown in~\figurename~\ref{fig:toy_example}~(a).

We quantize the FCDNN trained using the generated dataset and analyze the difference between the weight and activation quantization. In particular, we visualize the errors when varying the model depth or width. Two datasets are adopted for the evaluation of the trained model. The first is a test dataset. The correct answer dataset is constructed by dividing the area corresponding to each label by the radius of the semicircle. This dataset is used for quantitative analysis of the trained QDNNs by measuring accuracy. The other is a grid dataset that consists of $(x, y) \in \{ (x, y) | 0 < x <1, 0 <y <1 \}$. By visualizing the prediction on the $x$-$y$ plane, we can analyze whether the input-prediction mapping of a DNN is distorted or as added noise.

\begin{figure}[t]
    \centering
    \subfigure[128-4 (78.0\%)] {\includegraphics[width=0.21\linewidth]{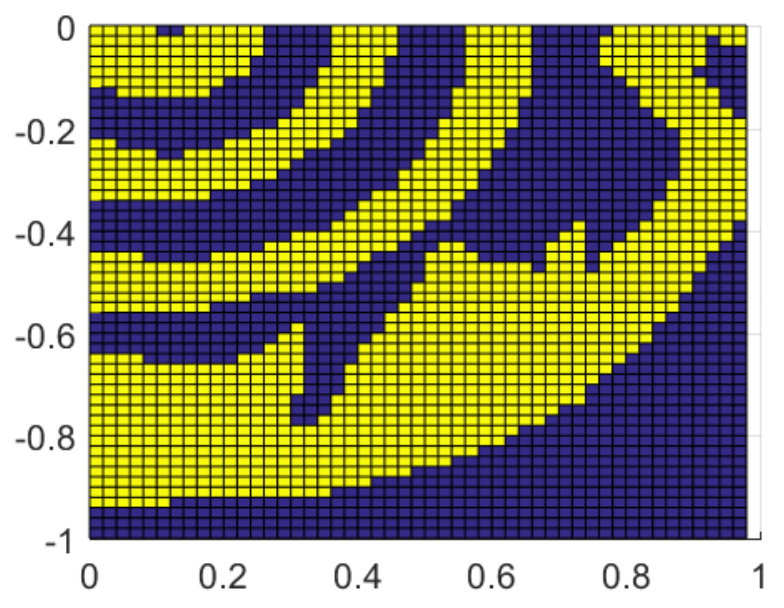}}
    \subfigure[(82.2\%)] {\includegraphics[width=0.21\linewidth]{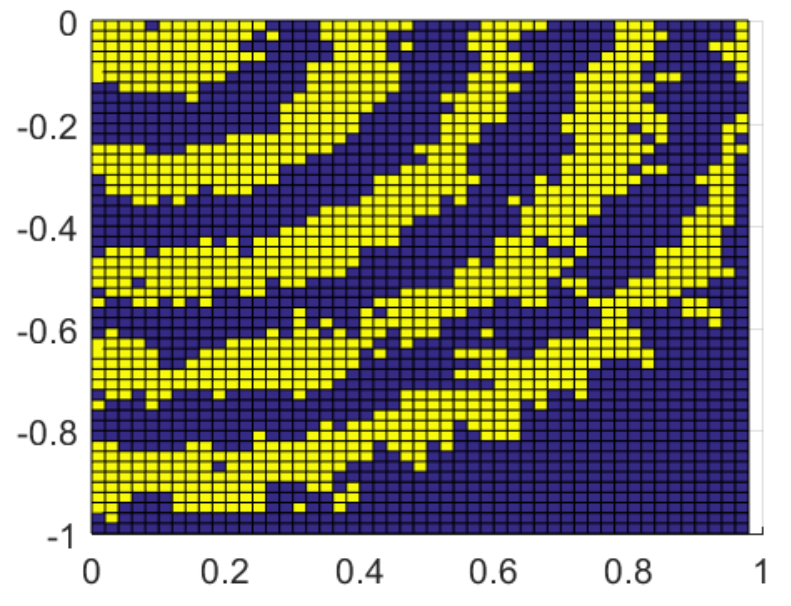}}
    \subfigure[(84.6\%)] {\includegraphics[width=0.21\linewidth]{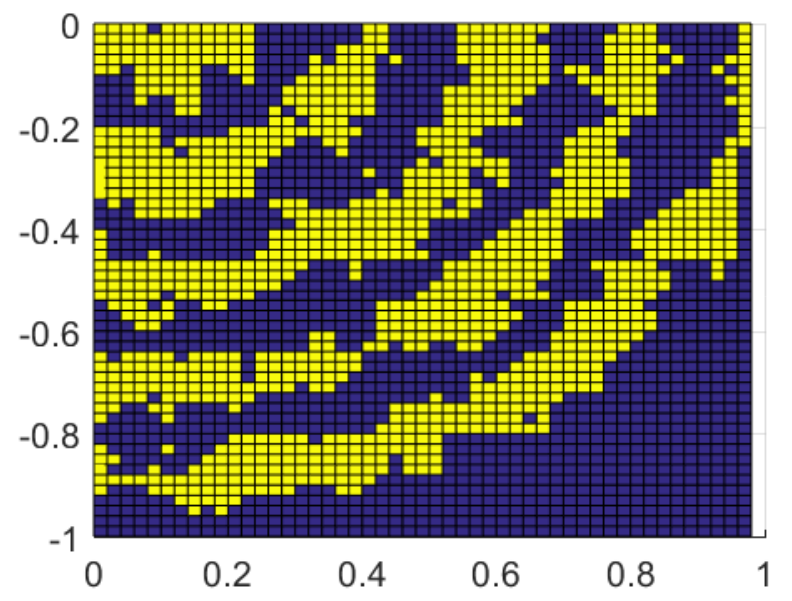}}
    \\
    \subfigure[256-4 (96.2\%)] {\includegraphics[width=0.21\linewidth]{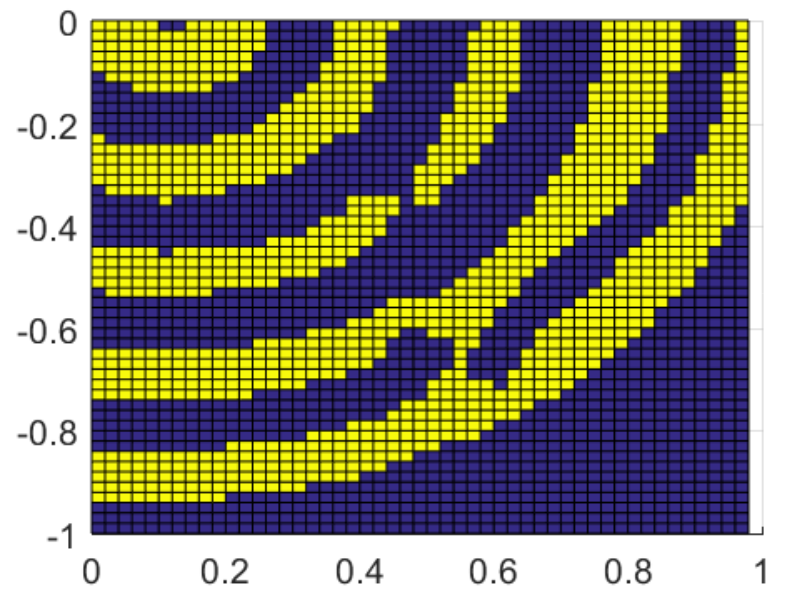}}
    \subfigure[(93.4\%)] {\includegraphics[width=0.21\linewidth]{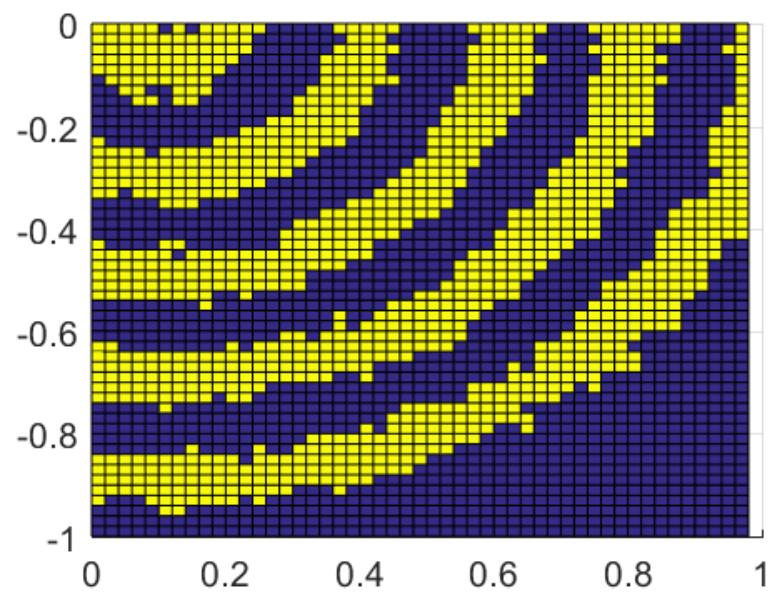}}
    \subfigure[(88.8\%)] {\includegraphics[width=0.21\linewidth]{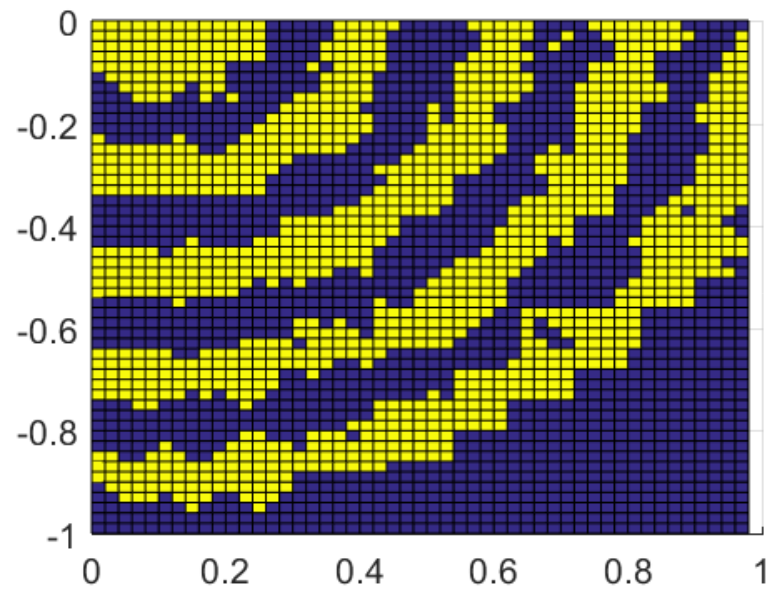}}
    %
    %
    \\
    \subfigure[128-8 (98.8\%)] {\includegraphics[width=0.21\linewidth]{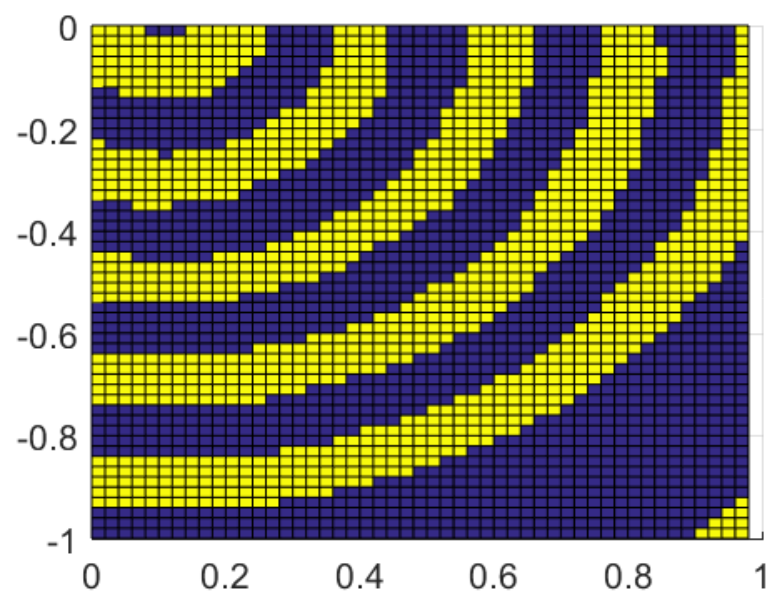}}
    \subfigure[(86.8\%)] {\includegraphics[width=0.21\linewidth]{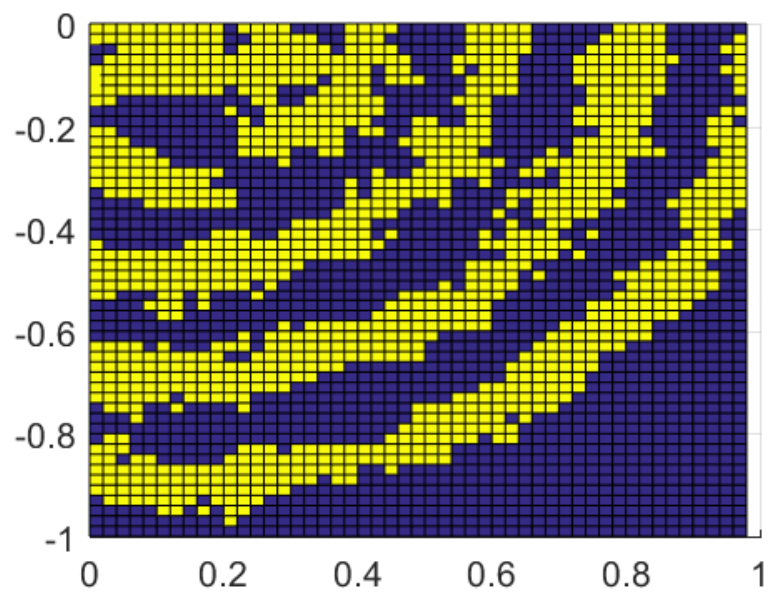}}
    \subfigure[(66.4\%)] {\includegraphics[width=0.21\linewidth]{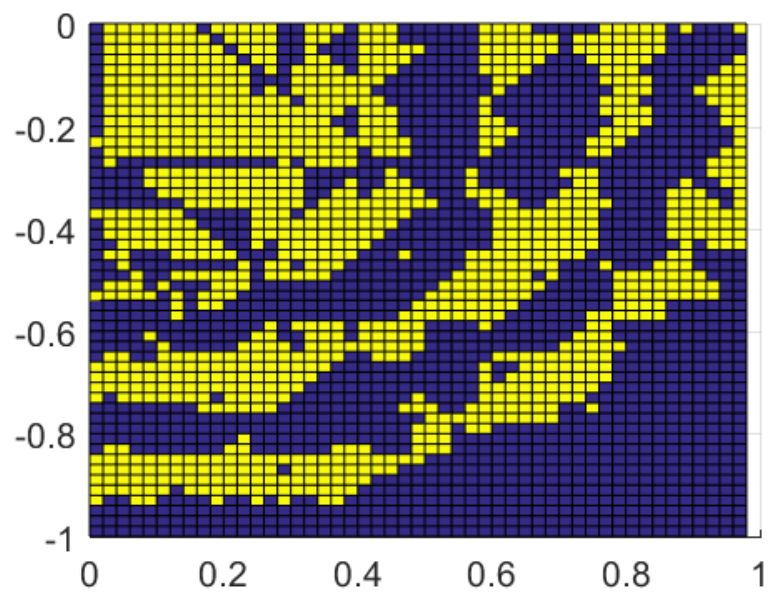}}
    
    \caption{Prediction results of QDNN as the width and depth increase. (\textbf{Left}) 2-bit weights. (\textbf{Middle}) 2-bit activations. (\textbf{Right}) 2-bit weights and activations.}
    \label{fig:toy_example_depth_width}
\end{figure}

\subsection{Results on Synthetic Dataset}
\label{subsec:toy_results}

We devise artificial DNN models for testing with the synthetic dataset. Fully-connected DNN (FCDNN) models are employed with varying depth and width. In addition, models with residual connections are also considered. We indicate the experimental models as ``width'' - ``depth'' of FCDNNs. The prediction results of floating-point models using the evaluation dataset are shown in~\figurename~\ref{fig:toy_example}. The 128-3 FCDNN shows quite a different prediction result from the actual data distribution. \figurename~\ref{fig:toy_example} (c,d) show that increasing the depth to 4 or the width to 256 is sufficient to learn the synthesized dataset quite faithfully. For the remaining experiments, we represent only the bottom-right quarter circle for detailed visualization. All QDNN results are reported after retraining. 

\begin{figure}[t]
    \centering
    \subfigure[128-4 (91.0\%)]
    {\includegraphics[width=0.21\linewidth]{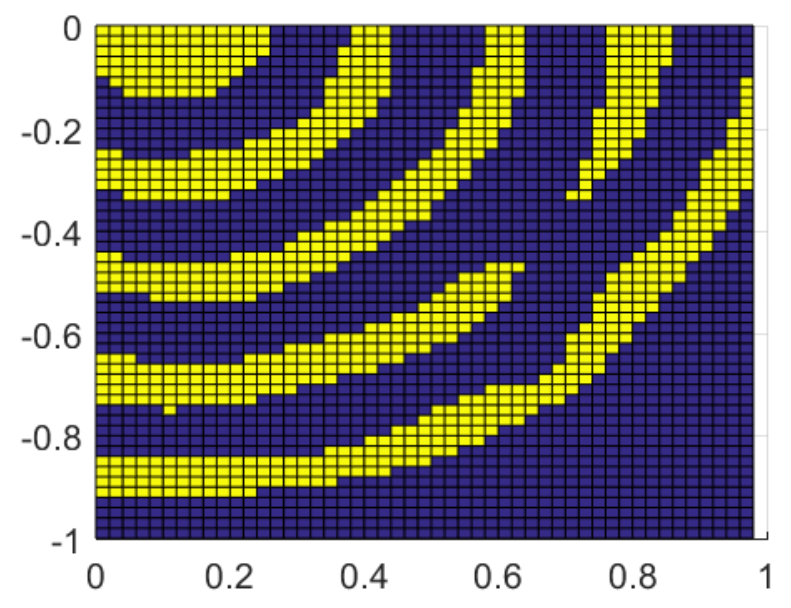}}
    \subfigure[(88.5\%)]
    {\includegraphics[width=0.21\linewidth]{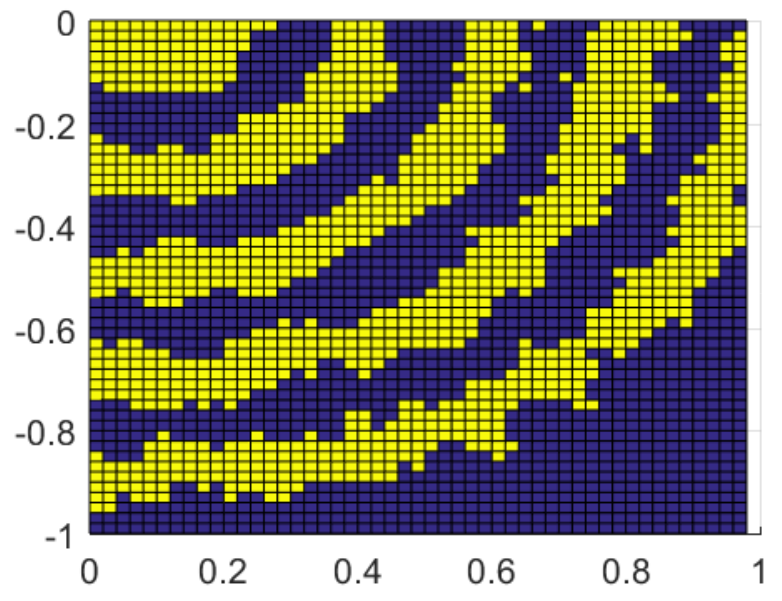}}
    \subfigure[(84.8\%)]
    {\includegraphics[width=0.21\linewidth]{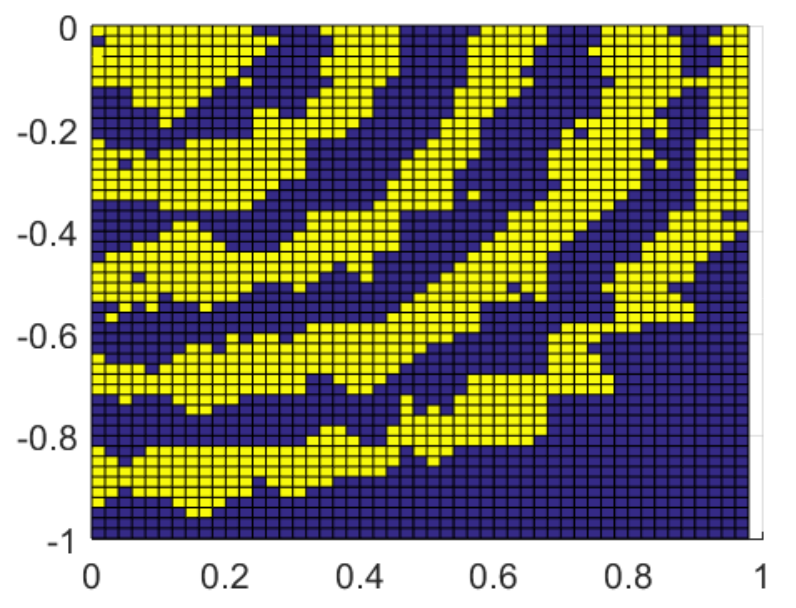}}
    \\
    \subfigure[128-8 (98.0\%)]
    {\includegraphics[width=0.21\linewidth]{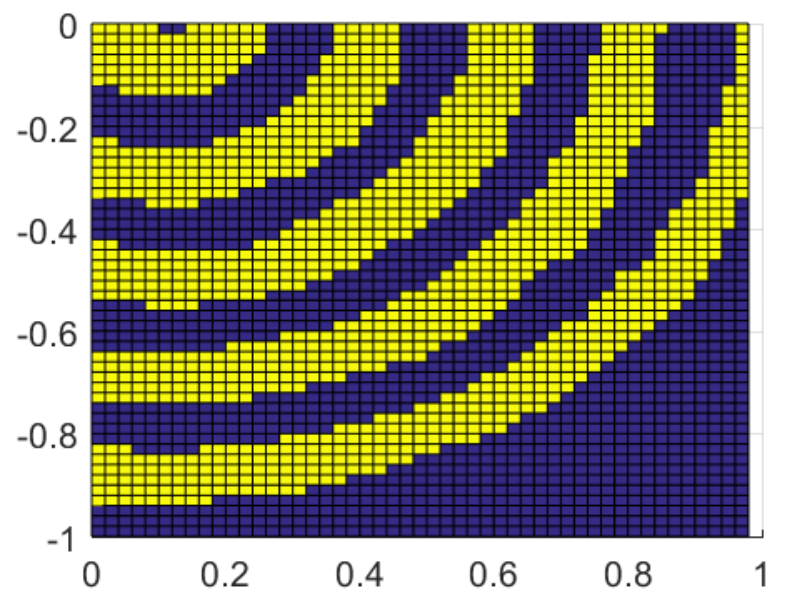}}
    \subfigure[(81.8\%)]
    {\includegraphics[width=0.21\linewidth]{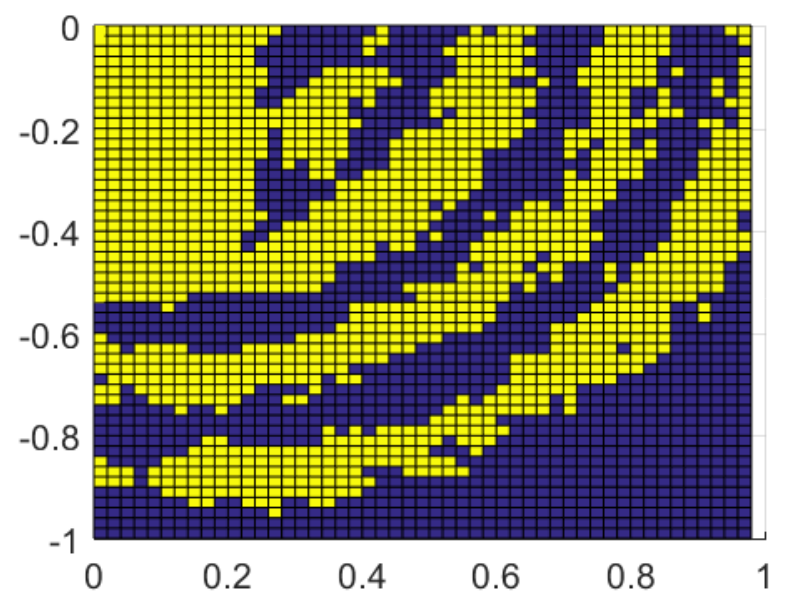}}
    \subfigure[(61.9\%)]
    {\includegraphics[width=0.21\linewidth]{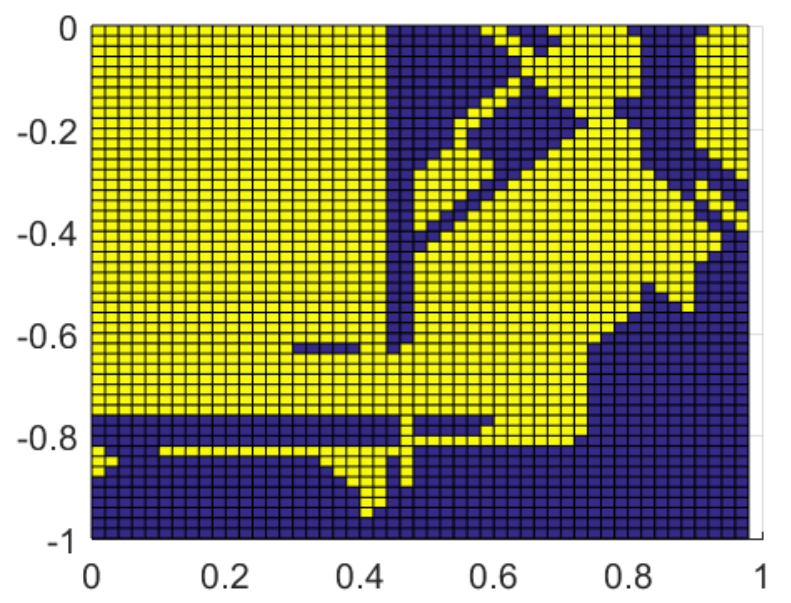}}
    
    
    \caption{Prediction results of QDNNs with residual connection. (\textbf{Left}) 2-bit weights. (\textbf{Middle}) 2-bit activations. (\textbf{Right}) 2-bit weights and activations.}
    \label{fig:toy_example_residual}
\end{figure}

\figurename~\ref{fig:toy_example_depth_width} (\textbf{Left}) shows the effects of the weight quantization according to the width and depth of FCDNNs. We can see that weight quantization distorts the input-prediction mapping of the DNN. The evaluation results of the weight quantized models resemble that of a small floating-point model, such as the 128-3 FCDNN shown in~\figurename~\ref{fig:toy_example} (b). The experiment results show that the decrease in learning ability occurs similarly when the model size is reduced or the weights are quantized. As studied in~\cite{neyshabur2018towards}, increasing the model size helps generalization. 
Our experiments show that the generalization capability decreases as the precision of the parameters is lowered. Thus, the effect of reduced generalization capability due to the weight quantization is not noticeable when the model size is large enough. The distortion with 2-bit weight quantization is barely found when the layer width or the number of layers is increased.

\figurename~\ref{fig:toy_example_depth_width} (\textbf{Middle}) shows the activation quantization results. The effect of the activation quantization is very different from that of the model capacity reduction in a DNN. Activation quantization appears to add noise to the prediction results. Although both the weight and activation quantization errors degrade the performance of DNNs, they behave in a completely different manner. When the activation is quantized to 2 bits, increasing the depth does not mitigate the noise added to the prediction results. Rather, the noise tends to worsen with weight quantization when the depth increases. Activation quantization is related to the dimension of each layer rather than the capacity of the model. Activation quantization in wide FCDNNs (\figurename~\ref{fig:toy_example_depth_width} (e)) is more robust than in deep FCDNNs (\figurename~\ref{fig:toy_example_depth_width} (h)). The effect of noise from the activation quantization is reduced because the number of dimensions of the hidden vector received as the input in each layer is increased. When quantizing both the weight and activation, the two errors are combined, and result in a noisy and distorted prediction, as shown in~\figurename~\ref{fig:toy_example_depth_width} (\textbf{Right}). The more results of different QDNNs are reported in Appendix \textbf{A}.

We also analyze the effect of residual connections on activation quantization. The residual connection helps train DNNs with an extremely large number of layers \cite{he2016deep}. In our experiment, the residual connections are implemented by adding each hidden output to the activation of the next layer. The result is multiplied by 0.5 to preserve the scale of the intermediate results. \figurename~\ref{fig:toy_example_residual} shows the results of the quantizing activation when the residual connection is applied. Residual connections help alleviate the distortion through weight quantization. However, FCDNNs with residual connections are more sensitive to activation quantization than the original models. With residual connections, the outputs of the quantized hidden layer are summed so that the activation quantization noise is also added. As a result, applying residual connections shows more noisy prediction in deep models, such as 128-8 FCDNNs.

\section{QDNN Optimization with Architectural Transformation and Improved Training}
\label{sec:training}

The visualization results with the synthesized data show that weight quantization decreases the generalization capability of DNNs, while activation quantization induces noised inference. Also, the quantization effects depend on the architecture very much. Based on this observation, we employ three approaches for QDNN optimization. The first one is modifying the architecture quantization-friendly. The second one is the training method for improved generalization. This technique is intended to reduce the effects of weight quantization. The third one is applying the regularization term that limits the amount of activation noise.

\begin{figure}[t]
    \centering
    \subfigure[] {\includegraphics[width=0.18\linewidth]{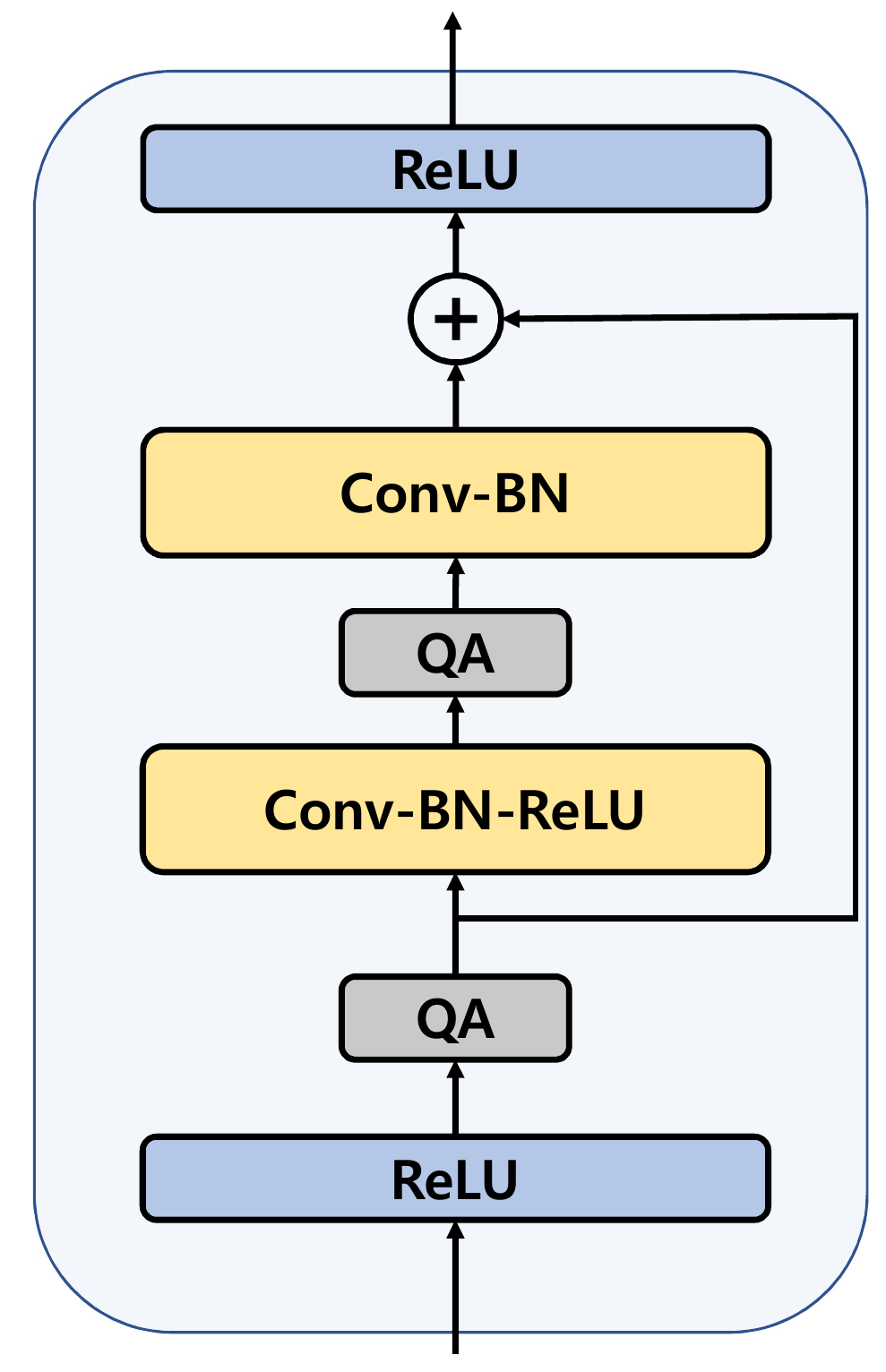}}
    \hskip 0.3in
    \subfigure[] {\includegraphics[width=0.18\linewidth]{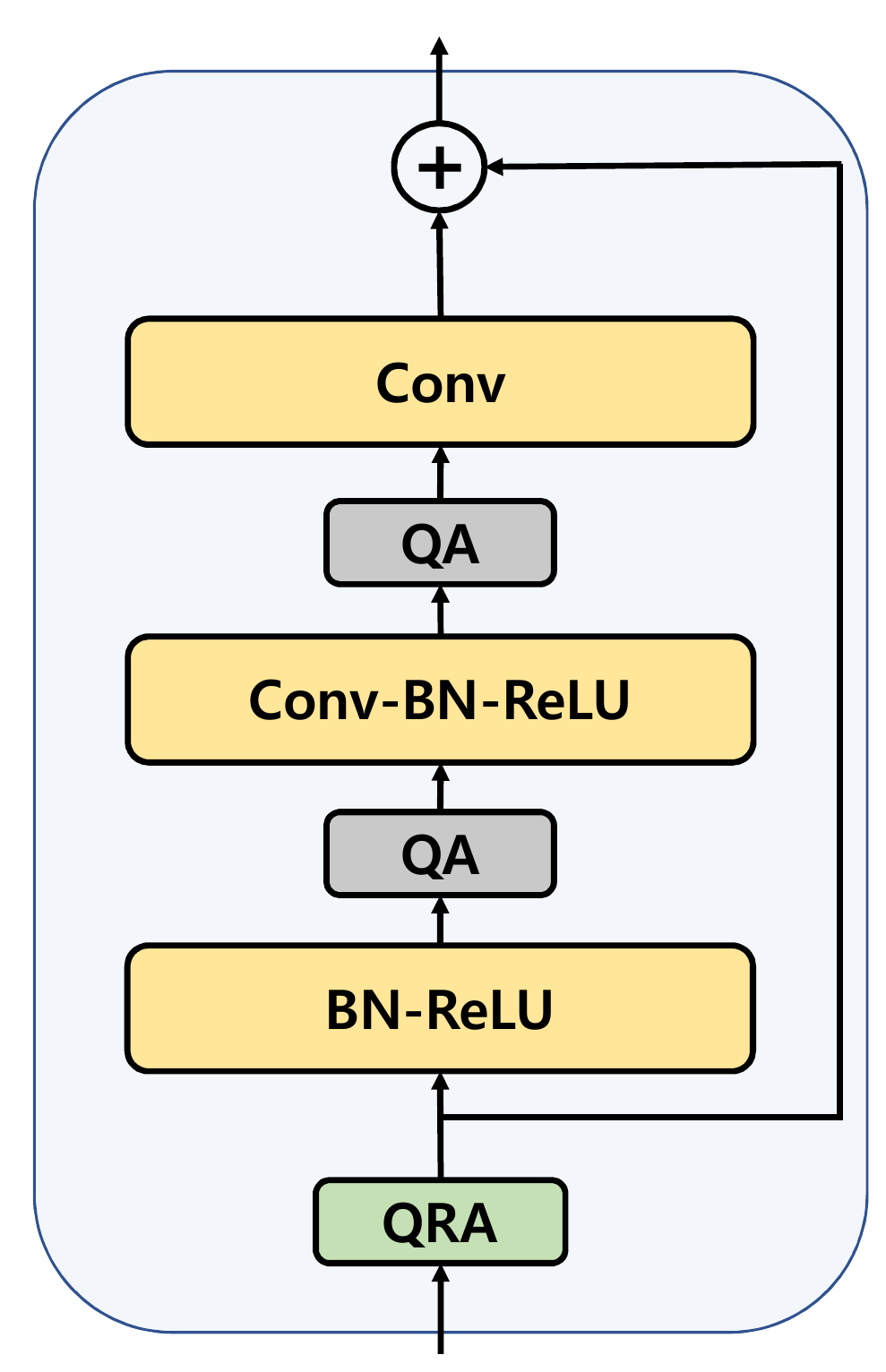}}
    \hskip 0.3in
    \subfigure[] {\includegraphics[width=0.18\linewidth]{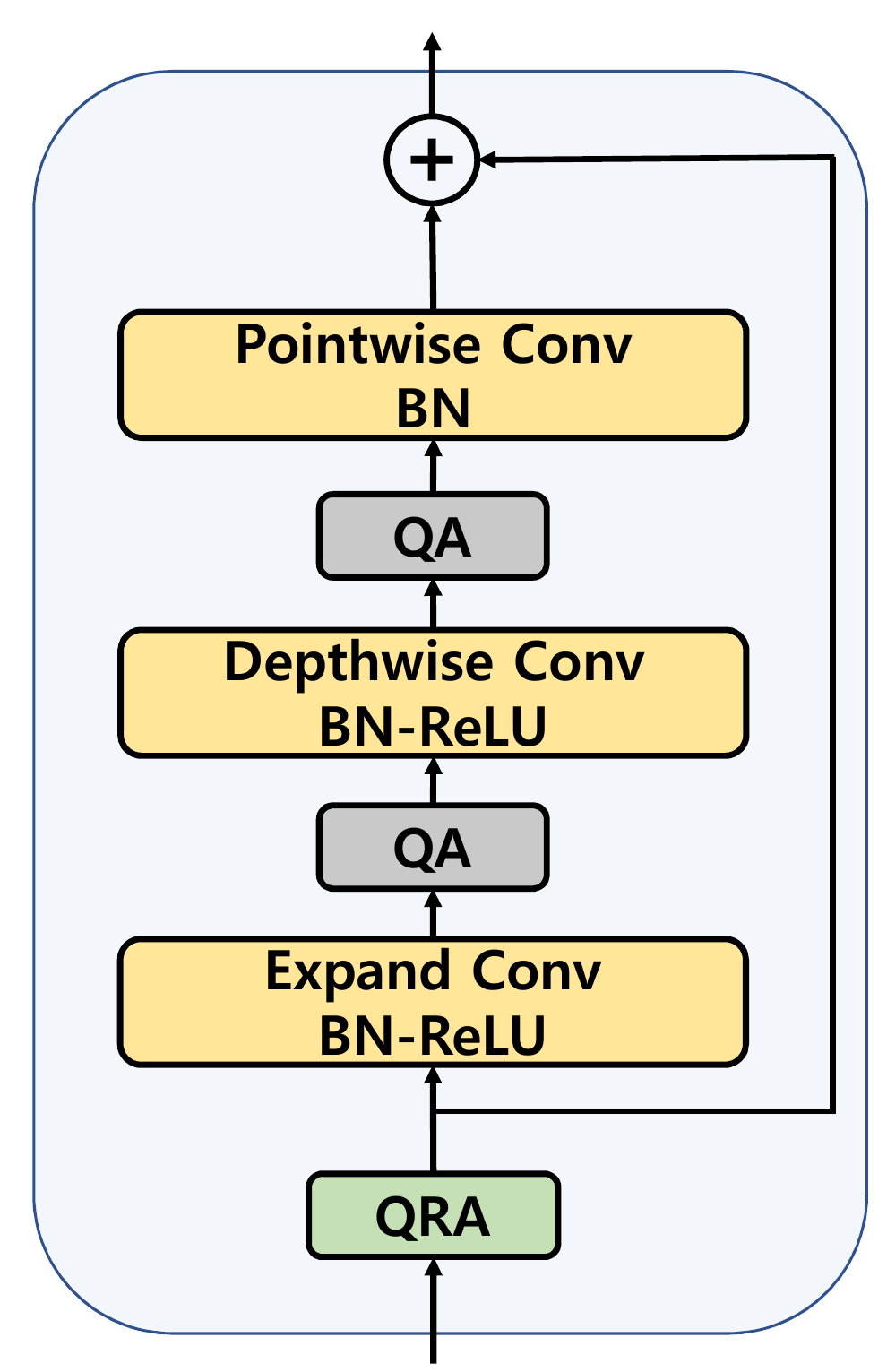}}
    \caption{Types of residual blocks. (a) basic block, (b) pre-activation block, and (c) depthwise block. QA and QRA denote the activation quantization operations.}
    \label{fig:skip_diagram}
\end{figure}

\subsection{Architecture Transformation for Improved Robustness to Quantization}

Deep CNN models are hard to train because of the gradient vanishing problem. The residual architecture was developed to solve this problem~\cite{he2016deep,wu2016google,vaswani2017attention}. In CNN with residual connections, increasing the depth, often over 100, usually helps to improve the performance. Of course, widening the networks also increases the performance~\cite{zagoruyko2016wide}. When the number of parameters is limited, increasing the depth is usually preferred because the model complexity rises in proportional to the depth, but squarely proportional to the width. However, our work in in Section~\ref{sec:analysis} shows that deep CNN models are very prone to activation quantization. 
The conventional approach for QDNN design is developing the best performing floating-point model, and then quantizing it in the best way possible. In this case, the best performing floating-point model prefers deeper ones, which are, however, prone to activation quantization. Thus, we need to consider the effects of weight and activation quantization even for the initial floating-point model design. Wide CNN models, which are considered parameter inefficient, often show better performance than deep ones when the activation is severely quantized.

Recent CNN models employ various residual blocks for improved performance or parameter efficiency. The most well-known residual blocks for CNNs are shown in~\figurename~\ref{fig:skip_diagram}. The depthwise block employed to MobileNetV2~\cite{sandler2018mobilenetv2} helps to reduce the number of parameters and computations. However, the quantization performances of these blocks are not well studied.

\subsection{Cyclical Learning Rate Scheduling for Improved Generalization}

We adopt the cyclic learning rate scheduling (CLR) as a way to reduce the effects of weight quantization. CLR increases and decreases the learning rate periodically, while conventional training usually reduces the learning rate in one direction. This method is known to increase the generalization capability of the model by leading to a flat loss surface \cite{smith2017cyclical}. Among a few different cyclical learning rate scheduling algorithms, we choose the one that alters the learning rate discretely, which is known to be more effective for generalization \cite{jastrzkebski2017three}. The maximum and minimum boundaries of the CLR are determined between the 100 and 0.1 times of the last learning rate of the retraining procedure, respectively. The learning rate changes 8 times in one cycle and exponentially decreases or increases. The CLR scheduling for each task is illustrated in Appendix~\textbf{B}.

\subsection{Regularization for Limiting the Activation Noise Amplification}

Training methods to increase noise robustness of DNNs have been studied in the field of adversarial training. Parseval networks reduce the Lipschitz constant so that the noise of the input is not amplified as the layer increases \cite{cisse2017parseval}. In particular, \cite{lin2019defensive} shows that activation quantized DNNs exacerbate performance degradation due to adversarial noise, and add the regularization term to the loss to keep the Lipschitz constant of each layer small. The regularization term is as follows:
\begin{align}
L_{Lip} = \frac{1}{2} \sum_{W_l} ||W_{l}^T W_{l} - I ||^2 . \label{eq:lip}
\end{align}
Note that convolution kernels are reshaped to $(k\times k \times c_{in}, c_{out})$ where $k$, $c_{in}$, and $c_{out}$ are the kernel size, input channels, and output channels, respectively. $L_{Lip}$ was applied to enhance the adversarial attack robustness of activation quantized DNNs \cite{lin2019defensive}. We show that $L_{Lip}$ can reduce the noise due to activation quantization itself. Also, we compare the effect of the regularization term on the performance of weight quantized DNNs.

\section{Experimental Results}
\label{sec:experiment}

\subsection{Visualizing the Effects of Quantization on the Segmentation Task}

\begin{figure*}[t]
    \centering
    \subfigure[Image] {\includegraphics[width=0.2\linewidth]{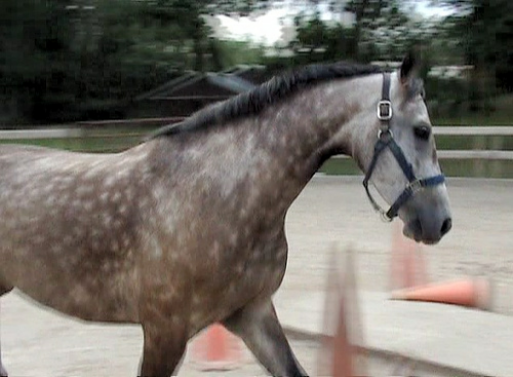}}
    \hskip 0.02\linewidth
    \subfigure[Float] {\includegraphics[width=0.2\linewidth]{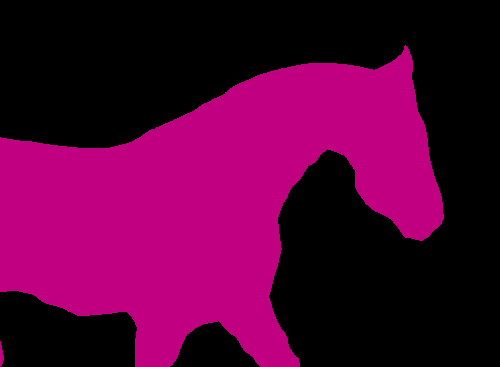}}
    \hskip 0.02\linewidth
    \subfigure[2-bit W] {\includegraphics[width=0.2\linewidth]{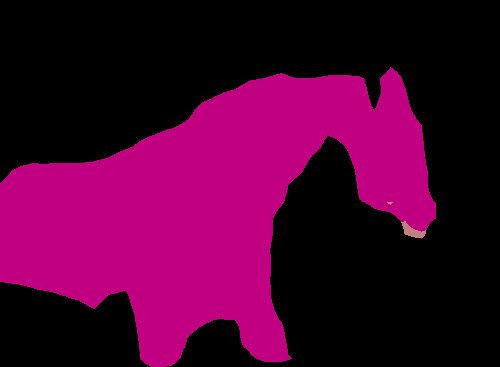}}
    \hskip 0.02\linewidth
    \subfigure[2-bit A]
    {\includegraphics[width=0.2\linewidth]{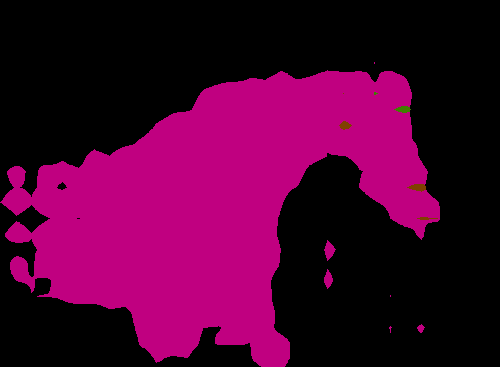}}
    \\
    \subfigure[Image] {\includegraphics[width=0.2\linewidth]{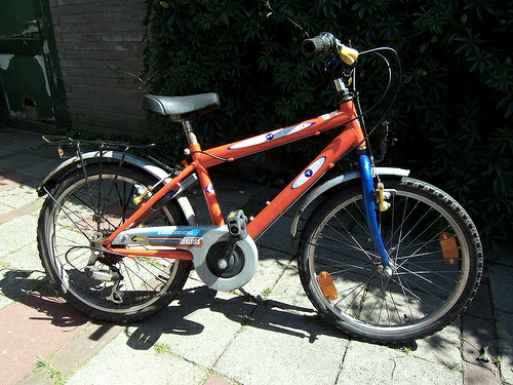}}
    \hskip 0.02\linewidth
    \subfigure[Float] {\includegraphics[width=0.2\linewidth]{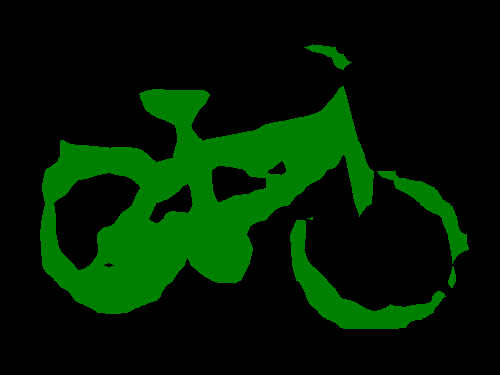}}
    \hskip 0.02\linewidth
    \subfigure[2-bit W] {\includegraphics[width=0.2\linewidth]{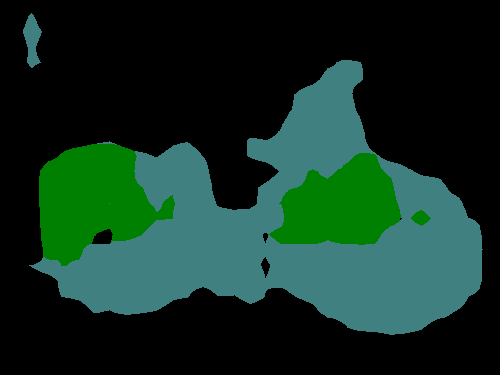}}
    \hskip 0.02\linewidth
    \subfigure[2-bit A] {\includegraphics[width=0.2\linewidth]{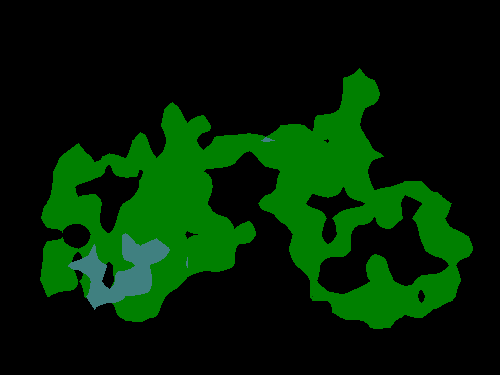}}
    
    \caption{Visualization of quantization errors on the PASCAL VOC segmentation benchmark. W and A are abbreviations for the weight and activation, respectively. Activation outputs are retained in floating-point precision on weight quantized model, and vice-versa.}
    \label{fig:vis_seg}
\end{figure*}

We first visualize the weight and activation quantization effects using a segmentation task. The PASCAL VOC 2012 dataset~\cite{everingham2015pascal} is used. The dataset contains 1,464 training images and 1,449 validation images. Each image is labeled at pixel-level with 20 object classes and a background class. The MobileNetV2 \cite{sandler2018mobilenetv2} is adopted, which is trained according to DeepLabV3~\cite{chen2017rethinking} with 10,582 augmented training images~\cite{hariharan2011semantic}~\footnote{We obtained the pretrained model from https://github.com/tensorflow/models/tree /master/research/deeplab}. The model is used as a floating-point pretrained model after fine-tuned using the original 1,464 training images for 30K iterations. The output stride is 16 and Atrous Spatial Pyramid Pooling (ASPP) \cite{chen2017deeplab} is not applied. The performance is measured in terms of mean intersection over union (mIOU) without multi-scaling and flipping input images. Only the original training images are used for retraining and fine-tuning. The retraining is conducted for 30K iterations with a batch size of 16. The initial learning rate is 1e-3 and the learning rate policy is the same as \cite{chen2017rethinking}. 

The segmentation results of the retrained QDNNs are visualized in~\figurename~\ref{fig:vis_seg}. Either weights or activations are quantized to 2 bits. When the object is simple, as shown in~\figurename~\ref{fig:vis_seg} (a), the weight quantized model seems to perform the segmentation fairly well. However, the results with the activation quantized model contains some noise on the section where the background and the object colors are similar. In the segmentation of a complex one, the weight quantized model fails to find the characteristics of the object, as shown in \figurename~\ref{fig:vis_seg} (e). We can even consider that the activation noise corrupted model segments the bicycle more faithfully than the weight quantized model. The visualization results imply that weight quantization degrades the generalization ability, and activation quantization induces noise. The experiment with the segmentation task confirms the observation with the synthetic dataset in Section~\ref{sec:analysis}. More segmentation results of QDNNs are compared in Appendix~\textbf{C}.

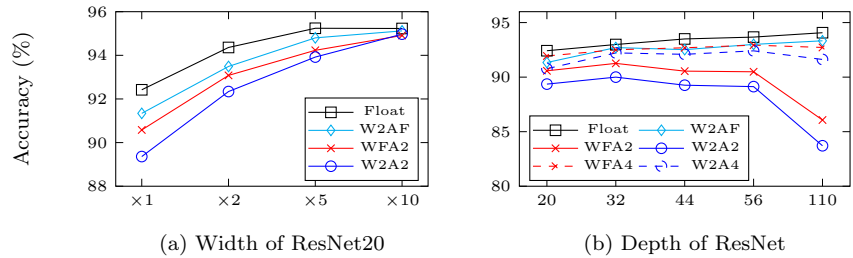
\begin{figure}[t]
\centering
\tikzset{mark options={mark size=2}}
\subfigure{
\begin{tikzpicture}
    \begin{axis}[
	width=0.47\linewidth,
	height = 0.32\linewidth,
	ymin=88.0,
	ymax=96.0,
	xtick=data,
	xticklabels= {$\times 1$ , $\times 2$, $\times 5$, $\times 10$},
	xtick align=inside, 
	major tick style={line width=0.010cm, black, font=\fontsize{4}{4}\selectfont},
	major tick length=0.10cm,
    xlabel= (a) Width of ResNet20,
    ylabel=Accuracy (\%),
    legend pos=south east,
    legend columns=1,
    tick label style={font=\fontsize{6}{4}\selectfont},
    label style={font=\fontsize{8}{4}\selectfont},
    legend style={font=\fontsize{4}{1}\selectfont, inner sep=1pt, row sep=-2pt},
    ]]
	\addplot[color=black,solid, mark=square]
	coordinates {(0, 92.42) (1, 94.36) (2, 95.24) (3, 95.23)};
	\addplot[color=cyan, solid,mark=diamond]
	coordinates {(0, 91.34) (1, 93.49) (2, 94.80) (3, 95.12)};
	\addplot[color=red,solid,mark=x]
	coordinates {(0, 90.58) (1, 93.08) (2, 94.23) (3, 94.91)};
	\addplot[color=blue, solid, mark=o]
	coordinates {(0, 89.36) (1, 92.34) (2, 93.92) (3, 94.97)};
	\legend{Float, W2AF, WFA2, W2A2}
    \end{axis}
  \end{tikzpicture}}
  \hskip 0.2in
  \subfigure{
\begin{tikzpicture}
    \begin{axis}[
	width=0.49\linewidth,
	height = 0.32\linewidth,
	ymin=80.0,
	ymax=96.0,
	xtick=data,
	xticklabels= {20 , 32, 44, 56, 110},
	xtick align=inside, 
	major tick style={line width=0.010cm, black},
	major tick length=0.10cm,
    xlabel= (b) Depth of ResNet,
    legend pos=south west,
    legend columns=2,
    tick label style={font=\fontsize{6}{4}\selectfont},
    label style={font=\fontsize{8}{4}\selectfont},
    legend style={font=\fontsize{4}{1}\selectfont, inner sep=1pt, row sep=-2pt},
    ]]
	\addplot[color=black,solid, mark=square]
	coordinates {(0, 92.42) (1, 92.99) (2, 93.50) (3, 93.67) (4, 94.07)};
	\addplot[color=cyan, solid,mark=diamond]
	coordinates {(0, 91.34) (1, 92.71) (2, 92.52) (3, 93.00) (4, 93.33)};
	\addplot[color=red,solid,mark=x]
	coordinates {(0, 90.58) (1, 91.26) (2, 90.55) (3, 90.49) (4, 86.07)};
	\addplot[color=blue, solid, mark=o]
	coordinates {(0, 89.36) (1, 90.00) (2, 89.26) (3, 89.13) (4, 83.71)};
	\addplot[color=red,dashed, mark=x]
	coordinates {(0, 91.94) (1, 92.51) (2, 92.69) (3, 92.93) (4, 92.72)};
	\addplot[color=blue,dashed, mark=o]
	coordinates {(0, 90.77) (1, 92.22) (2, 92.09) (3, 92.41) (4, 91.62)};
	\legend{Float, W2AF, WFA2, W2A2, WFA4, W2A4}
    \end{axis}
  \end{tikzpicture}}
\caption{Performance of quantized ResNet on the CIFAR-10 testset according to the (a) width of the ResNet20 and (b) depth of the ResNet. Legends represent the precision of `weights' (W) and `activation' (A). `F' denotes the floating-point precision.}
\label{fig:resnet_width_depth}
\end{figure}
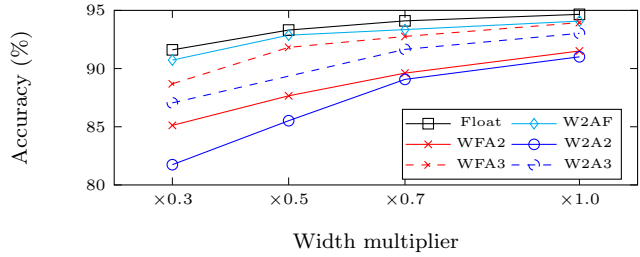
\begin{figure}[t]
\centering
\tikzset{mark options={mark size=2}}
\subfigure{
\begin{tikzpicture}
    \begin{axis}[
	width=0.7\linewidth,
	height = 0.32\linewidth,
	ymin=80.0,
	ymax=95.0,
	xtick=data,
	xmin=0.2,
	xmax=1.1,
	xticklabels= {$\times 0.3$ , $\times 0.5$, $\times 0.7$, $\times 1.0$},
	xtick align=inside, 
	major tick style={line width=0.010cm, black, font=\fontsize{4}{4}\selectfont},
	major tick length=0.10cm,
    xlabel= Width multiplier,
    ylabel=Accuracy (\%),
    legend pos=south east,
    legend columns=2,
    tick label style={font=\fontsize{6}{4}\selectfont},
    label style={font=\fontsize{8}{4}\selectfont},
    legend style={font=\fontsize{4}{1}\selectfont, inner sep=1pt, row sep=-1pt},
    ]]
	\addplot[color=black,solid, mark=square]
	coordinates {(0.3, 91.61) (0.5, 93.31) (0.7, 94.1) (1.0,94.66)};
	\addplot[color=cyan, solid,mark=diamond]
	coordinates {(0.3, 90.73) (0.5, 92.87) (0.7, 93.35) (1.0, 94.08)};
	\addplot[color=red,solid,mark=x]
	coordinates {(0.3, 85.12) (0.5, 87.66) (0.7, 89.61) (1.0, 91.51)};
	\addplot[color=blue, solid, mark=o]
	coordinates {(0.3, 81.74) (0.5, 85.52) (0.7, 89.07) (1.0, 91.01)};
	\addplot[color=red,dashed, mark=x]
	coordinates {(0.3, 88.69) (0.5, 91.82) (0.7, 92.76) (1.0, 93.95)};
	\addplot[color=blue,dashed, mark=o]
	coordinates {(0.3, 87.05) (0.7, 91.65) (1.0, 93.03)};
	\legend{Float, W2AF, WFA2, W2A2, WFA3, W2A3}
    \end{axis}
  \end{tikzpicture}}

\caption{CIFAR-10 test accuracy (\%) of MobileNetV2 according to the width multiplier.}
\label{fig:mobilenet_width}
\end{figure}

\begin{figure}[t]
\centering
\tikzset{mark options={mark size=2}}
\subfigure[Basic block]{
\begin{tikzpicture}
    \begin{axis}[
	width=0.37\linewidth,
	height = 0.28\linewidth,
	ymin=87.0,
	ymax=95.0,
	xtick=data,
	xticklabels= {8, 14, 20, 32, 44, 56},
	xtick align=inside, 
	major tick style={line width=0.010cm, black, font=\fontsize{4}{4}\selectfont},
	major tick length=0.10cm,
    xlabel=Number of layers,
    ylabel=Accuracy (\%),
    legend pos=south west,
    legend columns=1,
    tick label style={font=\fontsize{6}{4}\selectfont},
    label style={font=\fontsize{8}{4}\selectfont},
    legend style={font=\fontsize{4}{1}\selectfont, inner sep=1pt, row sep=-3pt},
    ]]
	\addplot[color=black,solid, mark=square]
	coordinates {(0, 93.26) (1, 93.62) (2, 94.14) (3, 94.24) (4, 94.27) (5, 93.58)};
	\addplot[color=cyan, solid,mark=diamond]
	coordinates {(0, 92.83) (1, 93.51) (2, 93.57) (3, 93.71) (4, 93.91) (5, 93.46)};
	\addplot[color=red,solid,mark=x]
	coordinates {(0, 92.58) (1, 93.30) (2, 92.94) (3, 92.42) (4, 91.4) (5, 90.58)};
	\addplot[color=blue, solid, mark=o]
	coordinates {(0, 92.32) (1, 93.11) (2, 92.65) (3, 91.84) (4, 90.69) (5, 89.77)};
	\legend{Float, W2AF, WFA2, W2A2}
    \end{axis}
  \end{tikzpicture}}
\subfigure[Pre-activation block]{
\begin{tikzpicture}
    \begin{axis}[
	width=0.37\linewidth,
	height = 0.28\linewidth,
	ymin=87.0,
	ymax=95.0,
	xtick=data,
	xticklabels= {8, 14, 20, 32, 44, 56},
	xtick align=inside, 
	major tick style={line width=0.010cm, black, font=\fontsize{4}{4}\selectfont},
	major tick length=0.10cm,
    xlabel= Number of layers,
    tick label style={font=\fontsize{6}{4}\selectfont},
    label style={font=\fontsize{8}{4}\selectfont},
    ]]
	\addplot[color=black,solid, mark=square]
	coordinates {(0, 93.1) (1, 94.14) (2, 94.25) (3, 94.07) (4, 94.31) (5, 93.96)};
	\addplot[color=cyan, solid,mark=diamond]
	coordinates {(0, 92.75) (1, 93.85) (2, 93.82) (3, 93.89) (4, 94.04) (5, 93.83)};
	\addplot[color=red,solid,mark=x]
	coordinates {(0, 91.89) (1, 92.44) (2, 92.42) (3, 91.51) (4, 90.58) (5, 88.45)};
	\addplot[color=blue, solid, mark=o]
	coordinates {(0, 91.03) (1, 91.81) (2, 91.81) (3, 89.03) (4, 86.26) (5, 85.45)};
    \end{axis}
  \end{tikzpicture}}
\subfigure[Depthwise block]{
\begin{tikzpicture}
    \begin{axis}[
	width=0.37\linewidth,
	height = 0.28\linewidth,
	ymin=87.0,
	ymax=95.0,
	xtick=data,
	xticklabels= {29, 38, 56, 74, 110},
	xtick align=inside, 
	major tick style={line width=0.010cm, black, font=\fontsize{4}{4}\selectfont},
	major tick length=0.10cm,
    xlabel=Number of layers,
    tick label style={font=\fontsize{6}{4}\selectfont},
    label style={font=\fontsize{8}{4}\selectfont},
    ]]
	\addplot[color=black,solid, mark=square]
	coordinates {(0, 93.64) (1, 93.65) (2, 94.27) (3, 93.94) (4, 94.43)};
	\addplot[color=cyan, solid,mark=diamond]
	coordinates {(0, 93.28) (1, 93.47) (2, 94.02) (3, 93.71) (4, 94.39)};
	\addplot[color=red,solid,mark=x]
	coordinates {(0, 93.38) (1, 92.93) (2, 91.7) (3, 90.5) (4, 88.75)};
	\addplot[color=blue, solid, mark=o]
	coordinates {(0, 92.33) (1, 92.57) (2, 90.61) (3, 89.88) (4, 87.47)};
    \end{axis}
  \end{tikzpicture}}
\caption{CIFAR-10 test accuracy (\%) of quantized ResNet according to the block types with varying depth and width. Note that the number of parameters of all experimented models are about 1M. }
\label{fig:resnet_blocks}
\end{figure}
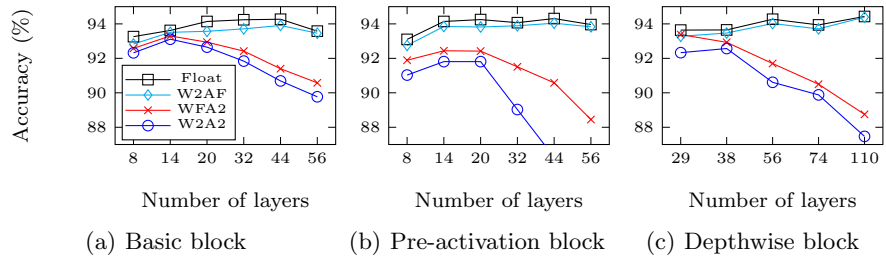

\subsection{The Width and Depth Effects on QDNNs}
\label{ch3:subsec:widthdepth}
We analyze the weight and activation quantization sensitivities when the depth and width of ResNet vary using the CIFAR-10 dataset~\cite{krizhevsky2009learning}. The depth refers the number of layers and the width is that of channels in a CNN. Simple data augmentation techniques, cropping and flipping, are applied as suggested in~\cite{lee2015deeply}. The batch size is 128 and the number of epochs for pretraining is 200. The SGD optimizer with a momentum of 0.9 is used. The learning rate starts at 0.1 and decays by 0.1 times at 100 and 150 epochs. The L2 regularization is applied with a scale of 5e-4. Quantized models are retrained for 100 epochs with the initial learning rate of 0.01, and the learning rate decreases by a factor of 0.1 at 50 and 80 epochs. We do not employ the L2 regularization when retraining of the quantized model. The number of layers is either 20, 32, 56, or 110 and the width multiplier ranges from x1 to x10 times of the original ResNet. The experimental model denoted as ResNet$\times I$ employs $ I $ times larger number of channels. The width-expanded ResNets are compared in~\figurename~\ref{fig:resnet_width_depth} (a). As the width of the ResNet20 increases, both weight and activation quantization errors decrease. The performance of ResNets when the depth of layers increases is shown in~\figurename~\ref{fig:resnet_width_depth} (b). When the depth increases, the 2-bit weight quantized models show improved performance, but those with 2-bit quantized activations exhibit degraded performance. At least, 4-bit activation quantization is needed for deep ResNet.

We also evaluate the quantization sensitivity of the MobileNetV2~\cite{sandler2018mobilenetv2}. The width multiplier is used to control the number of parameters of MobileNetV2. The MobileNetV2 performance on CIFAR-10 with various width-multiplication factors is shown in ~\figurename~\ref{fig:mobilenet_width}. Note that the MobileNetV2$\times1.0$ is the same with the original model except that the first stride of 2 is replaced to 1. As the width decreases, the performances of full-precision and weight quantized models degrade gradually. However, the activation quantized models exhibit severe performance loss as the width decreases. The results indicate that DNNs with small widths are more vulnerable to the activation quantization.


\subsection{QDNN Architecture Selection under the Parameter Constraint}

We compare the quantization sensitivity of CNNs according to the model structure under the constraint on the number of parameters. CNNs with three different block types: basic, pre-activation, depthwise blocks are evaluated using CIFAR-10 dataset. The number of parameters of these models is approximately one million. The performance of the quantized CNNs is shown in~\figurename~\ref{fig:resnet_blocks}. The experimented model configurations are summarized in Appendix~\textbf{D}. When the number of parameters is comparable, increasing the depth to a certain range helps to improve the performance of full precision and weight quantized models. However, the models with 2-bit activation quantization show poor performances when the depth increases beyond certain numbers. We can also find that depthwise blocks are more robust to the activation quantization. The 2-bit CNNs with the basic or pre-activation blocks show severe performance degradation when the depth is over 32 while the performance with depthwise blocks is improves until the depth of 38. For the CIFAR-10 dataset, the best performing quantized ResNet can be designed by choosing the depth of 14 with the basic blocks or 38 with the depthwise ones.

We change the depth and width of the ResNets and evaluate the quantization performance on the ImageNet dataset~\cite{russakovsky2015imagenet}. Data augmentation methods and hyperparameters are the same as~\cite{jung2019learning}. Pre-activation blocks are employed and the shortcut signals are quantized to 8 bits. 4-level 2-bit weight quantization is applied. The experimented ResNet structures are compared in~\tablename~\ref{ch3:table:imagenet}. Since ResNets for ImageNet classification have four groups of residual blocks, which are separated by the convolution layers with the stride of 2, the number of blocks is represented as a list: [first, second, third, fourth] groups. For example, ResNet18$\times1.4$ means that the depth is 18 with 8 number of blocks and the initial channel width is 90. The number of channels increases by a factor of 2 at every convolution layer with the stride of 2. 

The performance degradation by the quantization is much lower when the model is shallow under a comparable number of parameter. However, the top-1 accuracy of 2-bit ResNet18$\times 1.4$ is 0.3\% lower than the 2-bit ResNet34 because the floating-point performance is too low. When the depth, $L$, is 26 and the initial channel width, $D_{init}$ is 70, the top-1 accuracy of the floating-point model is 0.6\% lower but the 2-bit quantized model achieves 0.1\% higher top-1 accuracy when compared to the ResNet34. The performance improvements is more significant when the model is deeper. The top-1 accuracy degradation of 2-bit ResNet50 is 4.9~\cite{zhang2018lq} and 3.6(Ours). However, 2-bit ResNet44$\times 1.1$ shows 2.1\% top-1 accuracy drop. As a result, 2-bit ResNet44$\times 1.1$ achieves the 0.7\% top-1 accuracy improvement compared to the ResNet50 with the similar number of parameters and operations. The floating-point ResNet101 shows the top-1 accuracy of 77.5\%, which outperforms shallow and wider models with a comparable number of parameters. However, the 2-bit quantization of ResNet101 degrades the performance significantly, showing only about 20.1\% top-1 accuracy. As observed in Section~\ref{ch3:subsec:widthdepth}, the 2-bit activation quantized model degrades the performance dramatically when the model is very deep. On the other hand, the ResNet with the depth of 50 achieves 73.5\% top-1 accuracy even after 2-bit quantization. These results indicate that the performance of QDNNs can be improved simply by designing proper depth and width of the model.

\begin{table}[t]
\centering
\caption{Performance comparison of 2-bit ResNet on ImageNet validation set. `$L$' and `$D_{init}$' represent the number of layers and initial channel dimension, respectively. `$B$' is the stack of the residual blocks and represented separately according to the stride of 2.}
\setlength\tabcolsep{4pt}
\begin{tabular}{lccccc|ccc}
\toprule
\multicolumn{6}{c|}{Model}                                                                                               & \multicolumn{3}{c}{Top-1 Acc (\%)} \\ 
Type & $L$ & $D_{init}$ & $B$ & \begin{tabular}[c]{@{}c@{}}Params.\\ ($10^6$)\end{tabular} & \begin{tabular}[c]{@{}c@{}}Ops.\\ ($10^9$)\end{tabular} & Float & W2A2 & Diff. \\ \midrule
ResNet34~\cite{zhang2018lq} & 34 & 64 & {[}3, 4, 6, 3{]} & 21.8 & 3.7 & 73.8 & 69.8 & 4.0 \\ 
ResNet34~\cite{gong2019differentiable} & 34 & 64 & {[}3, 4, 6, 3{]} & 21.8 & 3.7 & 73.8 & 70.0  & 3.8 \\ 
ResNet34 & 34 & 64 & {[}3, 4, 6, 3{]} & 21.8 & 3.7 & 73.6 & 70.5 & 3.1 \\
ResNet18$\times1.4$ & 18 & 90 & {[}2, 2, 2, 2{]} & 22.8 & 3.5 & 72.6 & 70.2 & \textbf{2.4} \\
ResNet26$\times1.1$ & 26 & 70 & {[}3, 3, 3, 3{]} & 21.4& 3.3  & 73.0 & \textbf{70.6}  & \textbf{2.4} \\ \midrule
ResNet50~\cite{zhang2018lq} & 50 & 64 & {[}3, 4, 6, 3{]} & 25.6 & 4.1 & 76.4 & 71.5 & 4.9 \\ 
ResNet50 & 50 & 64 & {[}3, 4, 6, 3{]} & 25.6 & 4.1 & 76.3 & 72.7 & 3.6 \\ 
ResNet44$\times1.1$ & 44 & 72 & {[}3, 4, 5, 2{]} & 25.0 & 4.6 & 75.5 & \textbf{73.4} & \textbf{2.1} \\ \midrule
ResNet101 & 101 & 64 & {[}3, 4, 23, 3{]} & 44.6     & 7.9 & 77.5             & 20.1 & 57.4               \\
ResNet50$\times1.3$ & 50 & 85 & {[}3, 4, 6, 3{]} & 44.2 & 7.2 & 77.2 & \textbf{73.7} & \textbf{3.5} \\ \bottomrule

\end{tabular}
\label{ch3:table:imagenet}
\vskip 0.3in
\end{table}
\begin{table}[t]
\centering
\caption{Performance improvements of quantized MobileNetV2 on the PASCAL VOC 2012 validation set. $n_W$ and $n_A$ represent the precision of the weights and activations, respectively.}
\setlength\tabcolsep{10pt}
\begin{tabular}{lcc|c}
\toprule
Method     & $n_W$    & $n_A$    & mIOU           \\ \midrule
Pretrained model & Float & Float & 76.97          \\ 
\midrule
Retrain (baseline)   & 4-bit & Float & 73.63          \\
Fine-tune with CLR        & 4-bit & Float & \textbf{74.15} \\
Retrain with $L_{Lip}$       & 4-bit & Float & 73.22          \\ 
\midrule
Retrain (baseline)   & Float & 4-bit & 74.79          \\
Fine-tune with CLR        & Float & 4-bit & 74.71          \\
Retrain with $L_{Lip}$       & Float & 4-bit & \textbf{74.99} \\ \bottomrule
\end{tabular}
\label{table:seg_training}
\end{table}

\begin{table}[t]
\centering
\caption{Performance in terms of mIOU on the PASCAL VOC 2012 validation set when both weights and activations are quantized.}
\setlength\tabcolsep{10pt}
\begin{tabular}{l|cccc}
\toprule
                                 & \multicolumn{4}{c}{$n_W$ / $n_A$ (bits)}           \\ 
 \multicolumn{1}{c|}{Method}     & 8/8   & 6/6   & 4/4   & 3/3    \\ \midrule
Retrain (baseline)               & \textbf{76.56} & 75.79 & 71.16 & 59.64  \\
Retrain ($L_{Lip}$)              & 76.34 & 76.23 & 71.93 & 59.85  \\
$L_{Lip}$ + CLR                  & 76.46 & \textbf{76.40}      & \textbf{72.56} & \textbf{60.74}  \\ \bottomrule
\end{tabular}
\label{table:seg_training_both}
\end{table}

\begin{table}[t]
\centering
\caption{CIFAR-10 test accuracy (\%) improvements when retrained with $L_{Lip}$ and fine-tuned using CLR.}
\setlength\tabcolsep{10pt}
\begin{tabular}{c|cccc}
\toprule
   & \multicolumn{4}{c}{ResNet depth} \\ 
 $n_W$, $n_A$     & 20  & 32       & 56    & 110   \\ \midrule
 Float     & 92.42 & 92.99  & 93.67 & 94.07 \\
 2-bit     & 89.36 & 90.00  & 89.13 & 83.71        \\
 2-bit ($L_{Lip}$ + CLR)      & 89.68 & 90.60  & 90.24 & 87.77        \\
\bottomrule
\end{tabular}
\label{table:asq_cifar10}
\end{table}


\subsection{Results of Training Methods on QDNNs}

We assess the effects of the training methods on QDNN optimization using the segmentation task. The effects of applying the CLR for improved generalization and adding the regularization term, $L_{Lip}$, for noise robustness are summarized in Table ~\ref{table:seg_training}. The mIOU of the floating-point pretrained model is 76.97, that of the retrained 4-bit weight is 73.63, and that of retrained 4-bit activation is 74.79. CLR is applied with a period of 3K iterations and fine-tuning was performed for 15K iterations (i.e. 5 cycles). $L_{Lip}$ is added to the loss after multiplying the scaling factor of 1e-4. Applying the CLR increases the mIOU of the 4-bit weight quantized model by 0.5 but it is not effective for the activation quantized model. Retraining with the regularization term that reduces the Lipschitz constant improves the mIOU of the 4-bit activation quantized model. However, the performance of the weight quantized model is decreased when $L_{Lip}$ is applied. The performances of QDNNs when both weights and activations are quantized are shown in Table~\ref{table:seg_training_both}. The results show that the proposed approach for reducing the quantization effects of weights by CLR and activations by adding the Lipschitz loss works well when the precision is equal to or lower than 6-bit. But these techniques are not effective when the precision of quantization is 8-bit or larger. Even, the $L_{Lip}$ regularization degrades the performance of the 8-bit quantized model.

We also evaluate the effects of the training methods for the classification task. The performance degradation by severe quantization can be alleviated with $L_{Lip}$ and CLR as shown in Table~\ref{table:asq_cifar10}. $L_{Lip}$ is added to the loss with a factor of 1e-4 and CLR is applied for 40 epochs with the learning rates between 1e-3 and 1e- 5. Although we can improve the performance of QDNN considerably for deep networks by applying CLR and $L_{Lip}$ constraint, the best performing QDNN can be found when the depth is 32. 


\section{Concluding remarks}
\label{sec:conclude}


We have presented a holistic approach for the optimization of quantized deep neural networks (QDNNs). We first visualize the effects of the weight and activation quantization error using a synthetic dataset. The result clearly shows that the effects of weight and activation quantization are different. Especially, activation quantization severely degrades the performance of deep models. Through additional experiments with real tasks, we confirm that the optimal model structure under a parameter constraint is different for the full-precision and quantized DNNs because floating-point models usually prefer the deep networks but QDNNs tend to show improved performances on wide networks. We also show the effects of the DNN training schemes for improved generalization and noise reduction to optimize QDNNs. The proposed holistic approach can yield much better QDNN when compared to the conventional design approaches that start from the best performing floating-point models and optimize them using elaborate quantization and training methods.



\clearpage
%
%
\bibliographystyle{splncs04}
\bibliography{egbib}

\clearpage

\vskip 0.3in
\section*{\centering Appendix}
\vskip 0.2in

\subsection*{A. Additional Visualization Examples on the Synthetized Dataset}

\begin{figure}[h]
    \centering
    \subfigure[200-4]{
        \includegraphics[width=.175\linewidth]{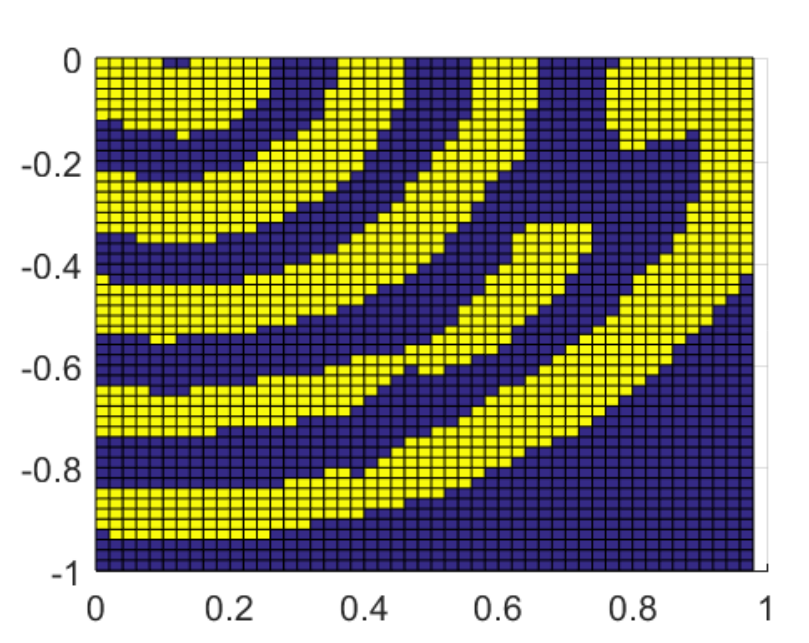}
    }
    \subfigure[384-4]{
        \includegraphics[width=.175\linewidth]{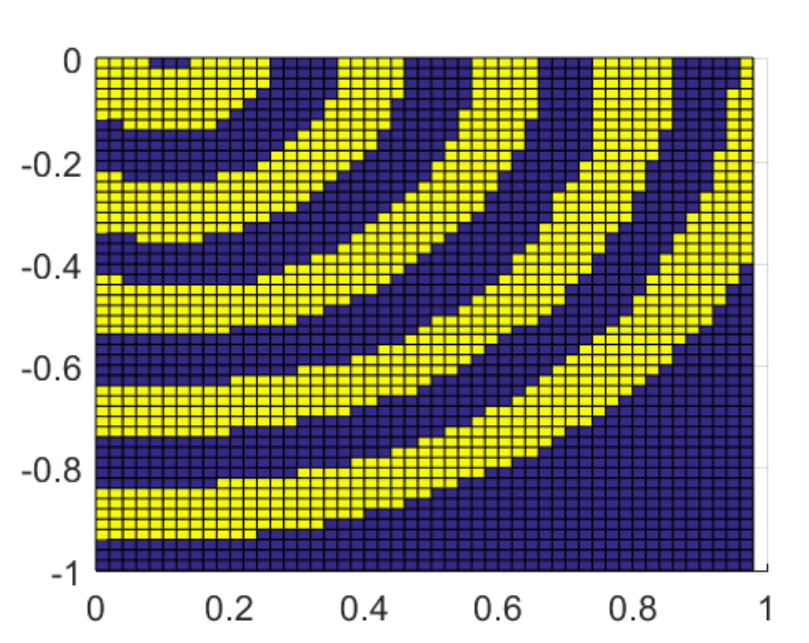}
        }
    \subfigure[128-5]{
        \includegraphics[width=.175\linewidth]{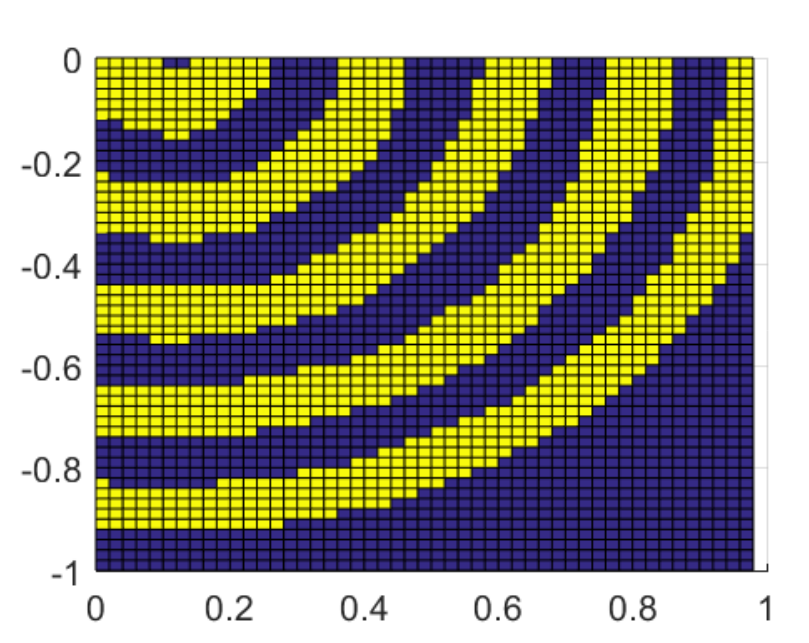}
    }
    \subfigure[128-6]{
        \includegraphics[width=.167\linewidth]{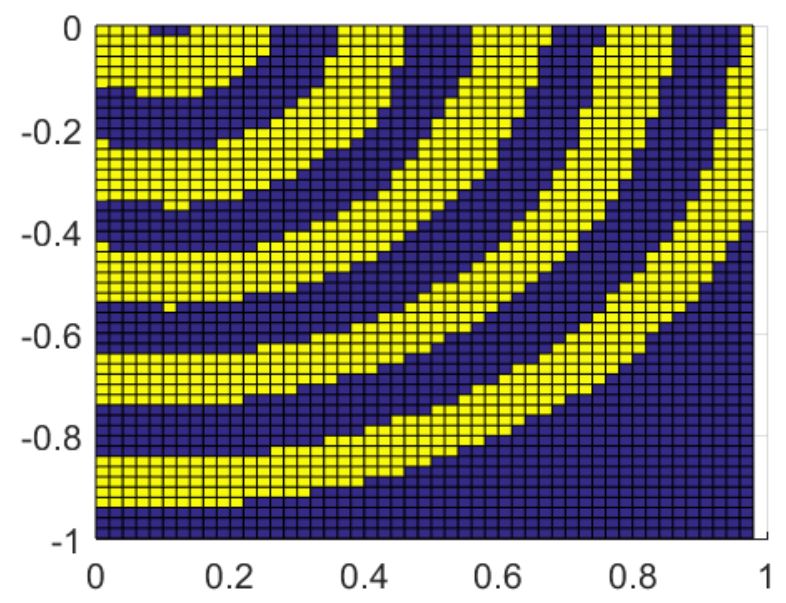}
    }
    \subfigure[128-8]{
        \includegraphics[width=.165\linewidth]{images2/128_8_w_2_a_32bits.pdf}
    }        
    
    \subfigure[200-4]{
        \includegraphics[width=.175\linewidth]{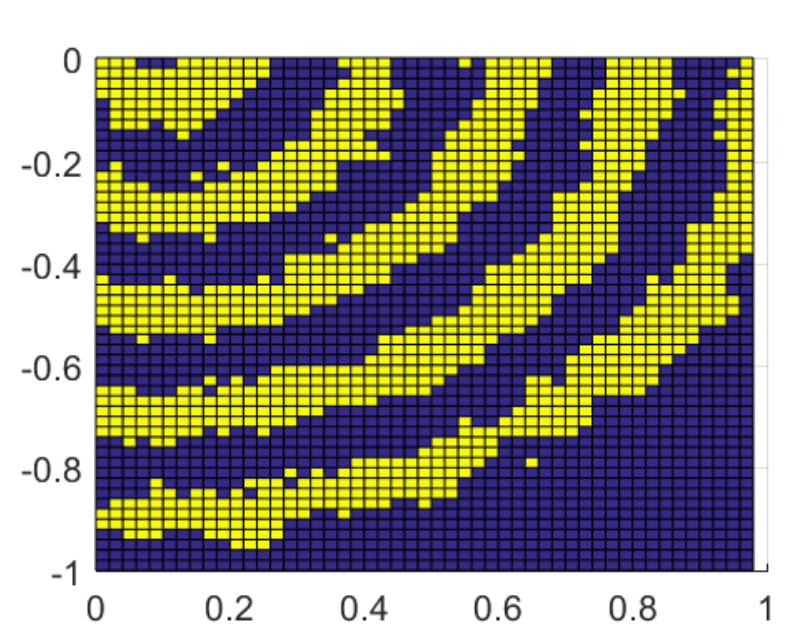}
    }    
    \subfigure[384-4]{
        \includegraphics[width=.175\linewidth]{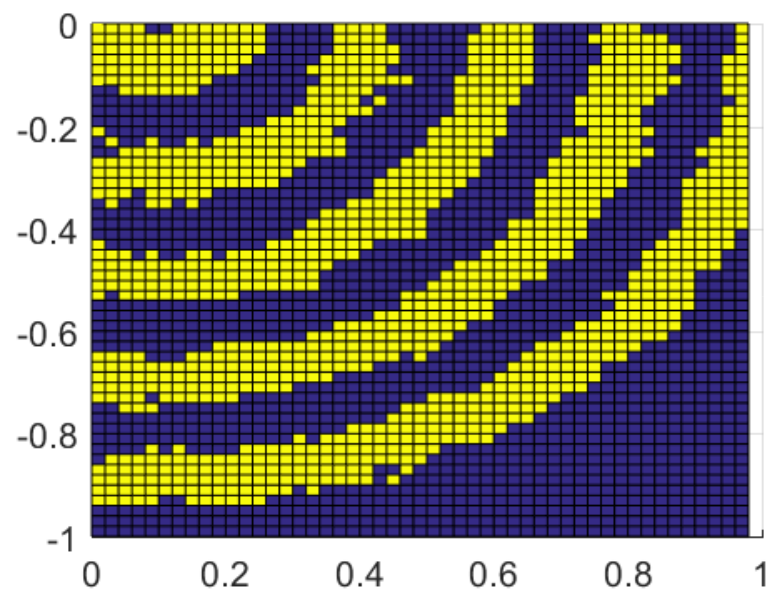}
    }    
    \subfigure[128-5]{
        \includegraphics[width=.175\linewidth]{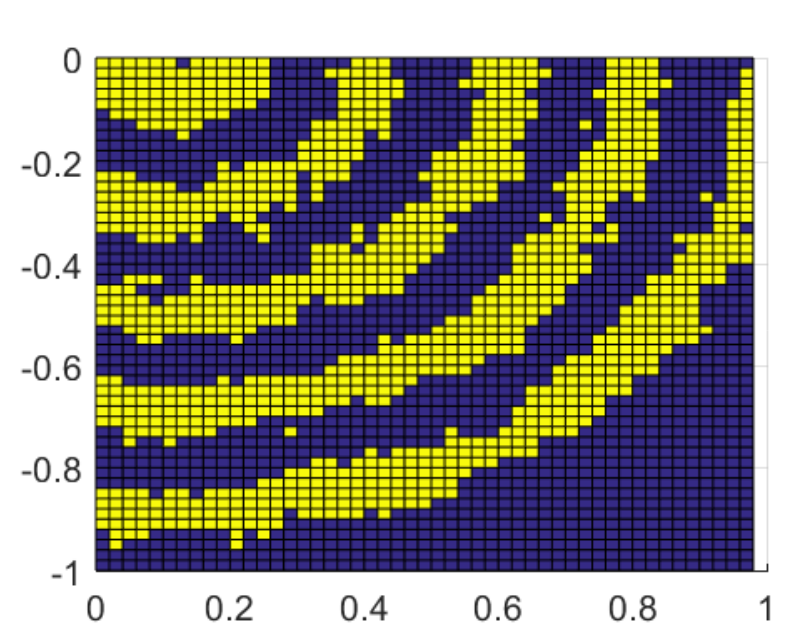}
        }
    \subfigure[128-6]{
        \includegraphics[width=.17\linewidth]{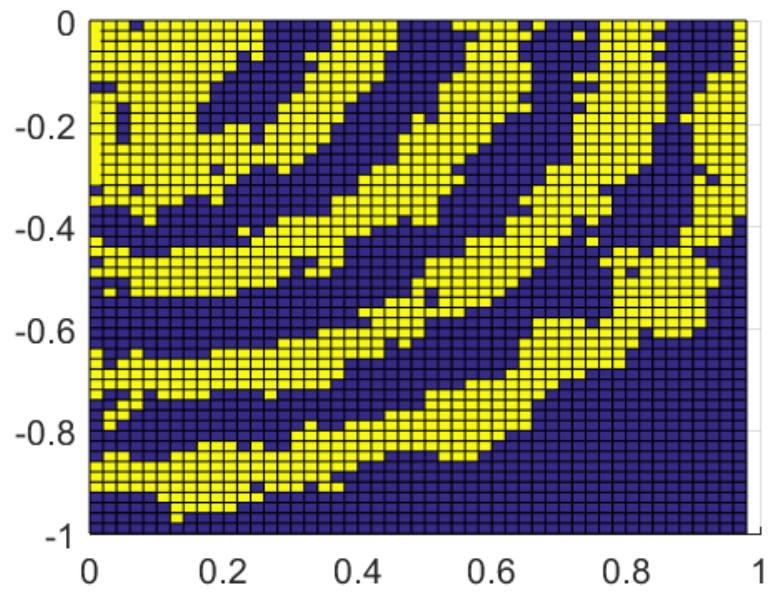}
        }
    \subfigure[128-8]{
        \includegraphics[width=.17\linewidth]{images2/128_8_w_32_a_2bits.pdf}
        }
    %
    \subfigure[200-4]{
        \includegraphics[width=.17\linewidth]{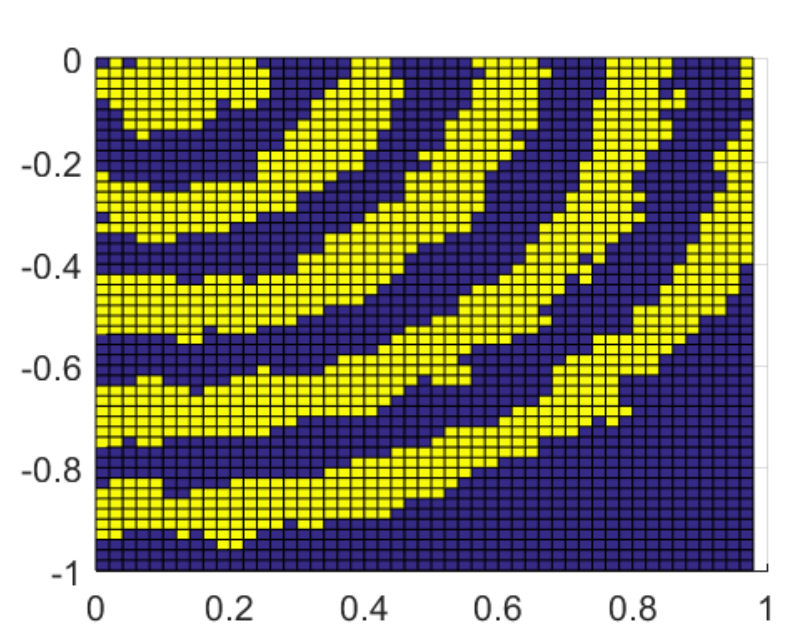}
        }
    \subfigure[384-4]{
        \includegraphics[width=.17\linewidth]{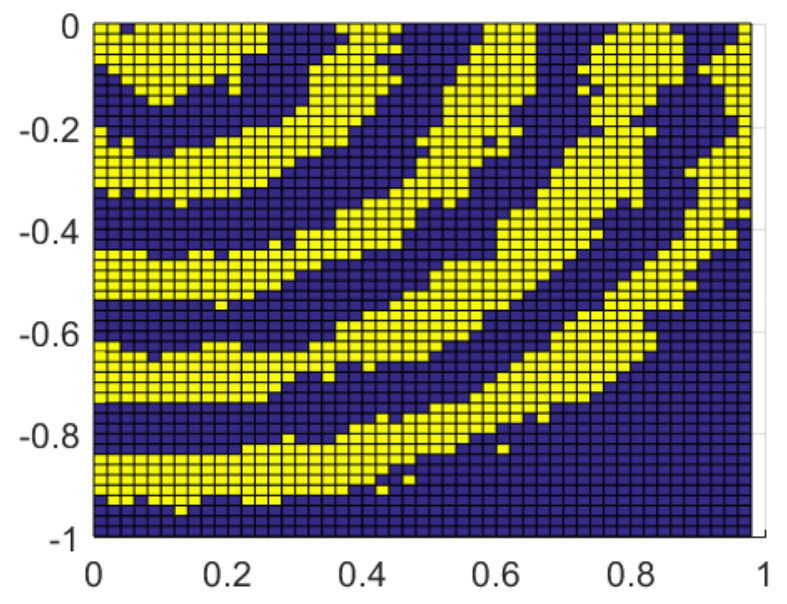}
        }
    \subfigure[128-5]{
        \includegraphics[width=.17\linewidth]{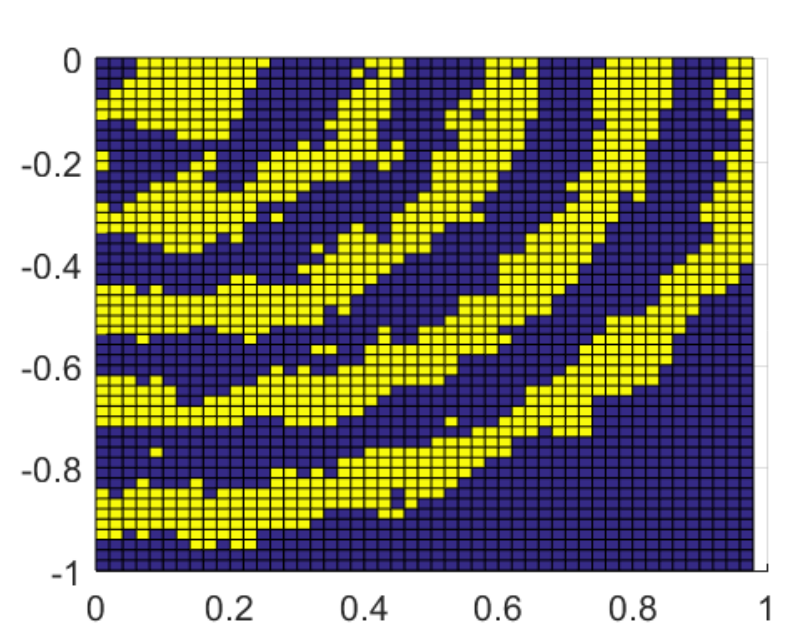}
        }
    \subfigure[128-6]{
        \includegraphics[width=.17\linewidth]{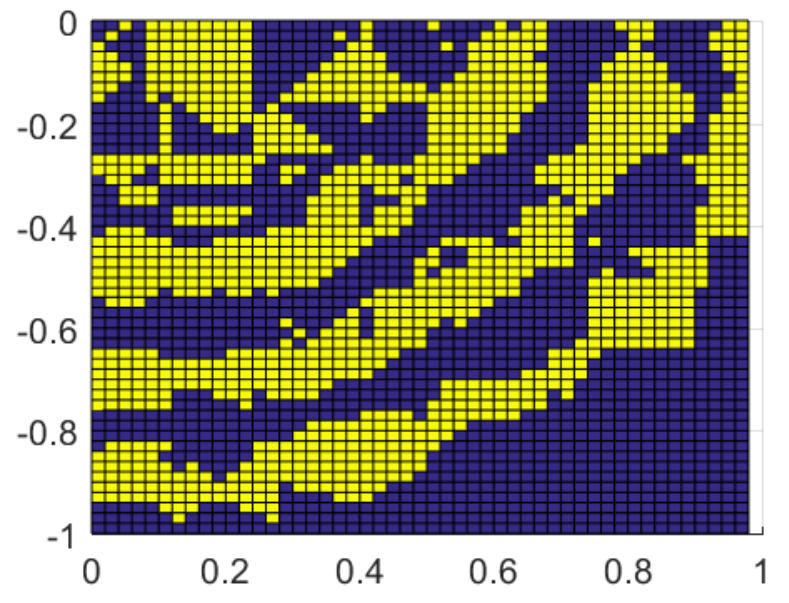}
        }
    \subfigure[128-8]{
        \includegraphics[width=.17\linewidth]{images2/128_8_w_2_a_2bits.pdf}
        }
    \caption{Synthetic data visualization results on FCDNNs as the width and depth increase. The experimented FCDNNs are denoted as `$D$'-`$H$', where $D$ is the dimension of hidden layers and $H$ is the number of layers. The models are either 2-bit wegiht quantized (\textbf{first row}), 2-bit activation quantized (\textbf{second row}), or 2-bit weight and activation quantized (\textbf{third row}).}
    \label{app_fig:toy_example_depth_width}
\end{figure}

Additional examples of visualization with synthetic dataset are summarized in~\figurename~\ref{app_fig:toy_example_depth_width}. The 200-4 FCDNN and 128-6 FCDNN contain 81K and 66K parameters, respectively. The deeper networks, 128-6 and 128-8 FCDNNs, are more robust to weight quantization. However, the wider one, 200-4 FCDNN is more resilient to activation quantization than the deeper networks. As discussed in Section 3 of the main contents, increasing the width alleviates the effects of weight and activation quantization errors, whereas increasing the depth only reduces the effects of weight quantization error. 

\begin{figure}[h]
    \centering
    \subfigure[128-5 (W2)]{
        \includegraphics[width=.2\linewidth]{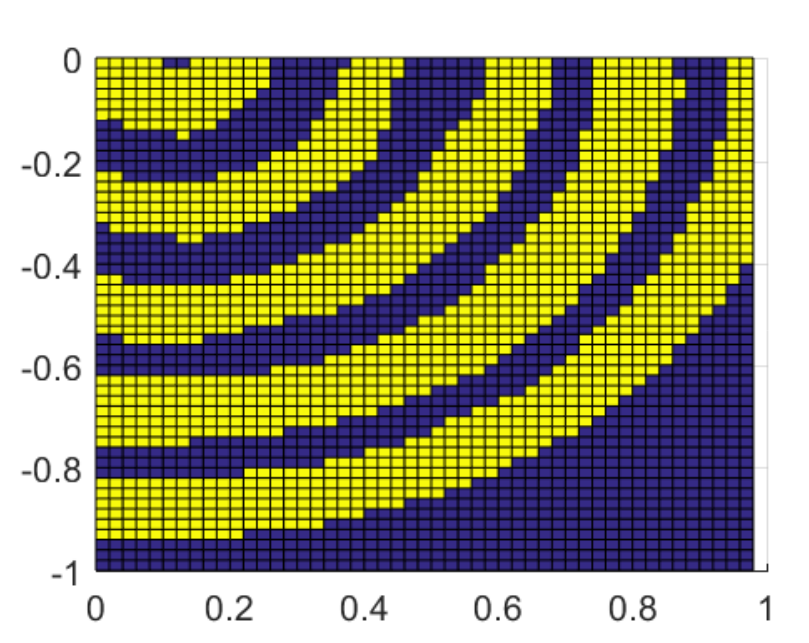}
        }
    \subfigure[128-6 (W2)]{
        \includegraphics[width=.2\linewidth]{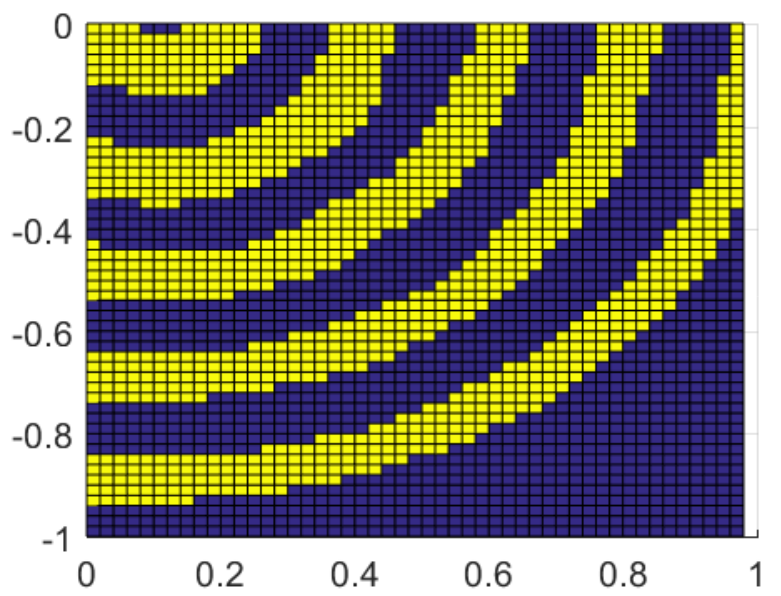}
        }
    \subfigure[128-8 (W2)]{
        \includegraphics[width=.2\linewidth]{images2/128_8_res_w_2bit.pdf}
        }
    \\
    \subfigure[128-5 (A2)]{
        \includegraphics[width=.2\linewidth]{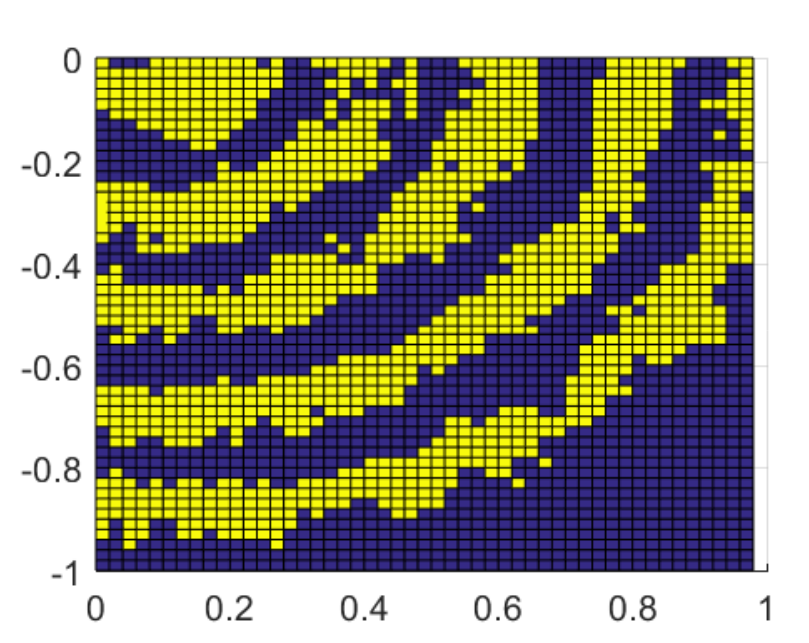}
        }
    \subfigure[128-6 (A2)]{
        \includegraphics[width=.2\linewidth]{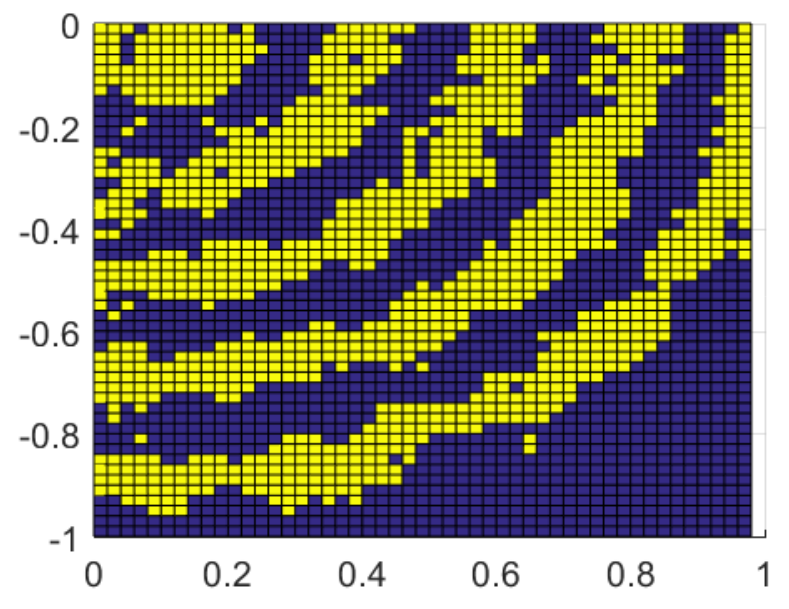}
        }
    \subfigure[128-8 (A2)]{
        \includegraphics[width=.2\linewidth]{images2/128_8_res_a_2bit.pdf}
        }
    \\
    \subfigure[128-5 (W2A2)]{
        \includegraphics[width=.2\linewidth]{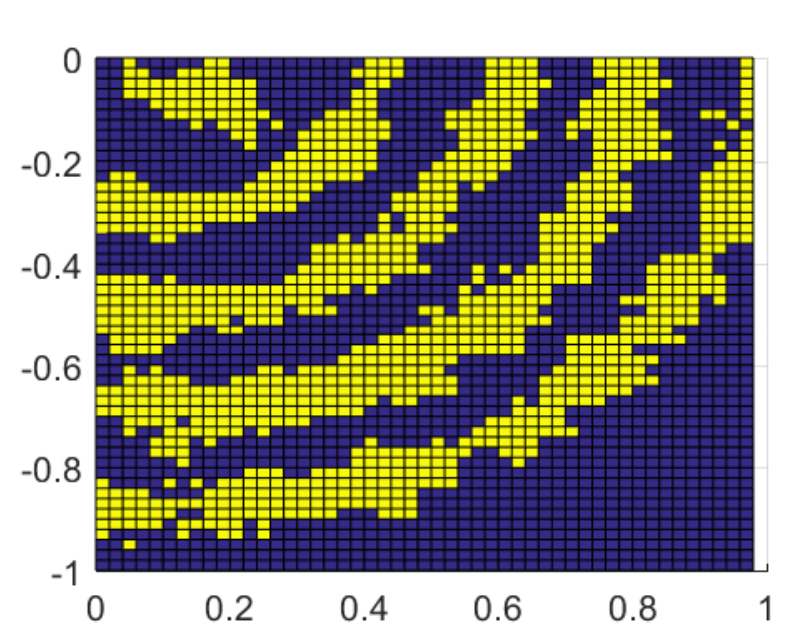}
        }
    \subfigure[128-6 (W2A2)]{
        \includegraphics[width=.2\linewidth]{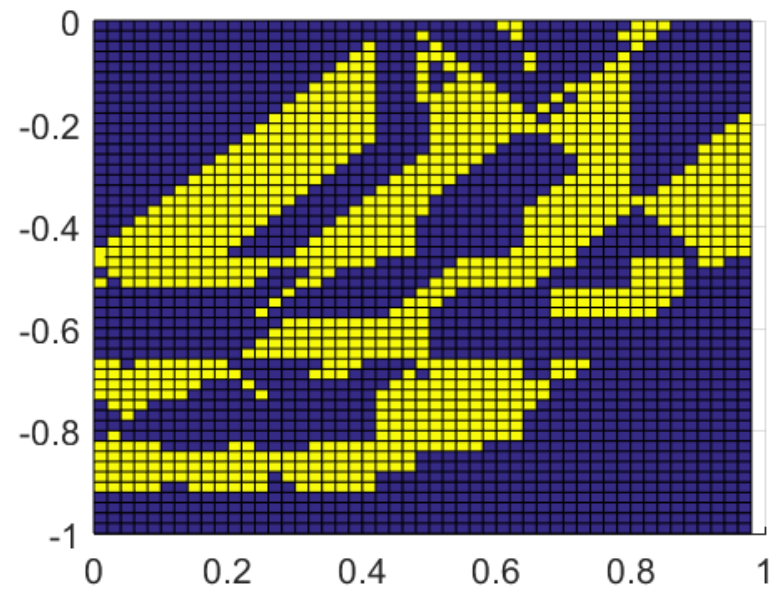}
        }
    \subfigure[128-8 (W2A2)]{
        \includegraphics[width=.2\linewidth]{images2/128_8_res_w_2_a_2bit.pdf}
        }
    %
    \caption{Synthetic data visualization results on FCDNNs with quantized residual connections.}
    \label{app_fig:toy_example_residual}
    \vskip -0.3in
\end{figure}

The visualization results with the residual connections are shown in~\figurename~\ref{app_fig:toy_example_residual}. As discussed in the main contents, increasing the depth with residual connections rather amplifies the activation quantization error when the shortcut outputs are quantized.

\clearpage

\subsection*{B. Details on Cyclic Learning Rate Scheduling}

\begin{figure}[h]
\centering
\vskip -0.2in
\tikzset{mark options={mark size=1}}
\begin{tikzpicture}
    \begin{axis}[
	width=0.8\linewidth,
	height = 0.25\linewidth,
	compat=1.12,
	xmin=0,
	xmax=600,
	ymode=log,
	xticklabels=\empty,
    xlabel=Iterations,
    ylabel=LR scale factor,
	xlabel shift=-3pt,
	ylabel shift=-3pt,
    tick label style={font=\footnotesize},
    label style={font=\footnotesize}
    ]]
	\addplot[color=black] file{data/clr.txt}; 
	\draw[color=red,dashed] (160, 0.008) rectangle (319, 1.2);
    \end{axis}
   \end{tikzpicture}
   \caption{Learning rate scale factor along training iterations. The red dashed box indicates a single cycle of CLR scheduling.}
   \label{app_fig:clr}
   \end{figure}
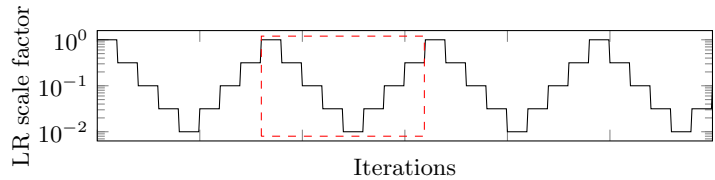

The cyclic learning rate (CLR) scheduling is shown in~\figurename~\ref{app_fig:clr}. The scale factor is multiplied to the base LR and changes 8 times in a single cycle; $\{1.0$, $\sqrt{0.1}$, $0.1$, $0.1\sqrt{0.1}$, $0.01$, 0$.1\sqrt{0.1}$, $0.1$, $\sqrt{0.1}$ $\}$. The cycle period and base LR for each dataset are as follows:

\noindent \textbf{CIFAR-10:} The base LR is 1e-4, which is 10 times larger than the last LR of the retraining. The cycle period is 8 epochs, thus the LR scaling factor changes at every epoch. 


\noindent \textbf{PASCAL VOC segmentation:} The last LR of the retraining becomes 0 with the polynomial policy for this task. We set the base LR to 1e-5, which is 10 times smaller than the initial LR of the retraining. The cycle period is 3K iterations.

\subsection*{C. Additional Visualization Examples on the PASCAL VOC Segmentation Benchmark}

\begin{figure}[h]
    \centering
    \subfigure{
        \includegraphics[width=.17\linewidth]{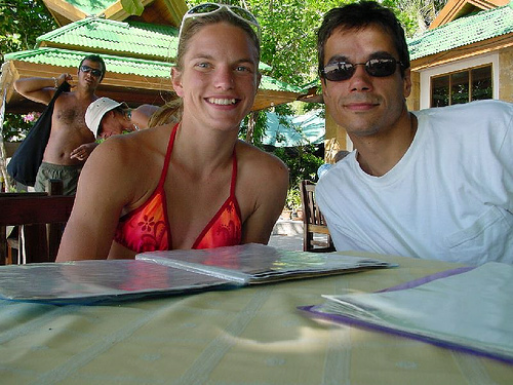}
    }
    \subfigure{
        \includegraphics[width=.17\linewidth]{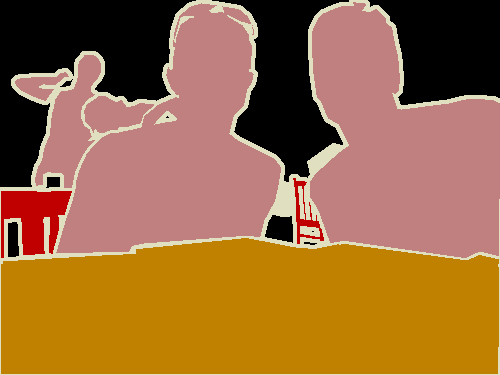}
    }
    \subfigure{
        \includegraphics[width=.17\linewidth]{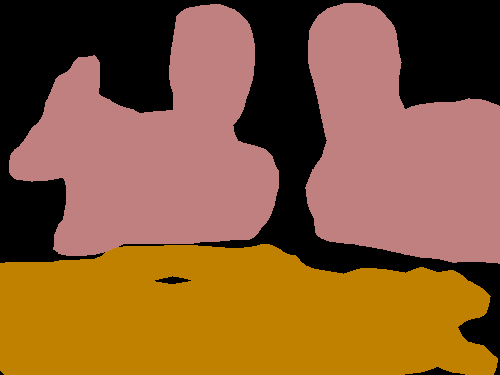}
    }
    \subfigure{
        \includegraphics[width=.17\linewidth]{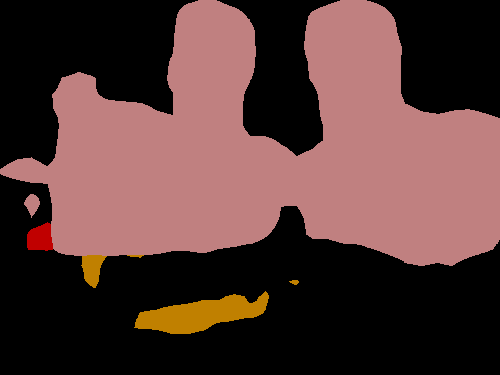}
    }
    \subfigure{
        \includegraphics[width=.17\linewidth]{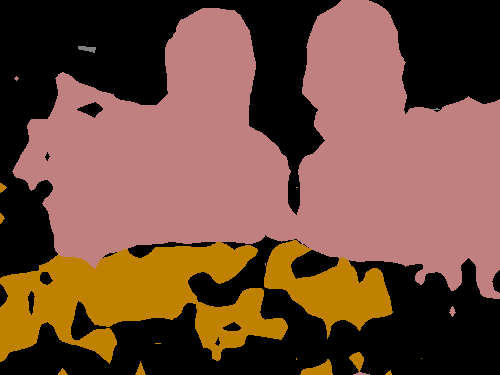}
    }
    \vskip 0.03in
    \subfigure{
        \includegraphics[width=.17\linewidth]{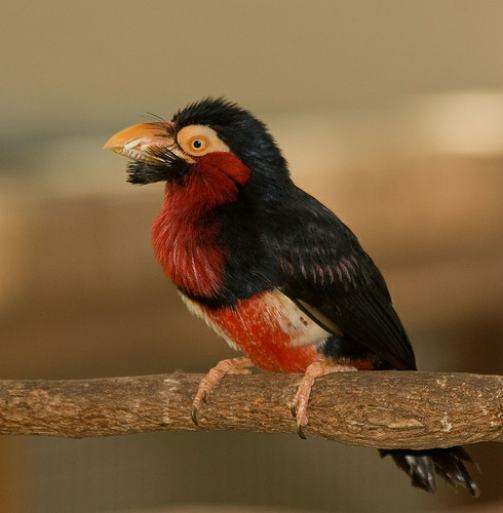}
    }
    \subfigure{
        \includegraphics[width=.17\linewidth]{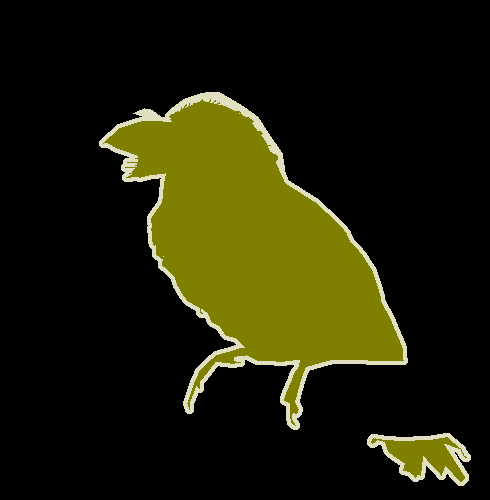}
    }
    \subfigure{
        \includegraphics[width=.17\linewidth]{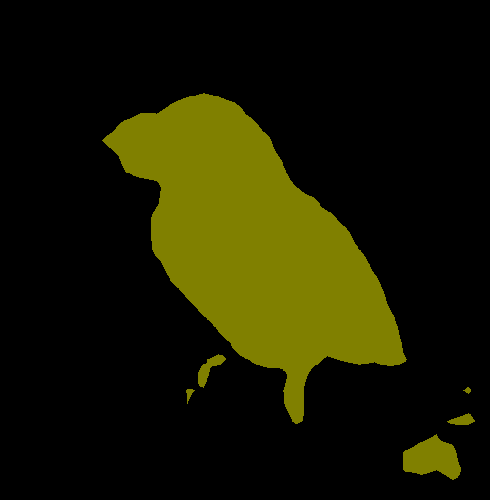}
    }
    \subfigure{
        \includegraphics[width=.17\linewidth]{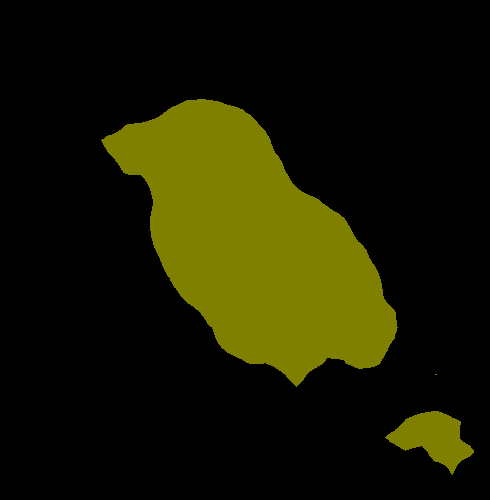}
    }
    \subfigure{
        \includegraphics[width=.17\linewidth]{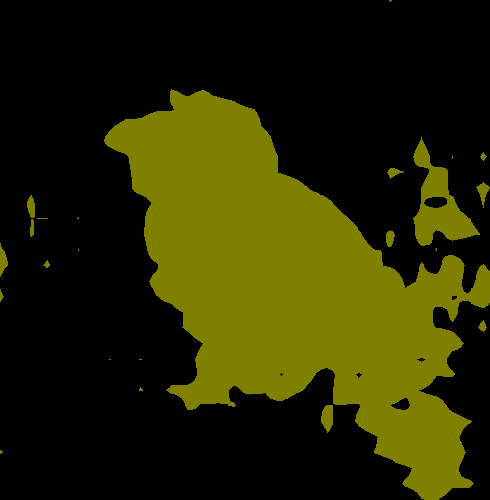}
    }
   \vskip 0.03in
    \subfigure{
        \includegraphics[width=.17\linewidth]{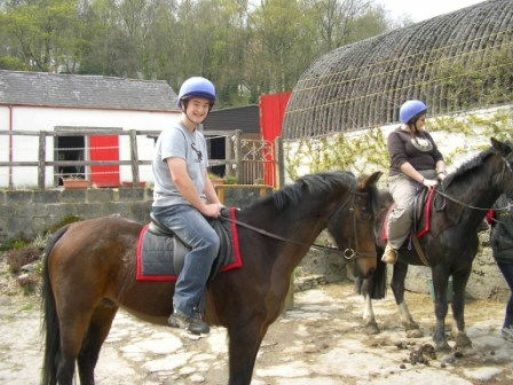}
    }
    \subfigure{
        \includegraphics[width=.17\linewidth]{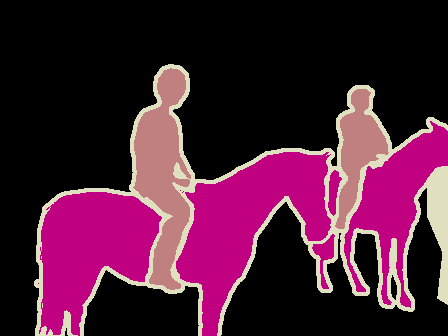}
    }
    \subfigure{
        \includegraphics[width=.17\linewidth]{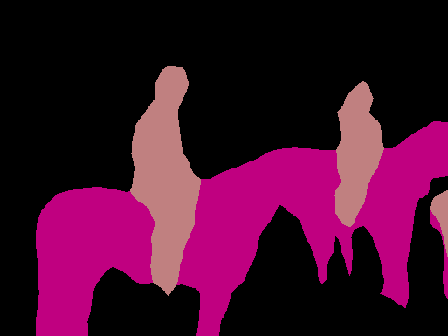}
    }
    \subfigure{
        \includegraphics[width=.17\linewidth]{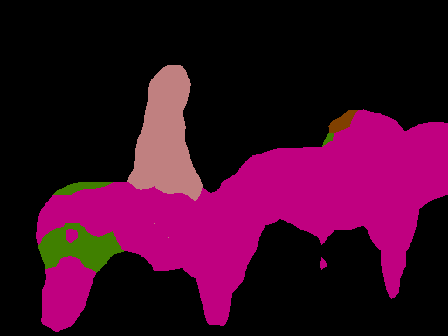}
    }
    \subfigure{
        \includegraphics[width=.17\linewidth]{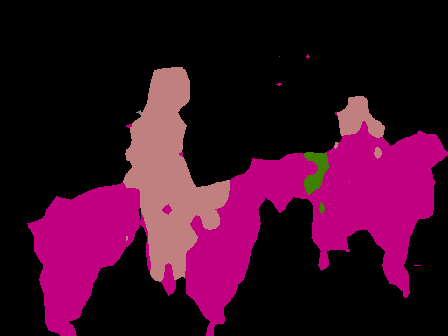}
    }
    \vskip 0.03in
    \subfigure{
        \includegraphics[width=.17\linewidth]{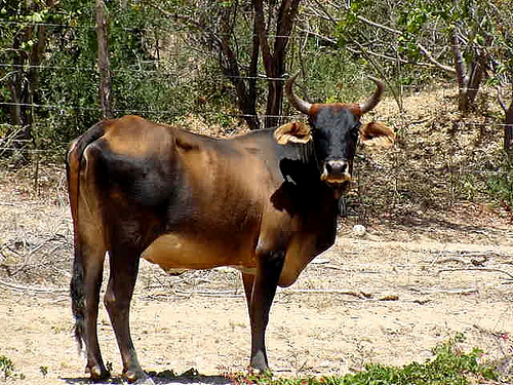}
    }
    \subfigure{
        \includegraphics[width=.17\linewidth]{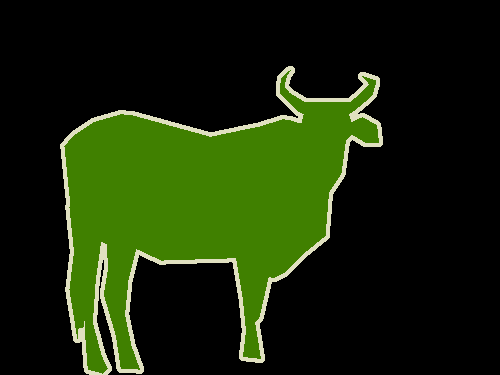}
    }
    \subfigure{
        \includegraphics[width=.17\linewidth]{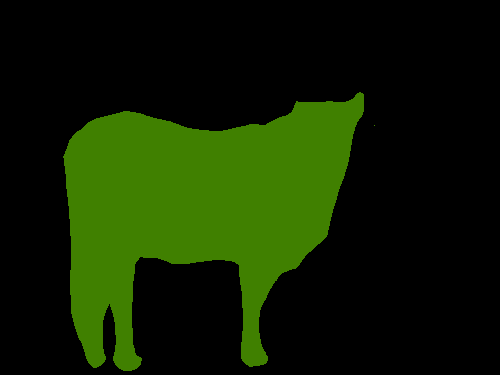}
    }
    \subfigure{
        \includegraphics[width=.17\linewidth]{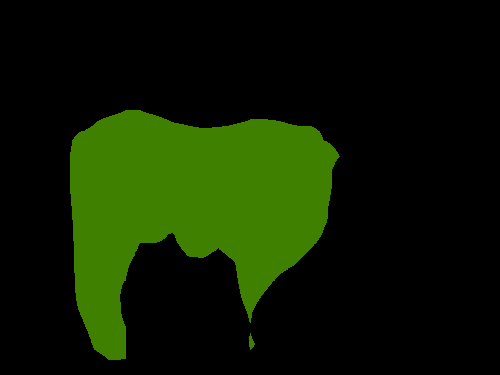}
    }
    \subfigure{
        \includegraphics[width=.17\linewidth]{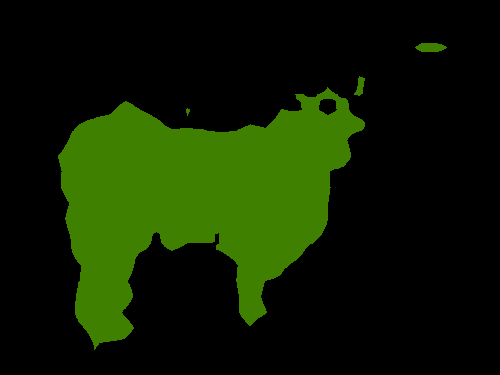}
    }
    \vskip 0.03in
    \subfigure{
        \includegraphics[width=.17\linewidth]{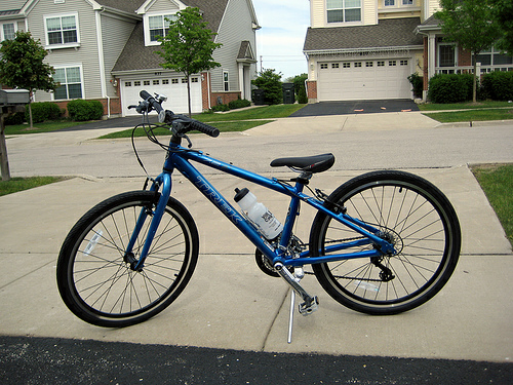}
    }
    \subfigure{
        \includegraphics[width=.17\linewidth]{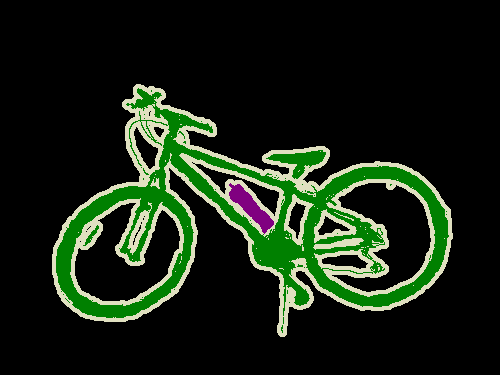}
    }
    \subfigure{
        \includegraphics[width=.17\linewidth]{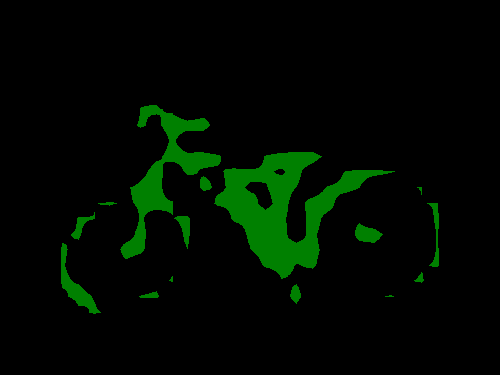}
    }
    \subfigure{
        \includegraphics[width=.17\linewidth]{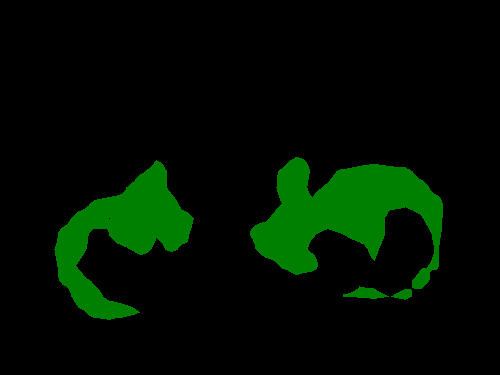}
    }
    \subfigure{
        \includegraphics[width=.17\linewidth]{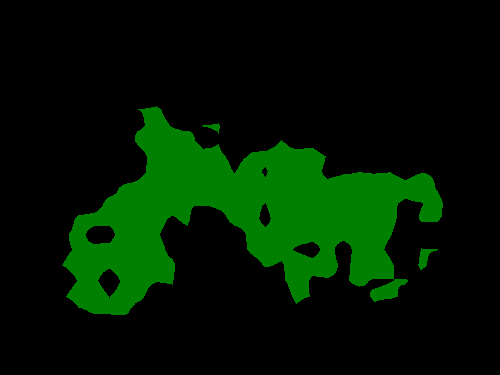}
    }
    \vskip 0.03in
    \subfigure{
        \includegraphics[width=.17\linewidth]{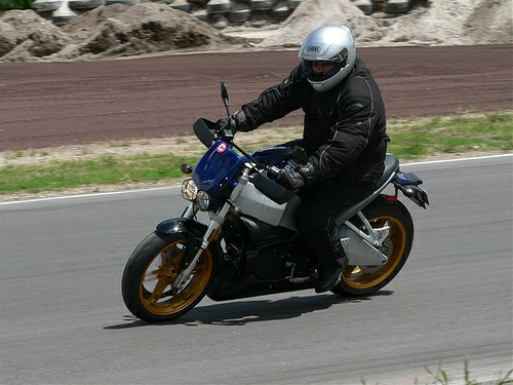}
    }
    \subfigure{
        \includegraphics[width=.17\linewidth]{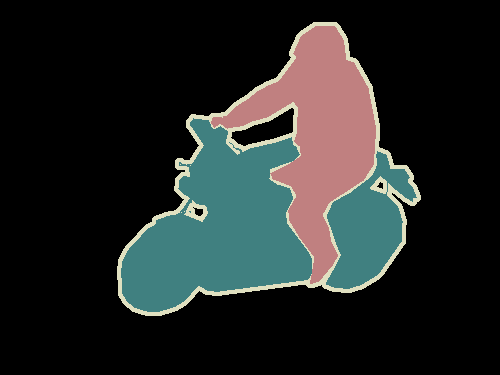}
    }
    \subfigure{
        \includegraphics[width=.17\linewidth]{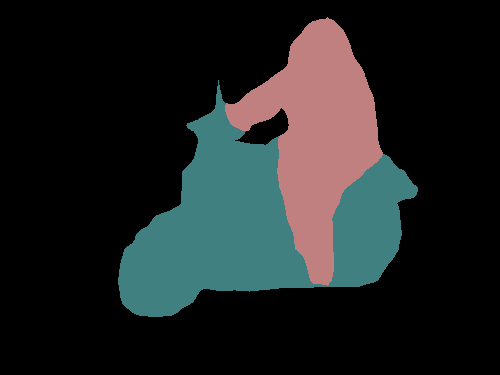}
    }
    \subfigure{
        \includegraphics[width=.17\linewidth]{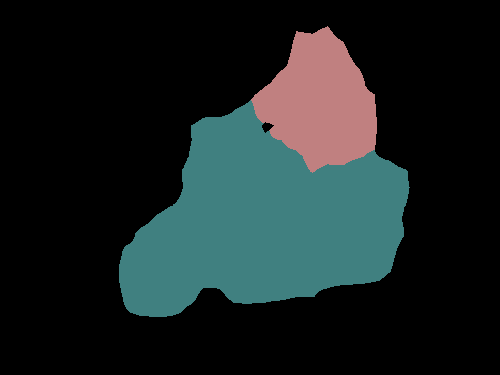}
    }
    \subfigure{
        \includegraphics[width=.17\linewidth]{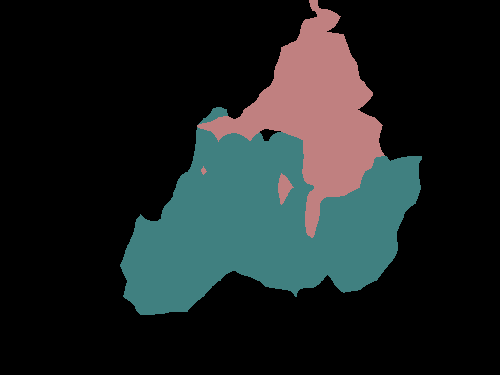}
    }
    \caption{Visualization examples on the PASCAL VOC segmentation validation set. The first and second columns represent the input images and the ground truth labels. Third to fifth columns show the visualization results of the floating-point, 2-bit weight quantized, and 2-bit activation quantized models, respectively.}
    \label{app_fig:segmentation}
\end{figure}
\clearpage

\subsection*{D. Details of Experimented Model Structures}

\begin{table}[h]
\centering
\caption{Quantization performance of various model structures on CIFAR-10 testset. `$L$', `$D_{init}$', and `$B$' represent the number of layers, initial channel dimension, and number of blocks, respectively.}
\setlength\tabcolsep{5pt}
\begin{tabular}{ccccc|cccc}
\toprule
\multicolumn{5}{c|}{Model}                                                                                                       & \multicolumn{4}{c}{Test Acc(\%)} \\
Block type                                                                      & $L$ & $D_{init}$ &  $B$ &  Params. & Float   & W2AF   & WFA2  & W2A2  \\ \midrule
\multirow{6}{*}{Basic block}                                                    & 8        & 59         & $1\times3$         & 1.05 M    & 93.26  & 92.83  & 92.58  & 92.32 \\
                                                                                & 14       & 40         & $2\times3$         & 1.10 M    & 93.62  & 93.51  & \textbf{93.30}  & \textbf{93.11} \\
                                                                                & 20       & 31         & $3\times3$         & 1.03 M    & 94.14  & 93.57  & 92.94  & 92.65 \\
                                                                                & 32       & 24         & $5\times3$         & 1.06 M    & 94.24  & 93.71  & 92.42  & 91.84 \\
                                                                                & 44       & 20         & $7\times3$         & 1.04 M    & \textbf{94.27}  & \textbf{93.91}  & 91.40  & 90.69 \\
                                                                                & 56       & 18         & $9\times3$         & 1.09 M    & 93.58  & 93.46  & 90.58  & 89.77 \\ \midrule
\multirow{6}{*}{\begin{tabular}[c]{@{}c@{}}Pre-activation\\ block\end{tabular}} & 8        & 59         & $1\times3$         & 1.05 M    & 93.10  & 92.75  & 91.89  & 91.03 \\
                                                                                & 14       & 40         & $2\times3$         & 1.10 M    & 94.14  & 93.85  & \textbf{92.44}  & \textbf{91.81} \\
                                                                                & 20       & 31         & $3\times3$         & 1.03 M    & 94.25  & 93.82  & 92.42  & \textbf{91.81} \\
                                                                                & 32       & 24         & $5\times3$         & 1.06 M    & 94.07  & 93.89  & 91.51  & 89.03 \\
                                                                                & 44       & 20         & $7\times3$         & 1.04 M    & \textbf{94.31}  & \textbf{94.04}  & 90.58  & 86.26 \\
                                                                                & 56       & 18         & $9\times3$         & 1.09 M    & 93.96  & 93.83  & 88.45  & 85.45 \\ \midrule
\multirow{5}{*}{\begin{tabular}[c]{@{}c@{}}Depthwise\\ block\end{tabular}}      & 29       & 38         & $3\times3$         & 1.11 M    & 93.64  & 93.28  & \textbf{93.38}  & 92.33 \\
                                                                                & 38       & 32         & $4\times3$         & 1.08 M    & 93.65  & 93.47  & 92.93  & \textbf{92.57} \\
                                                                                & 56       & 26         & $6\times3$         & 1.11 M    & 94.27  & 94.02  & 91.70  & 90.61 \\
                                                                                & 74       & 22         & $8\times3$         & 1.08 M    & 93.94  & 93.71  & 90.50  & 89.88 \\
                                                                                & 110      & 18         & $12\times3$         & 1.13 M    & \textbf{94.43}  & \textbf{94.39}  & 88.75  & 87.47 \\ \bottomrule
\end{tabular}
\label{app_table:structure}
\end{table}

The details of the model structure for Fig. 8 in the main contents are shown in Table~\ref{app_table:structure}. The number of parameters in all the experimented models is approximately one million. All CNNs consist of three block groups. `$B$' indicates the block structure; (number of blocks in each group) $\times$ (number of groups). The first block in second and third block groups have a stride of 2. ‘$D$’ represents the channel width of the first layer and the channel width increases by 2 times at the beginning of second and third block groups. Floating-point and weight quantized networks perform better on deeper models, while shallow and wider models shows better performance when activations are quantized.

\end{document}


\title{Supplementary Materials for Submission 7556}

\author{}

\maketitle

\section*{A. More visualization examples on the synthetized dataset} \nonumber

\begin{figure}[h]
    \centering
    
    \subfigure[200-4 (W2)] {\includegraphics[width=0.19\linewidth]{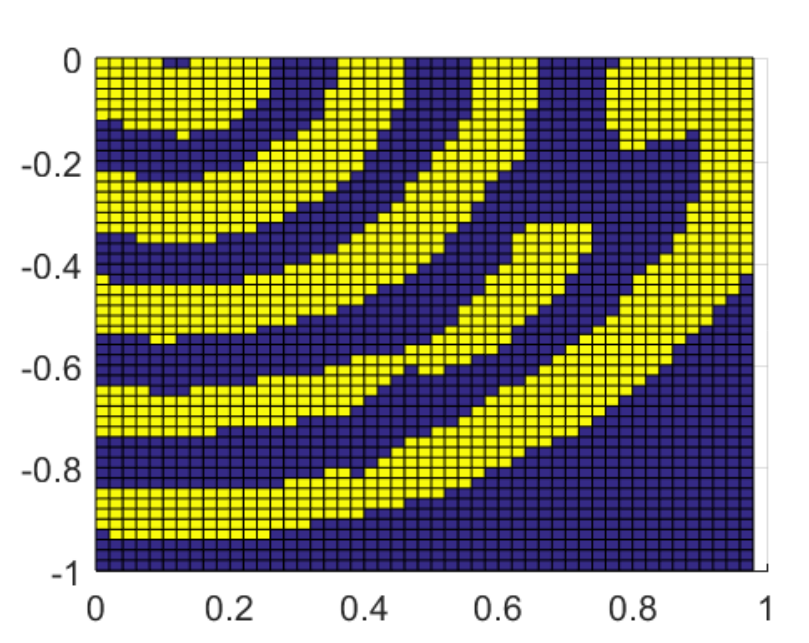}}
    \subfigure[384-4 (W2)] {\includegraphics[width=0.19\linewidth]{images_appendix/pred_300_4_w_2.pdf}}
    \subfigure[128-5 (W2)] {\includegraphics[width=0.19\linewidth]{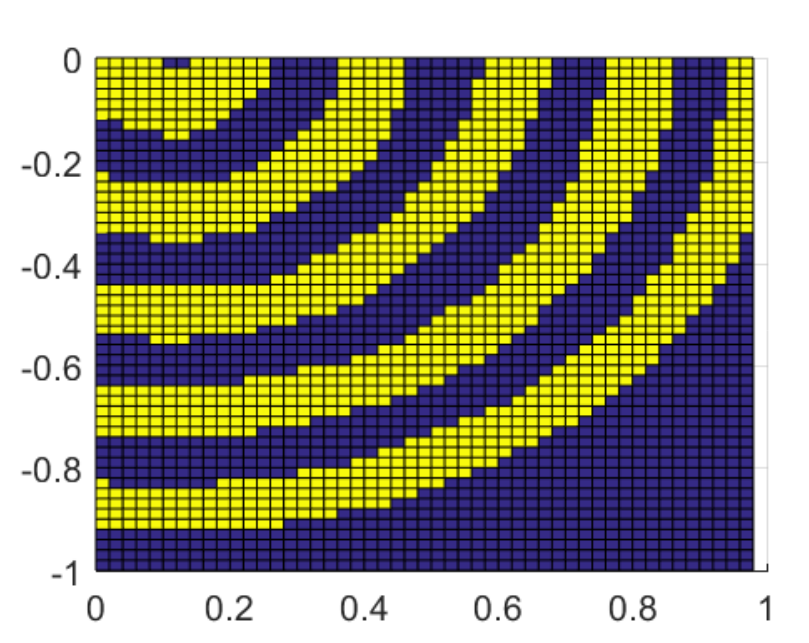}}
    \subfigure[128-6 (W2)] {\includegraphics[width=0.185\linewidth]{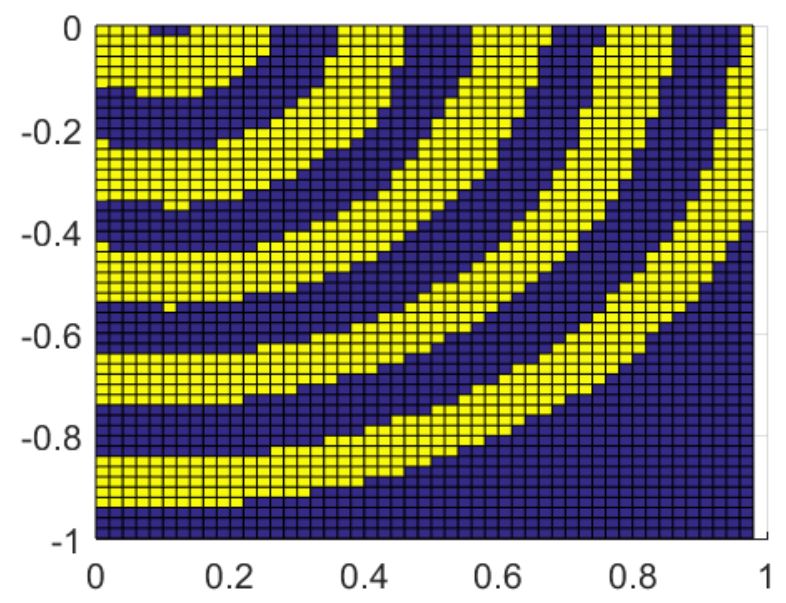}}
    \subfigure[128-8 (W2)] {\includegraphics[width=0.18\linewidth]{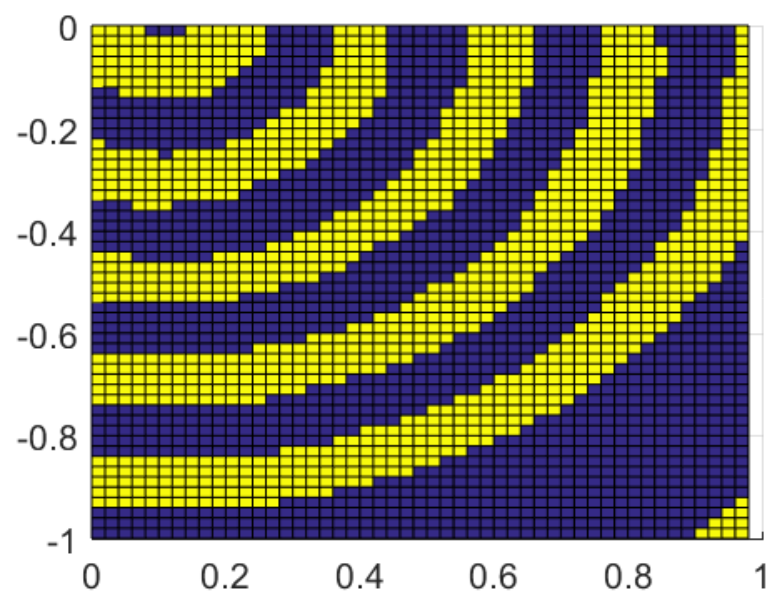}}
    
    \subfigure[200-4 (A2)] {\includegraphics[width=0.19\linewidth]{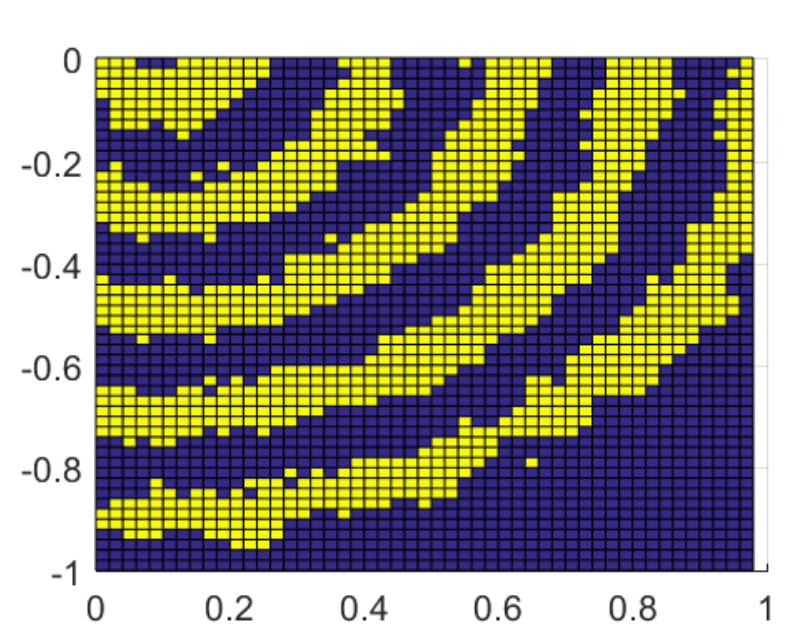}}
    \subfigure[384-4 (A2)] {\includegraphics[width=0.19\linewidth]{images2/384_4_w_32_a_2bits.pdf}}
    \subfigure[128-5 (A2)] {\includegraphics[width=0.19\linewidth]{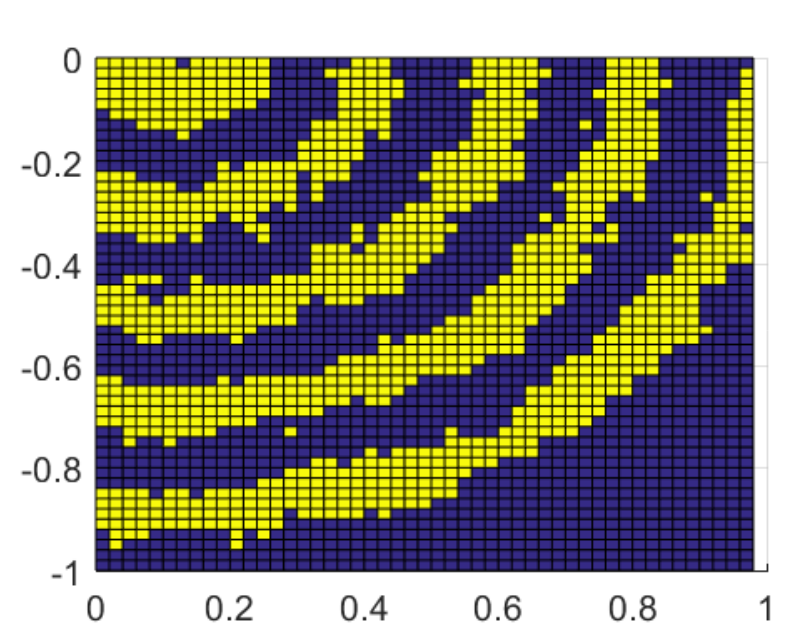}}
    \subfigure[128-6 (A2)] {\includegraphics[width=0.185\linewidth]{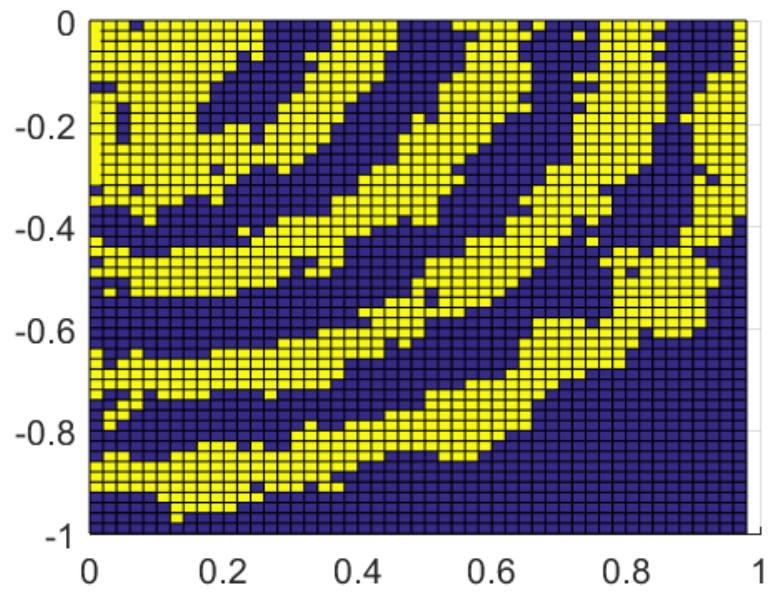}}
    \subfigure[128-8 (A2)] {\includegraphics[width=0.185\linewidth]{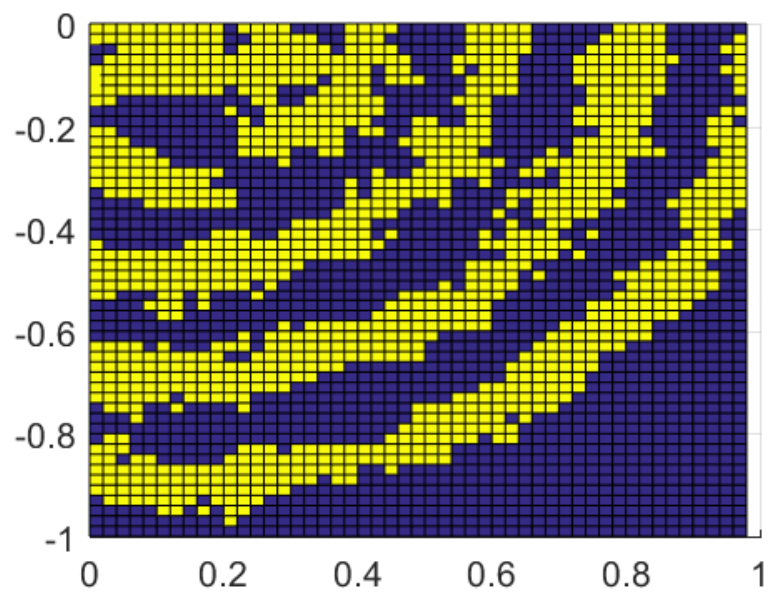}}
    
    \subfigure[200-4 (W2A2)] {\includegraphics[width=0.19\linewidth]{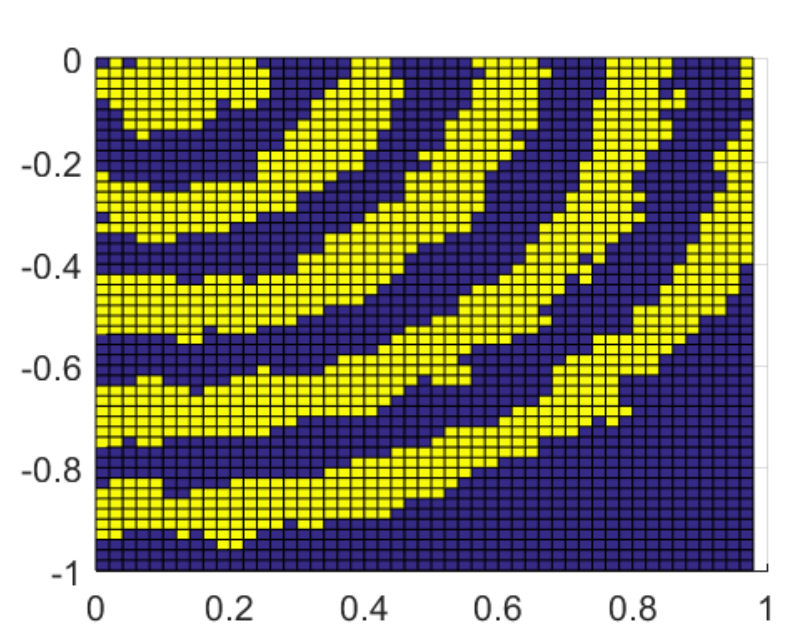}}
    \subfigure[384-4 (W2A2)] {\includegraphics[width=0.19\linewidth]{images2/384_4_w_2_a_2bits.pdf}}
    \subfigure[128-5 (W2A2)] {\includegraphics[width=0.19\linewidth]{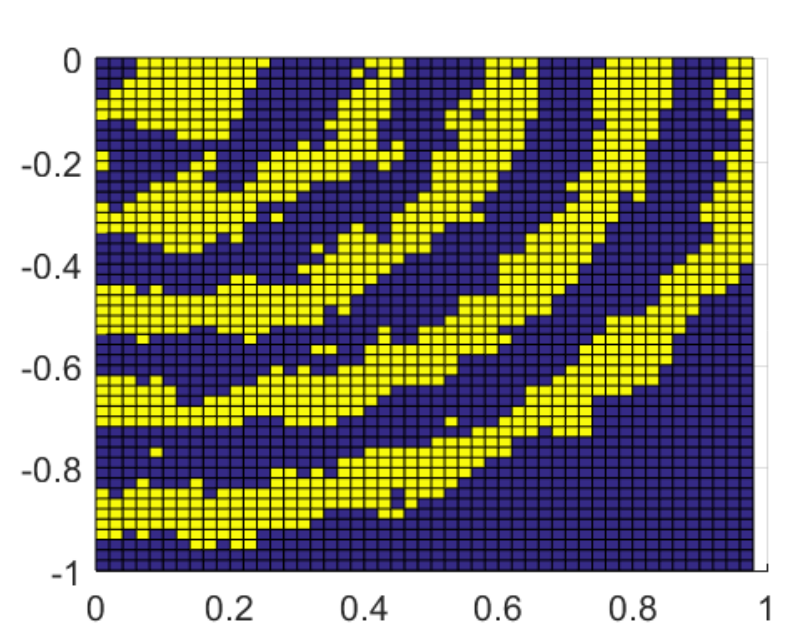}}
    \subfigure[128-6 (W2A2)] {\includegraphics[width=0.185\linewidth]{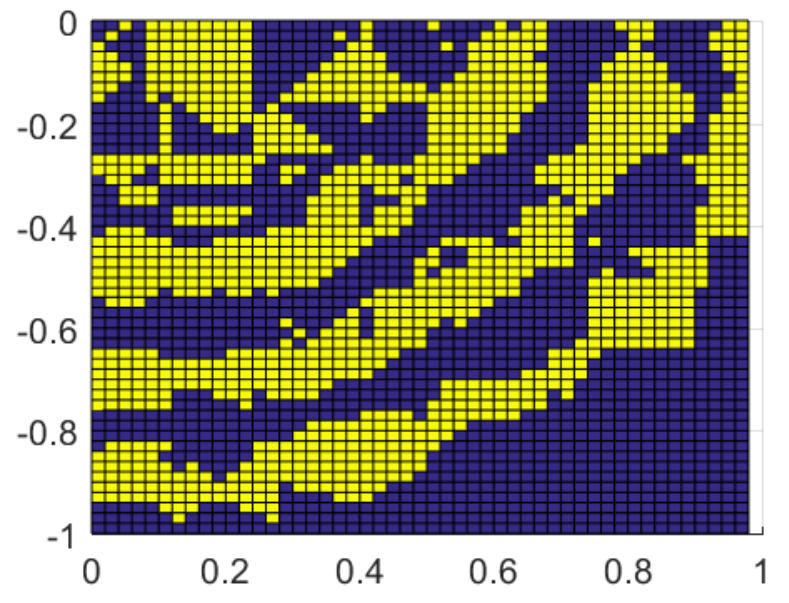}}
    \subfigure[128-8 (W2A2)] {\includegraphics[width=0.185\linewidth]{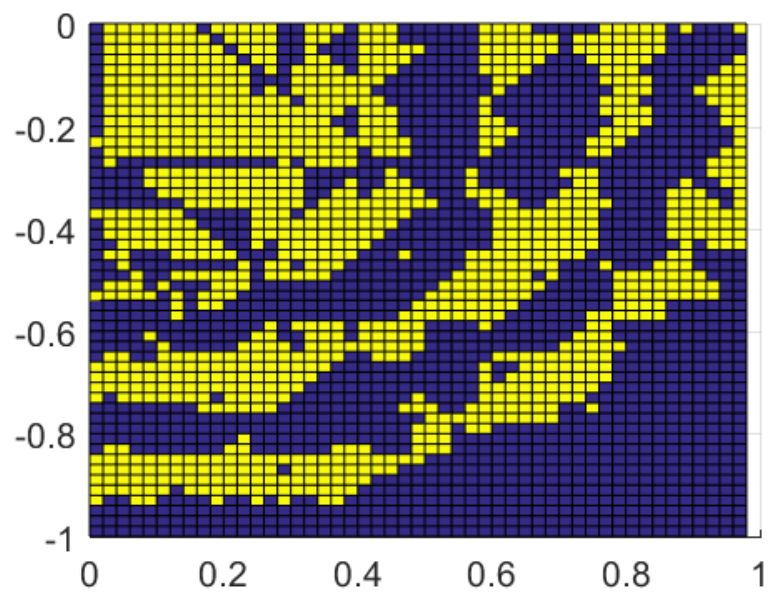}}
    
    \caption{Synthetic data visualization results on FCDNNs as the width and depth increase. The experimented FCDNNs are denoted as `$D$'-`$H$', where $D$ is the dimension of hidden layers and $H$ is the number of layers. W2 and A2 represent the 2-bit weight and 2-bit activation quantization, respectively.}
    \label{app_fig:toy_example_depth_width}
\end{figure}

More examples of visualization with synthetic dataset are shown in Figure~\ref{app_fig:toy_example_depth_width}. 200-4 FCDNN and 128-6 FCDNN contain 81K and 66K parameters, respectively. The deeper networks, 128-6 and 128-8 FCDNNs, are more robust to weight quantization. However, the wider one, 200-4 FCDNN is more resilient to activation quantization. As discussed in the main part of this paper, increasing the width alleviates the effects of weight and activation quantization errors, whereas increasing the depth only reduces the effects of weight quantization error. 


The visualization results with the residual connections are shown in Figure~\ref{app_fig:toy_example_residual}. As discussed in the main content, increasing the depth with residual connections rather amplifies the activation quantization error.


\begin{figure}[h]
    \centering
    \subfigure[128-5 Residual (W2)]
    {\includegraphics[width=0.23\linewidth]{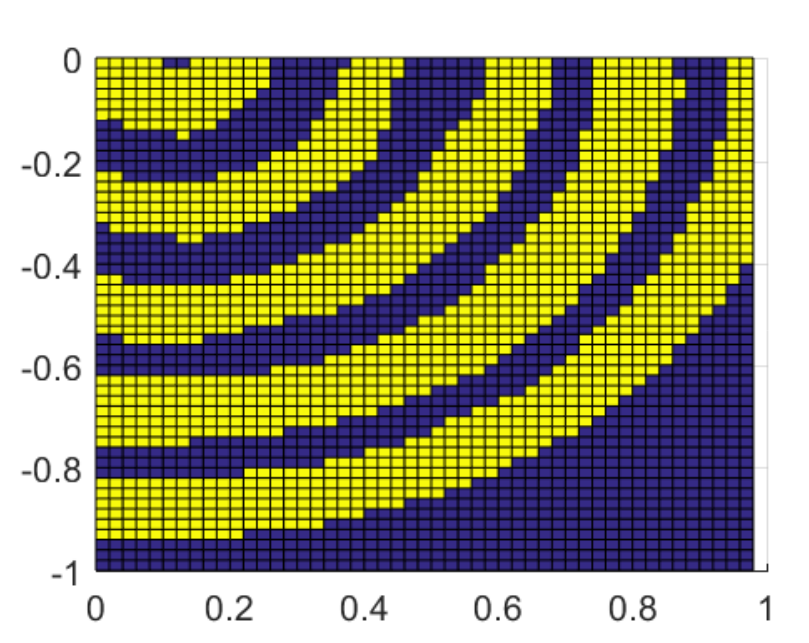}}
    \subfigure[128-6 Residual (W2)]
    {\includegraphics[width=0.23\linewidth]{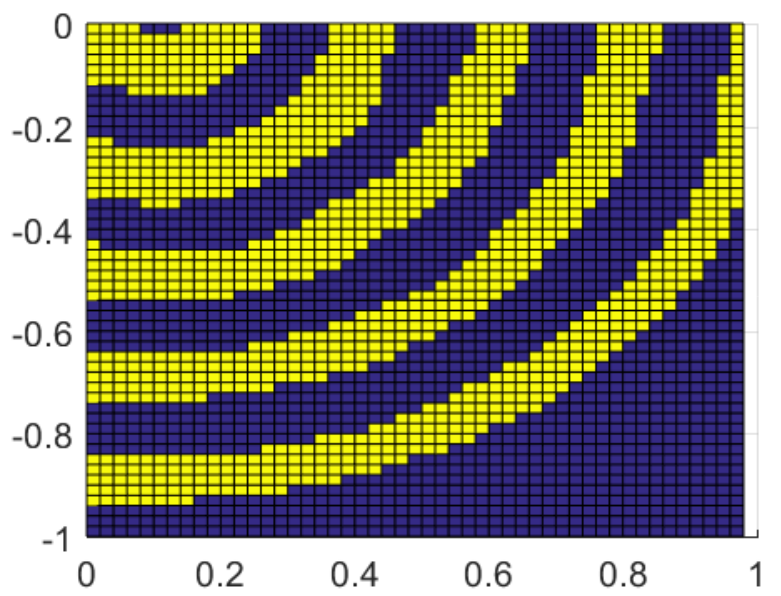}}
    \subfigure[128-8 Residual (W2)]
    {\includegraphics[width=0.23\linewidth]{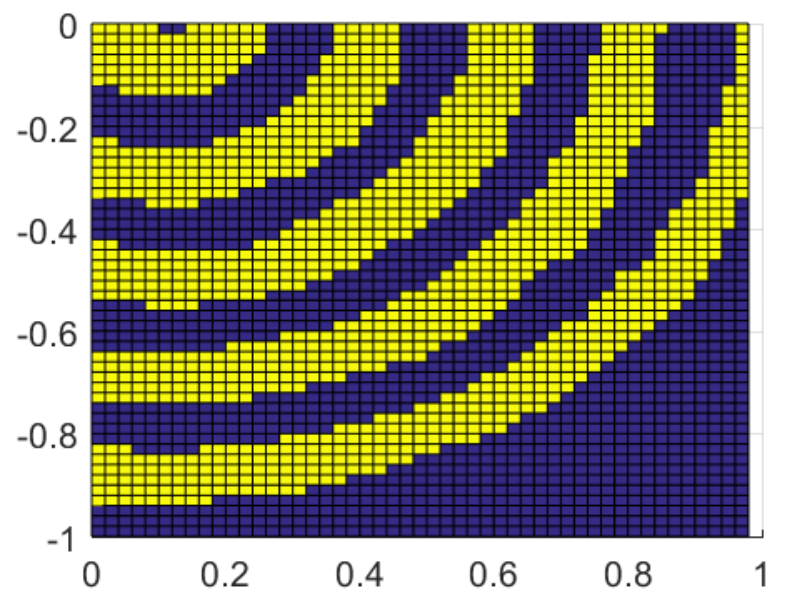}}

    \subfigure[128-5 Residual (A2)]
    {\includegraphics[width=0.23\linewidth]{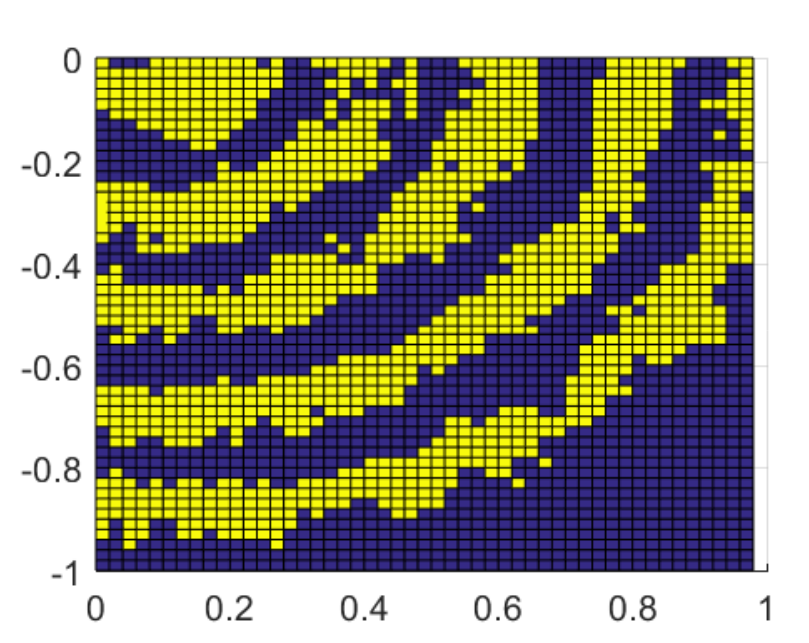}}
    \subfigure[128-6 Residual (A2)]
    {\includegraphics[width=0.23\linewidth]{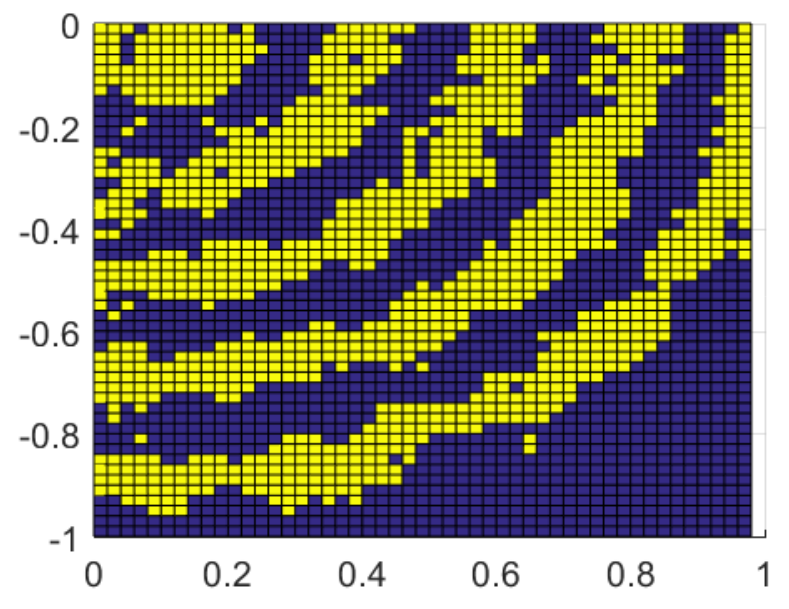}}
    \subfigure[128-8 Residual (A2)]
    {\includegraphics[width=0.23\linewidth]{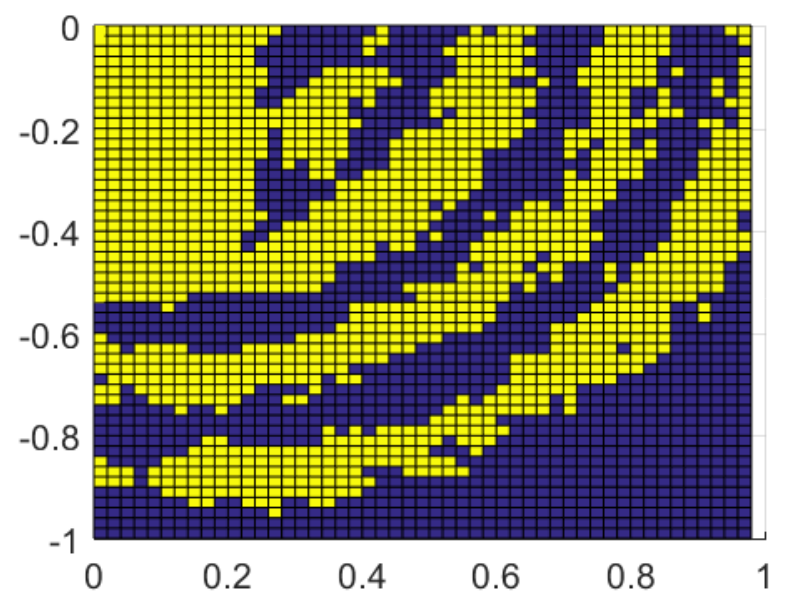}}

    \subfigure[128-5 Residual (W2A2)]
    {\includegraphics[width=0.23\linewidth]{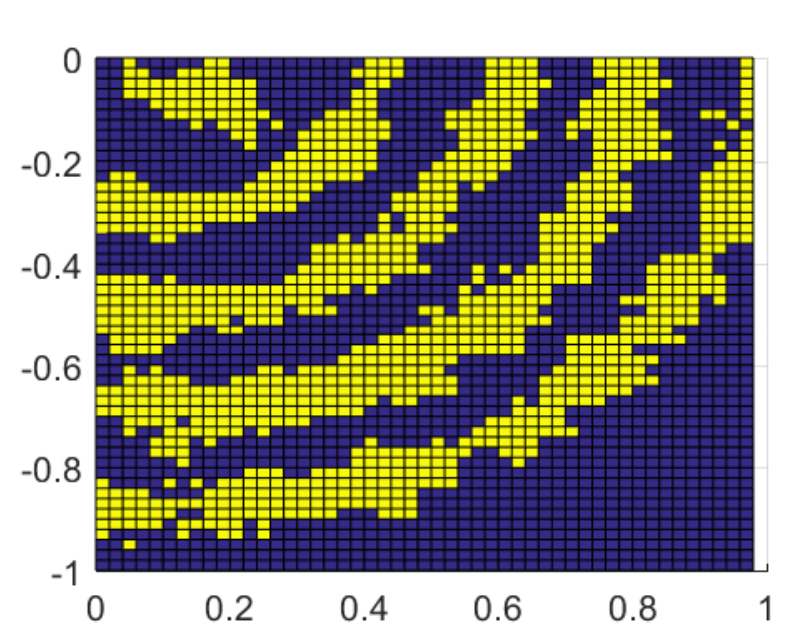}}
    \subfigure[128-6 Residual (W2A2)]
    {\includegraphics[width=0.23\linewidth]{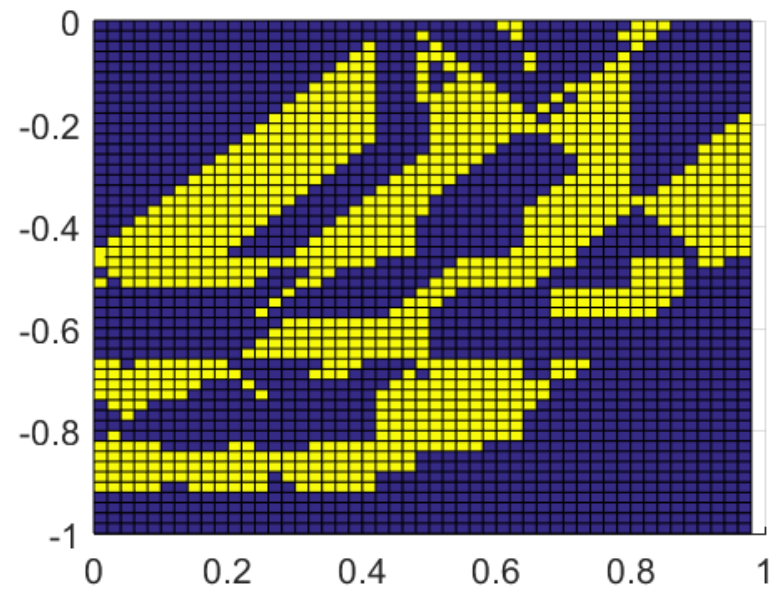}}
    \subfigure[128-8 Residual (W2A2)]
    {\includegraphics[width=0.23\linewidth]{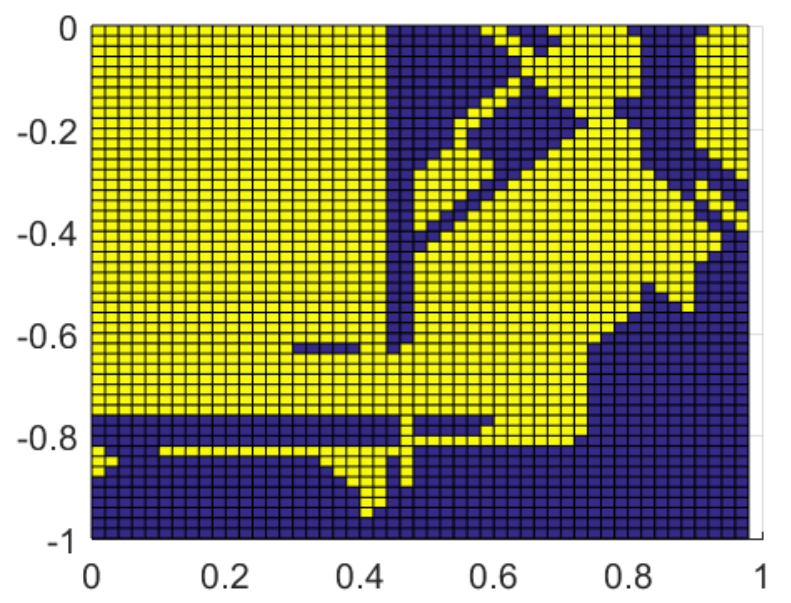}}
    
    \vskip 0.1in
    \caption{Synthetic data visualization results on FCDNNs with residual connections.}
    \label{app_fig:toy_example_residual}
\end{figure}


\section*{B. Cyclic learning rate scheduling}

\begin{figure}[h]
\centering
\tikzset{mark options={mark size=1}}
\begin{tikzpicture}
    \begin{axis}[
	width=0.8\linewidth,
	height = 0.25\linewidth,
	compat=1.12,
	xmin=0,
	xmax=600,
	ymode=log,
	xticklabels=\empty,
    xlabel=Iterations,
    ylabel=LR scale factor,
	xlabel shift=-3pt,
	ylabel shift=-3pt,
    tick label style={font=\footnotesize},
    label style={font=\footnotesize}
    ]]
	
	\addplot[color=black] file{data/clr.txt}; 
	\draw[color=red,dashed] (160, 0.008) rectangle (319, 1.2);
    \end{axis}
   \end{tikzpicture}
   \caption{Learning rate scale factor along training iterations. The red dashed box indicates a single cycle of CLR scheduling.}
   \label{app_fig:clr}
   \end{figure}

The cyclic learning rate (CLR) scheduling is shown in Figure~\ref{app_fig:clr}. The scale factor is multiplied to the base LR and changes 8 times in a single cycle; $\{1.0, \sqrt{0.1}, 0.1, 0.1\sqrt{0.1}, 0.01, 0.1\sqrt{0.1}, 0.1, \sqrt{0.1} \}$. The cycle period and base LR for each dataset are as follows:

\textbf{CIFAR-10:} The base LR is 1e-4, which is 10 times larger than the last LR of the retraining. The cycle period is 8 epochs, thus the LR scaling factor changes at every epoch. 

\textbf{ImageNet:} The base LR is 4e-4 and the cycle period is 8 epochs.

\textbf{PASCAL VOC segmentation:} The last LR of the retraining becomes 0 with the polynomial policy for this task. We set the base LR to 1e-5, which is 10 times smaller than the initial LR of the retraining. The cycle period is 3K iterations.

\section*{C. More visualization examples on the PASCAL VOC segmentation benchmark}

\begin{figure}[h]
    \centering
    \subfiguretopcaptrue
    \subfigure[Image] {\includegraphics[width=0.18\linewidth]{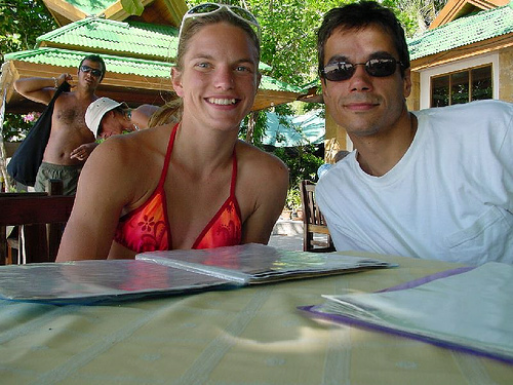}}
    \subfigure[Ground truth] {\includegraphics[width=0.18\linewidth]{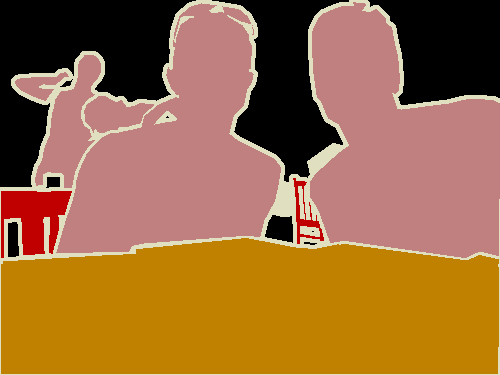}}
    \subfigure[Float] {\includegraphics[width=0.18\linewidth]{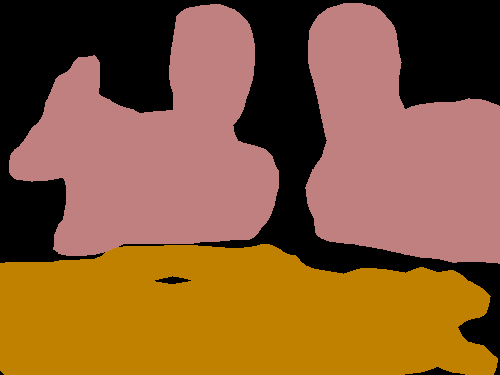}}
    \subfigure[Weight quantized] {\includegraphics[width=0.18\linewidth]{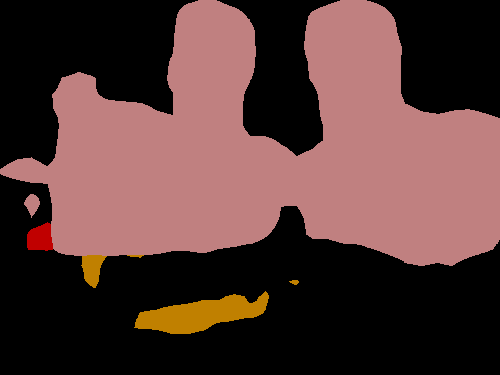}}
    \subfigure[Activation quantized] {\includegraphics[width=0.18\linewidth]{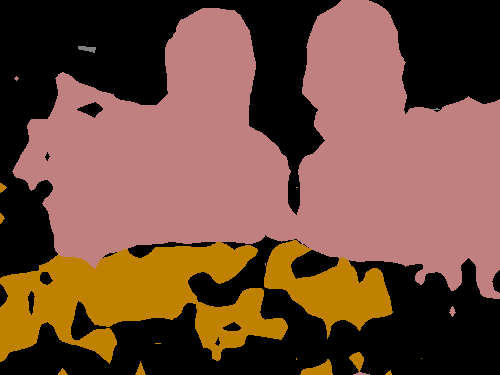}}
   
    \subfigure {\includegraphics[width=0.18\linewidth]{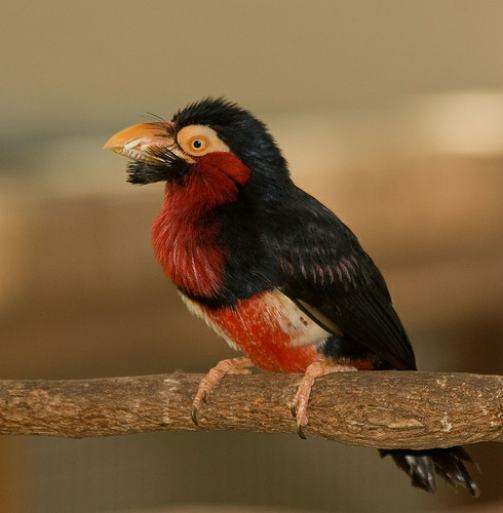}}
    \subfigure {\includegraphics[width=0.18\linewidth]{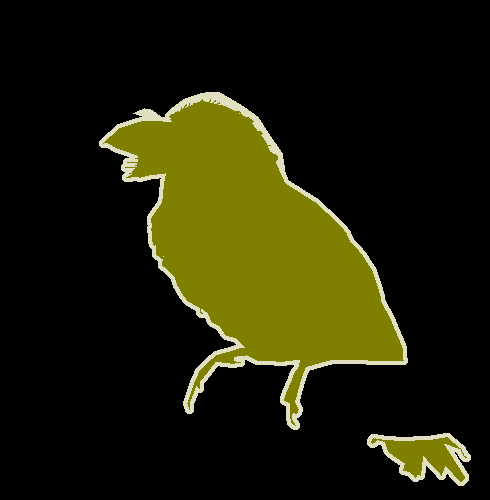}}
    \subfigure {\includegraphics[width=0.18\linewidth]{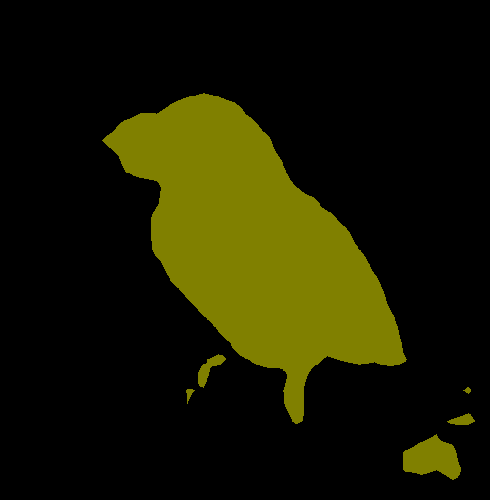}}
    \subfigure {\includegraphics[width=0.18\linewidth]{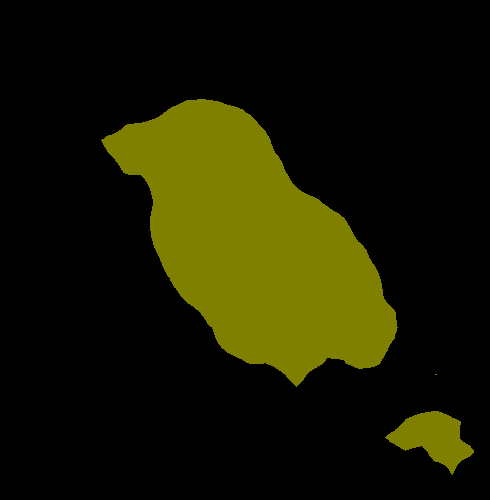}}
    \subfigure {\includegraphics[width=0.18\linewidth]{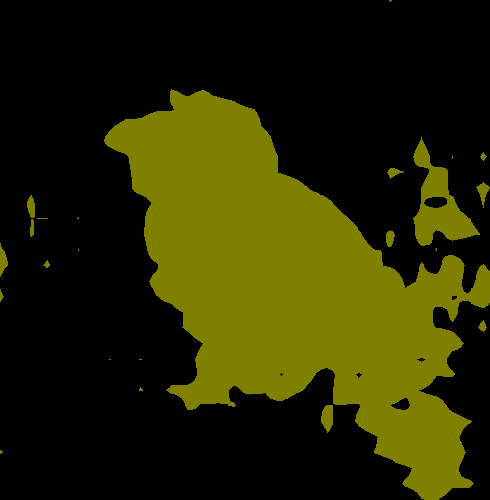}}
    
    \subfigure {\includegraphics[width=0.18\linewidth]{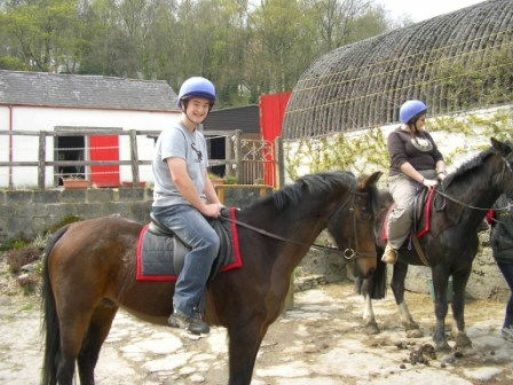}}
    \subfigure {\includegraphics[width=0.18\linewidth]{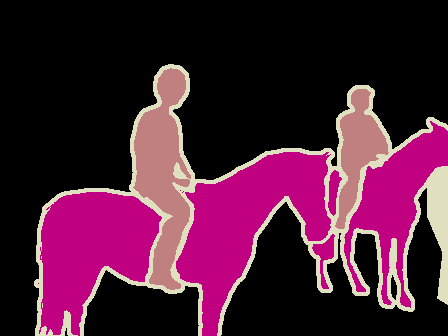}}
    \subfigure {\includegraphics[width=0.18\linewidth]{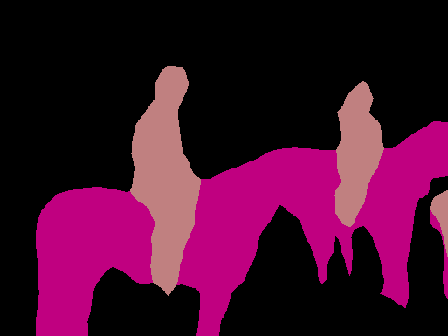}}
    \subfigure {\includegraphics[width=0.18\linewidth]{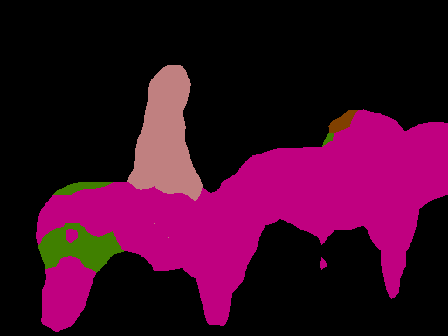}}
    \subfigure {\includegraphics[width=0.18\linewidth]{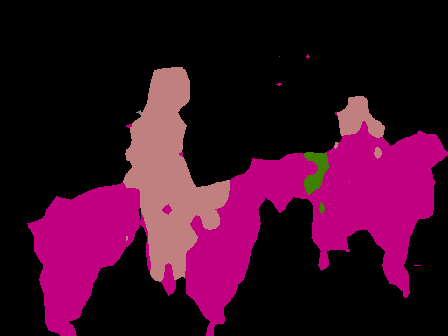}}
    
    \subfigure {\includegraphics[width=0.18\linewidth]{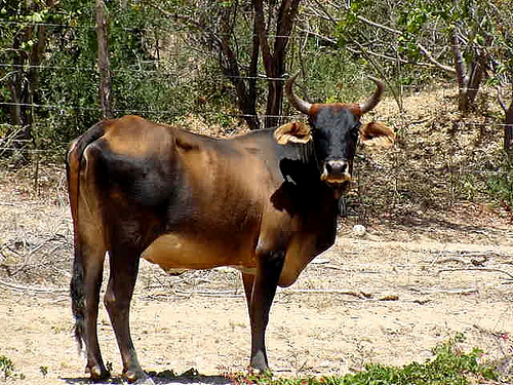}}
    \subfigure {\includegraphics[width=0.18\linewidth]{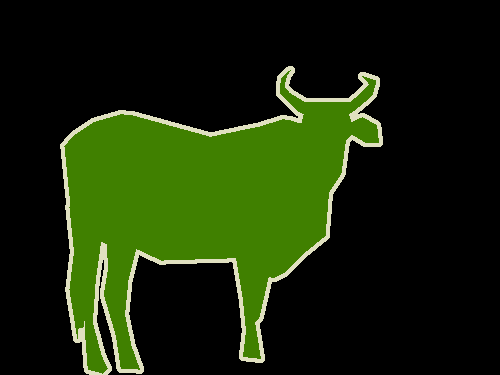}}
    \subfigure {\includegraphics[width=0.18\linewidth]{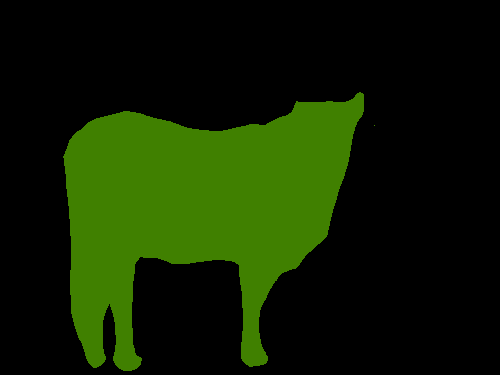}}
    \subfigure {\includegraphics[width=0.18\linewidth]{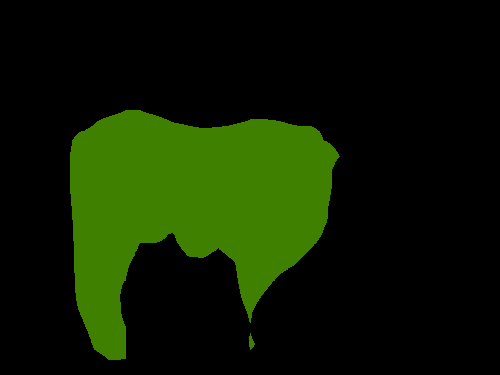}}
    \subfigure {\includegraphics[width=0.18\linewidth]{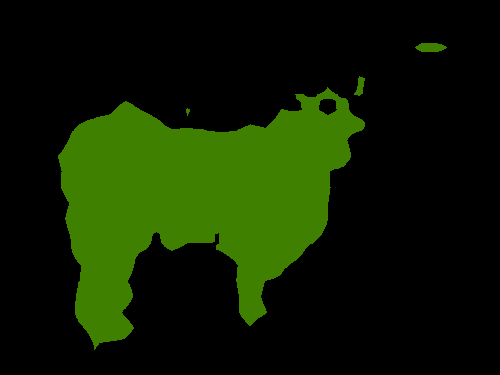}}
    
    \subfigure {\includegraphics[width=0.18\linewidth]{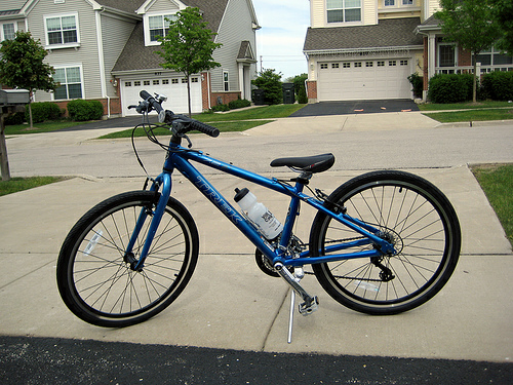}}
    \subfigure {\includegraphics[width=0.18\linewidth]{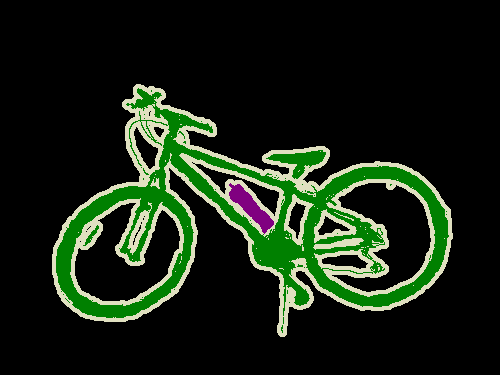}}
    \subfigure {\includegraphics[width=0.18\linewidth]{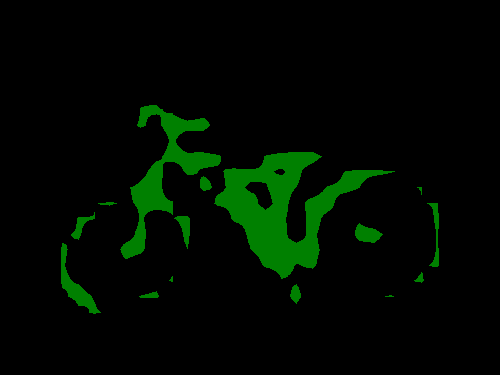}}
    \subfigure {\includegraphics[width=0.18\linewidth]{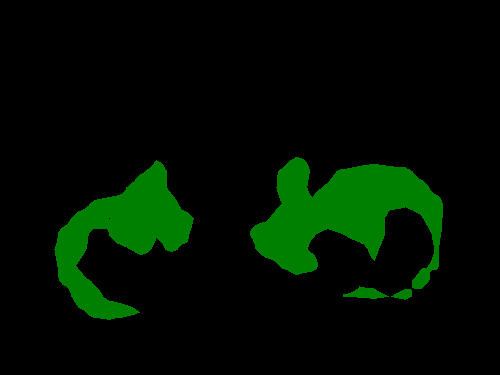}}
    \subfigure {\includegraphics[width=0.18\linewidth]{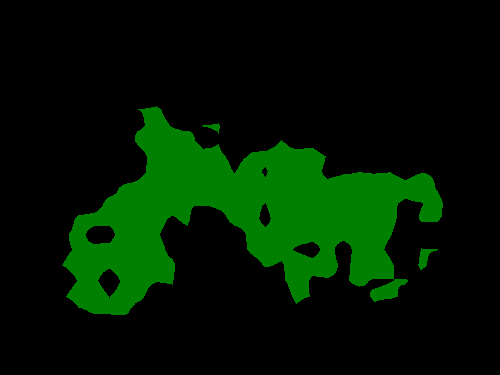}}
    
    \subfigure {\includegraphics[width=0.18\linewidth]{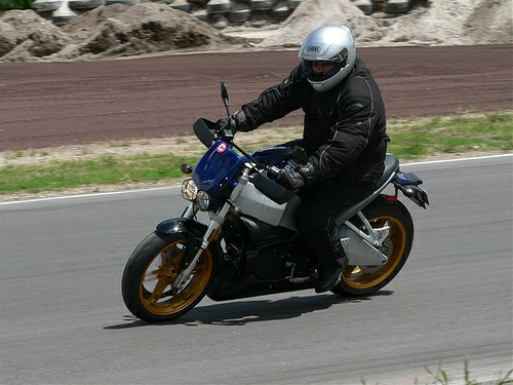}}
    \subfigure {\includegraphics[width=0.18\linewidth]{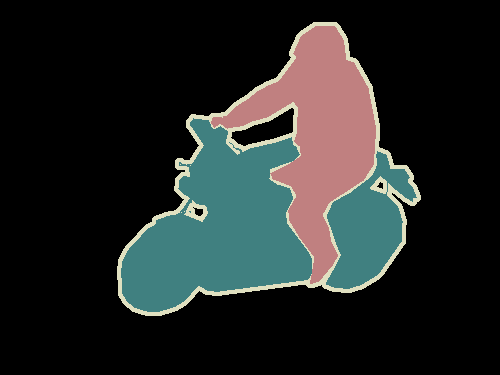}}
    \subfigure {\includegraphics[width=0.18\linewidth]{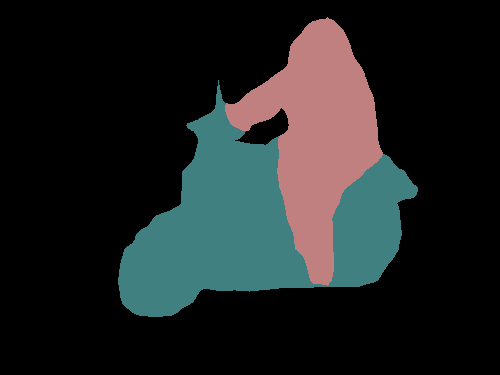}}
    \subfigure {\includegraphics[width=0.18\linewidth]{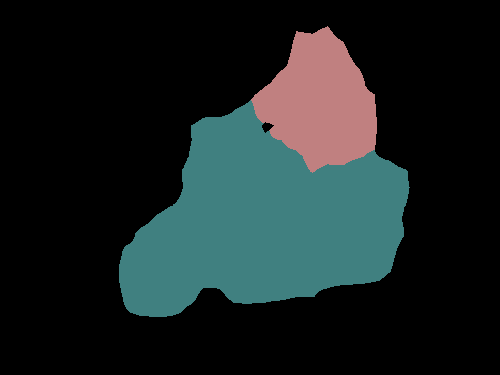}}
    \subfigure {\includegraphics[width=0.18\linewidth]{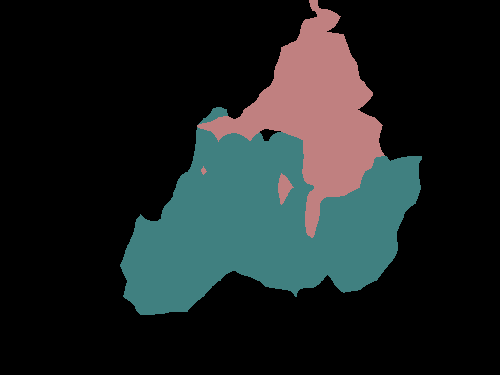}}
    
    
    \vskip 0.1in
    \caption{Visualization examples on the PASCAL VOC segmentation validation set. The first and second columns represent the input images and the ground truth labels. Third to fifth columns show the visualization results of the floating-point, 2-bit weight quantized, and 2-bit activation quantization models, respectively.}
    \label{app_fig:segmentation}
\end{figure}


\pagestyle{headings}
\mainmatter
\def\ECCVSubNumber{6068}  

\title{Supplementary Materials for Submission \ECCVSubNumber} 

\titlerunning{ECCV-20 Supplementary materials for submission ID \ECCVSubNumber} 
\author{Anonymous ECCV submission}
\institute{Paper ID \ECCVSubNumber}

\maketitle

\section*{A. Additional Visualization Examples on the Synthetized Dataset}

\begin{figure}[h]
    \centering
    \begin{subfigure}{.19\linewidth}
        \includegraphics[width=1\linewidth]{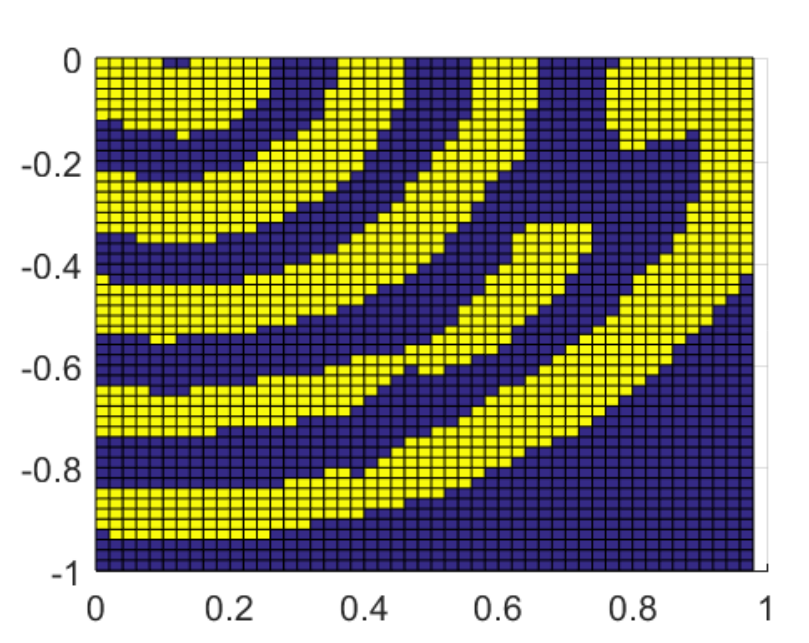}
        \caption{200-4 (W2)}
    \end{subfigure}
    \begin{subfigure}{.19\linewidth}
        \includegraphics[width=1\linewidth]{images_appendix/pred_300_4_w_2.pdf}
        \caption{384-4 (W2)}
    \end{subfigure}
    \begin{subfigure}{.19\linewidth}
        \includegraphics[width=1\linewidth]{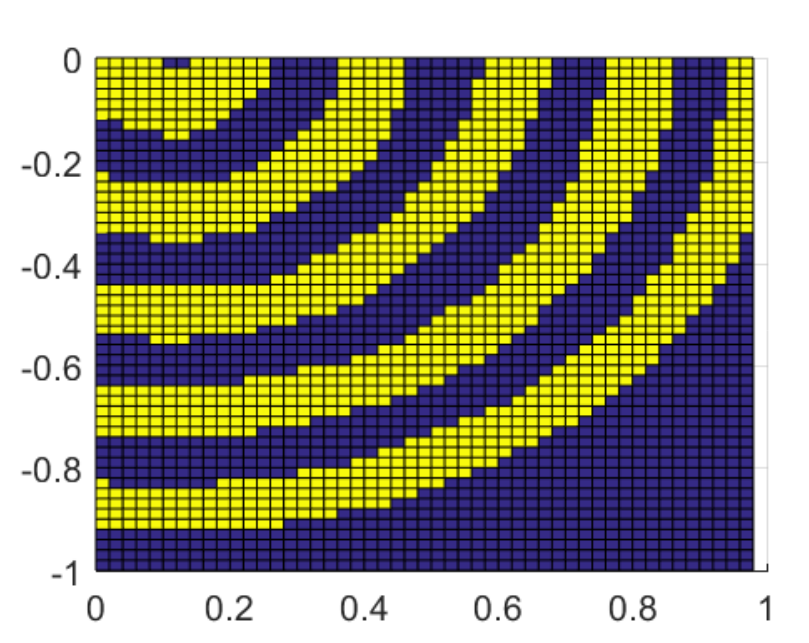}
        \caption{128-5 (W2)}
    \end{subfigure}
    \begin{subfigure}{.182\linewidth}
        \includegraphics[width=1\linewidth]{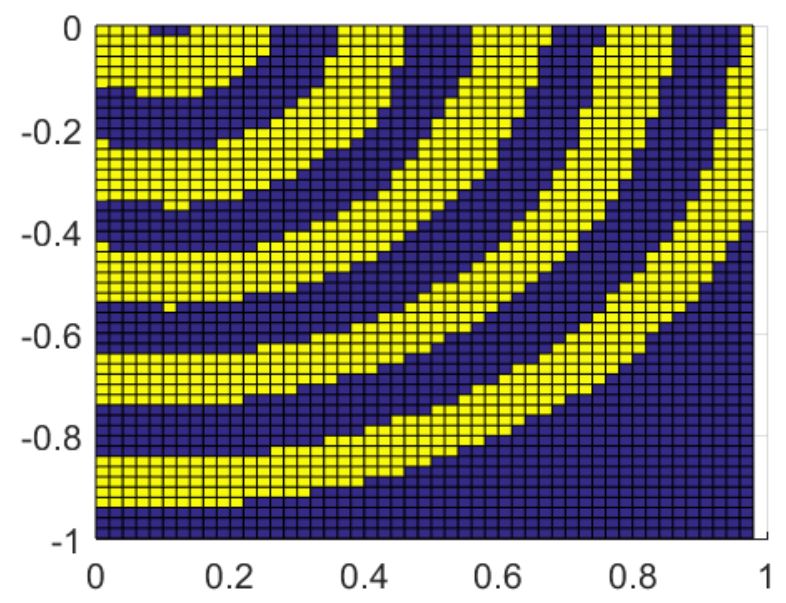}
        \caption{128-6 (W2)}
    \end{subfigure}
    \begin{subfigure}{.18\linewidth}
        \includegraphics[width=1\linewidth]{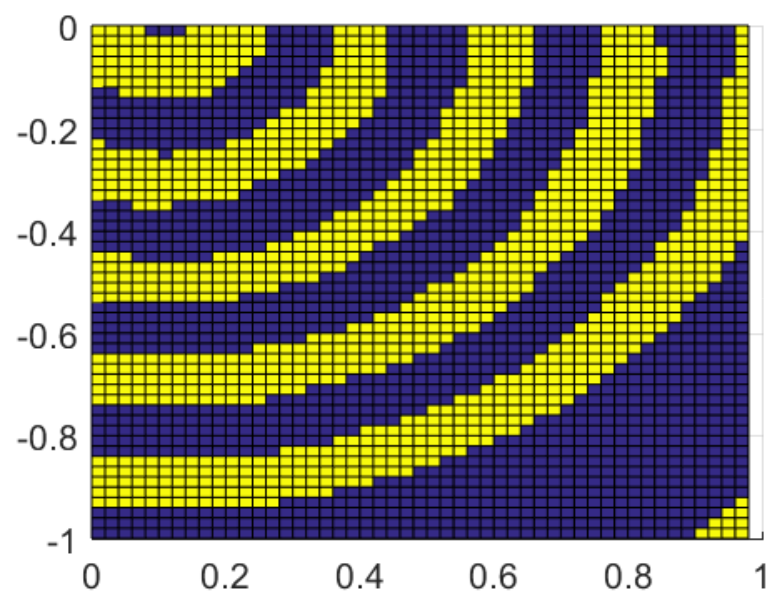}
        \caption{128-8 (W2)}
    \end{subfigure}
    \begin{subfigure}{.19\linewidth}
        \includegraphics[width=1\linewidth]{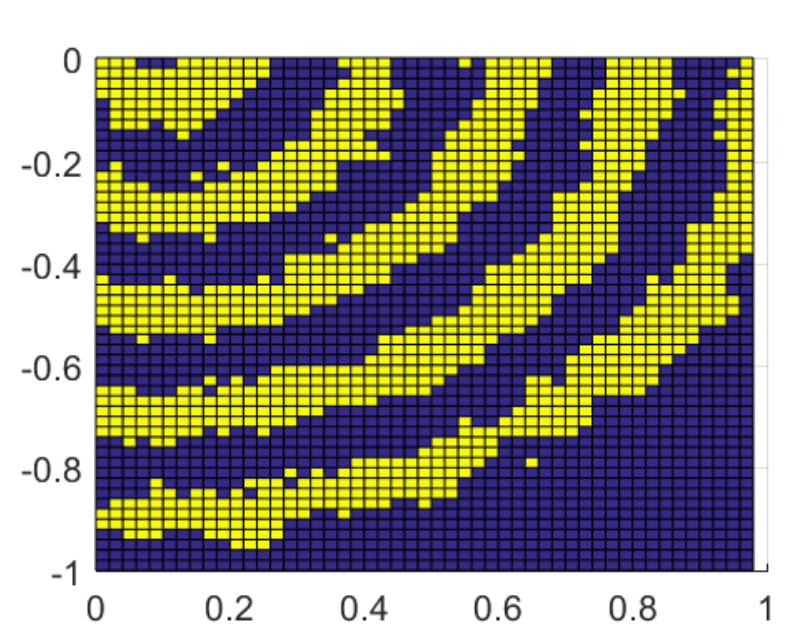}
        \caption{200-4 (A2)}
    \end{subfigure}
    \begin{subfigure}{.19\linewidth}
        \includegraphics[width=1\linewidth]{images2/384_4_w_32_a_2bits.pdf}
        \caption{384-4 (A2)}
    \end{subfigure}
    \begin{subfigure}{.19\linewidth}
        \includegraphics[width=1\linewidth]{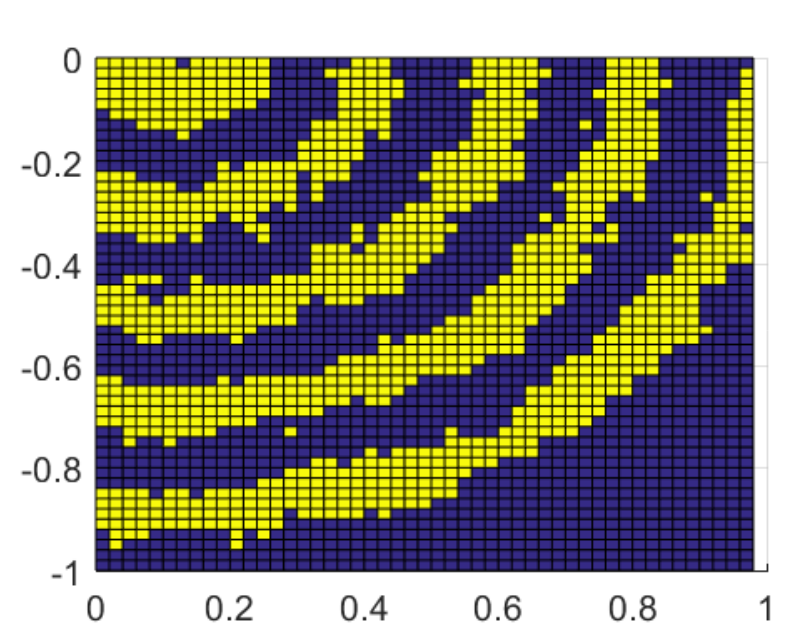}
        \caption{128-5 (A2)}
    \end{subfigure}
    \begin{subfigure}{.185\linewidth}
        \includegraphics[width=1\linewidth]{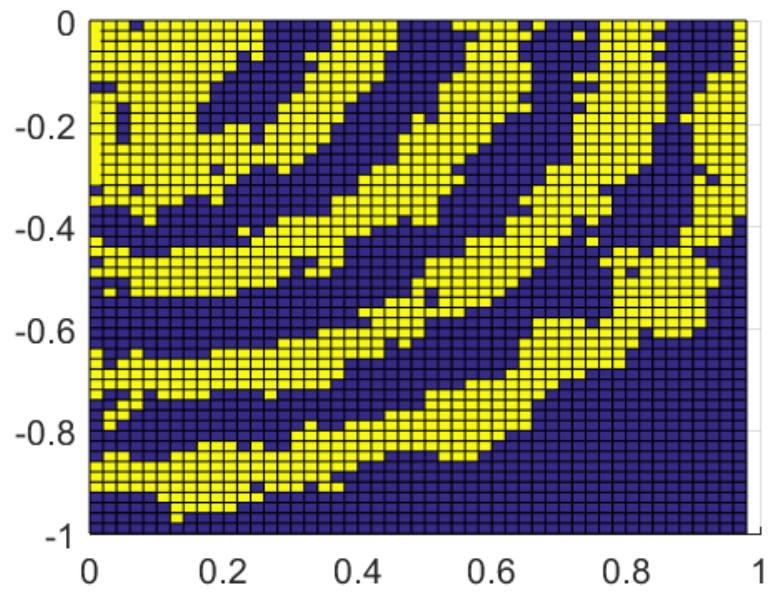}
        \caption{128-6 (A2)}
    \end{subfigure}
    \begin{subfigure}{.185\linewidth}
        \includegraphics[width=1\linewidth]{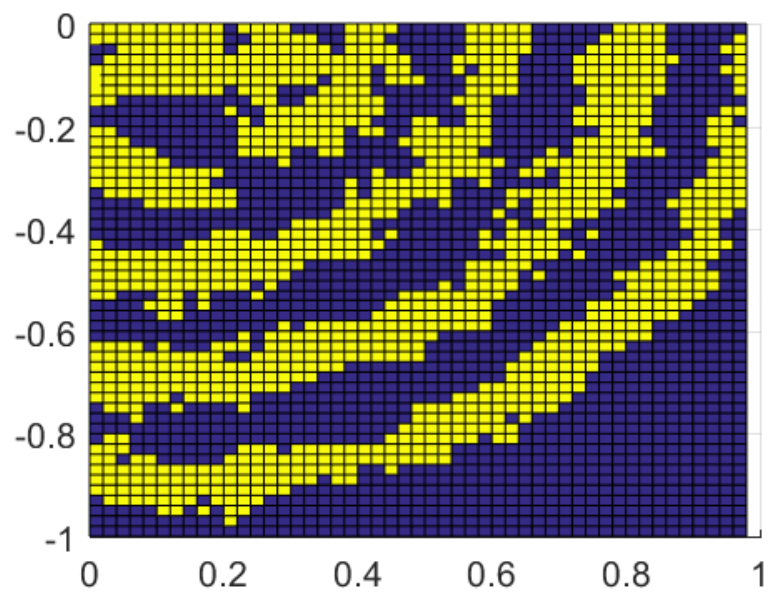}
        \caption{128-8 (A2)}
    \end{subfigure}
    \begin{subfigure}{.185\linewidth}
        \includegraphics[width=1\linewidth]{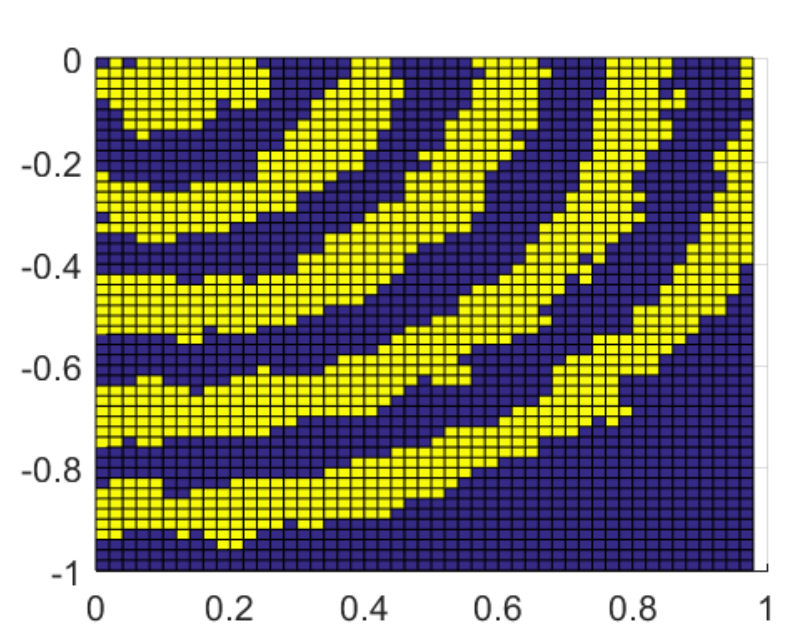}
        \caption{200-4 (W2A2)}
    \end{subfigure}
    \begin{subfigure}{.185\linewidth}
        \includegraphics[width=1\linewidth]{images2/384_4_w_2_a_2bits.pdf}
        \caption{384-4 (W2A2)}
    \end{subfigure}
    \begin{subfigure}{.185\linewidth}
        \includegraphics[width=1\linewidth]{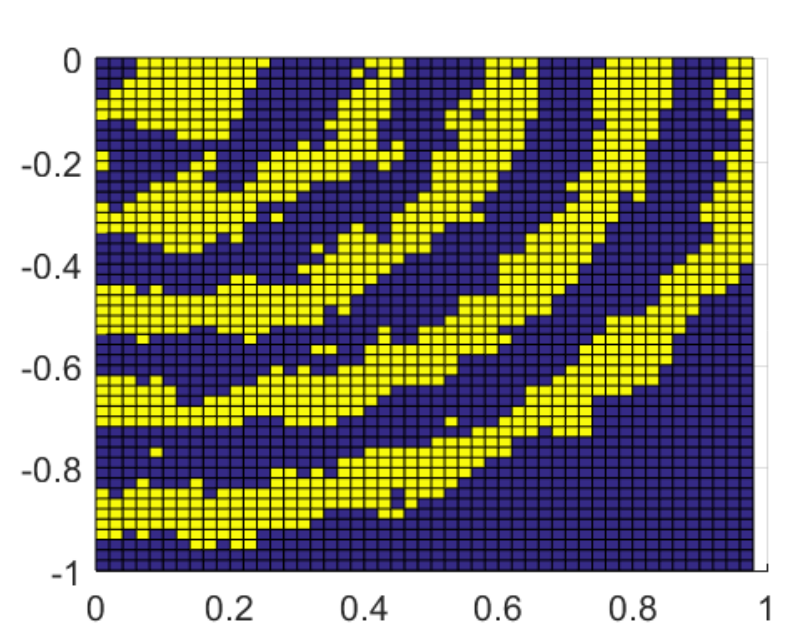}
        \caption{128-5 (W2A2)}
    \end{subfigure}
    \begin{subfigure}{.185\linewidth}
        \includegraphics[width=1\linewidth]{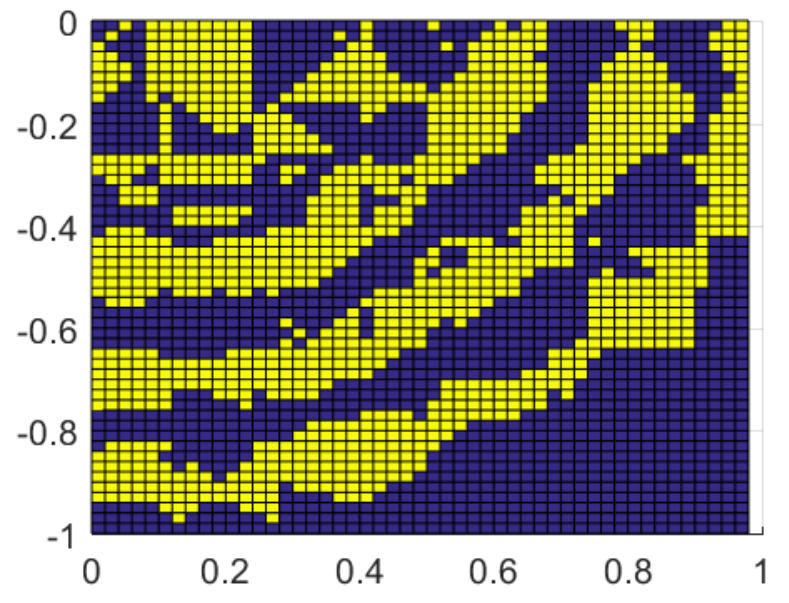}
        \caption{128-6 (W2A2)}
    \end{subfigure}
    \begin{subfigure}{.185\linewidth}
        \includegraphics[width=1\linewidth]{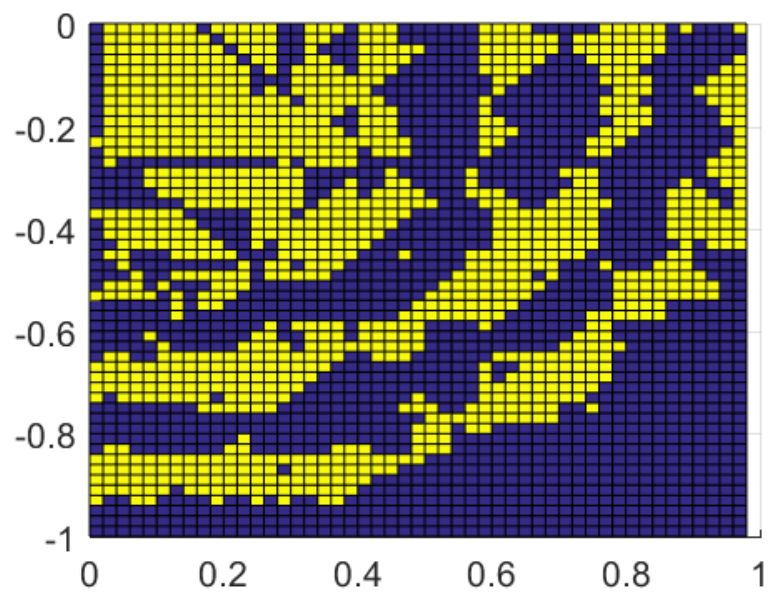}
        \caption{128-8 (W2A2)}
    \end{subfigure}
    \caption{Synthetic data visualization results on FCDNNs as the width and depth increase. The experimented FCDNNs are denoted as `$D$'-`$H$', where $D$ is the dimension of hidden layers and $H$ is the number of layers. W2 and A2 represent the 2-bit weight and 2-bit activation quantization, respectively.}
    \label{app_fig:toy_example_depth_width}
\end{figure}

Additional examples of visualization with synthetic dataset are shown in~\figurename~\ref{app_fig:toy_example_depth_width}. 200-4 FCDNN and 128-6 FCDNN contain 81K and 66K parameters, respectively. The deeper networks, 128-6 and 128-8 FCDNNs, are more robust to weight quantization. However, the wider one, 200-4 FCDNN is more resilient to activation quantization than the deeper networks. As discussed in Section 3 of the main contents, increasing the width alleviates the effects of weight and activation quantization errors, whereas increasing the depth only reduces the effects of weight quantization error. 


The visualization results with the residual connections are shown in~\figurename~\ref{app_fig:toy_example_residual}. As discussed in the main contents, increasing the depth with residual connections rather amplifies the activation quantization error when the shortcut outputs are quantized.


\begin{figure}[h]
    \centering
    \begin{subfigure}{.2\linewidth}
        \includegraphics[width=1\linewidth]{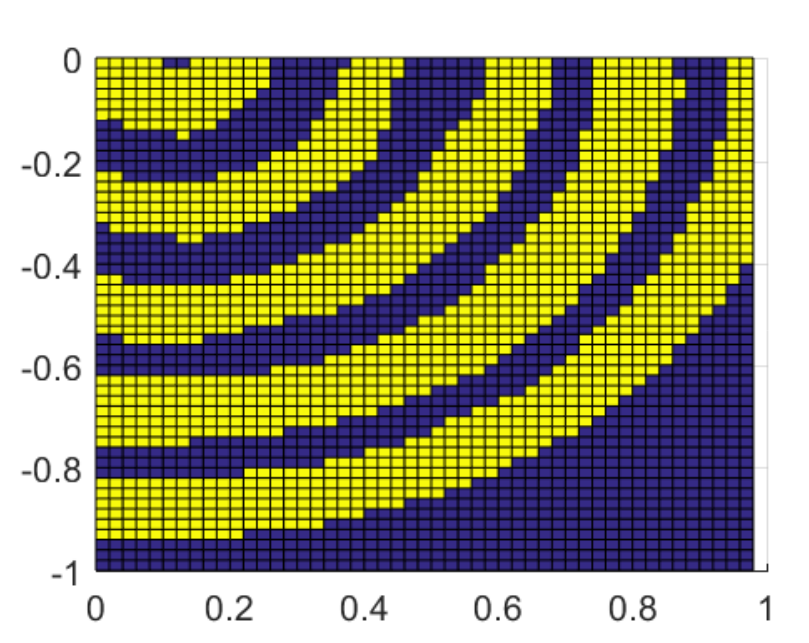}
        \caption{128-5 (W2)}
    \end{subfigure}
    \begin{subfigure}{.2\linewidth}
        \includegraphics[width=1\linewidth]{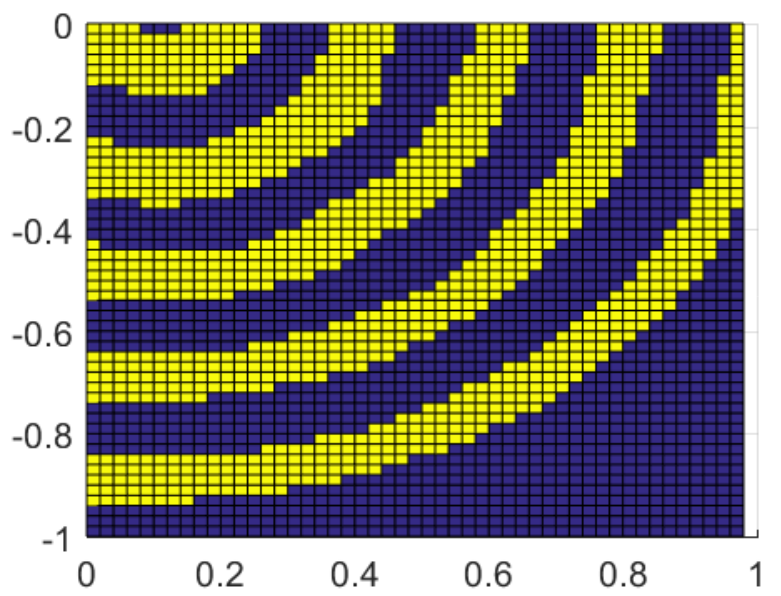}
        \caption{128-6 (W2)}
    \end{subfigure}
    \begin{subfigure}{.2\linewidth}
        \includegraphics[width=1\linewidth]{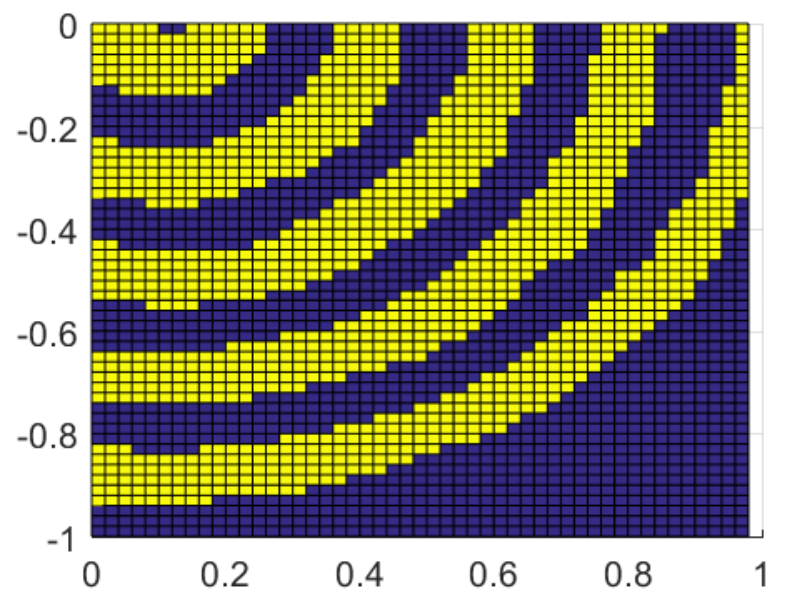}
        \caption{128-8 (W2)}
    \end{subfigure}
    \\
    \begin{subfigure}{.2\linewidth}
        \includegraphics[width=1\linewidth]{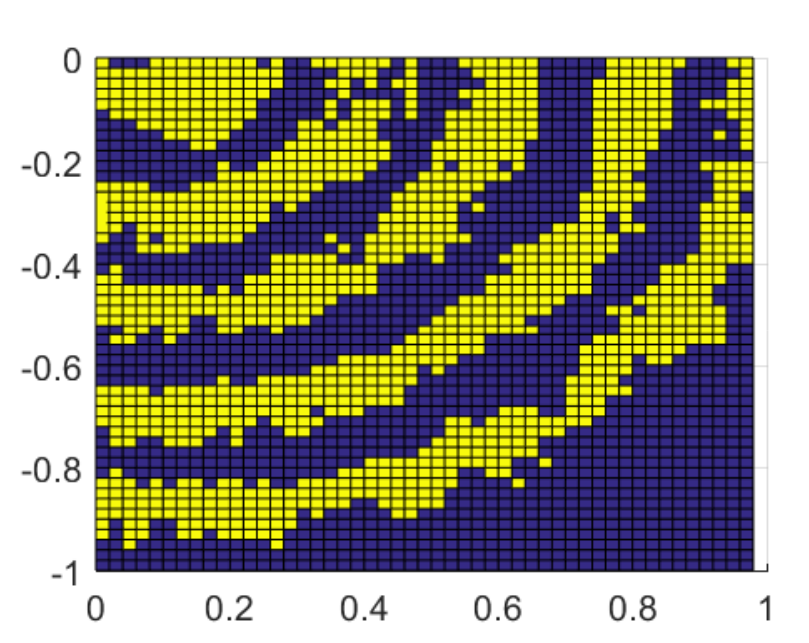}
        \caption{128-5 (A2)}
    \end{subfigure}
    \begin{subfigure}{.2\linewidth}
        \includegraphics[width=1\linewidth]{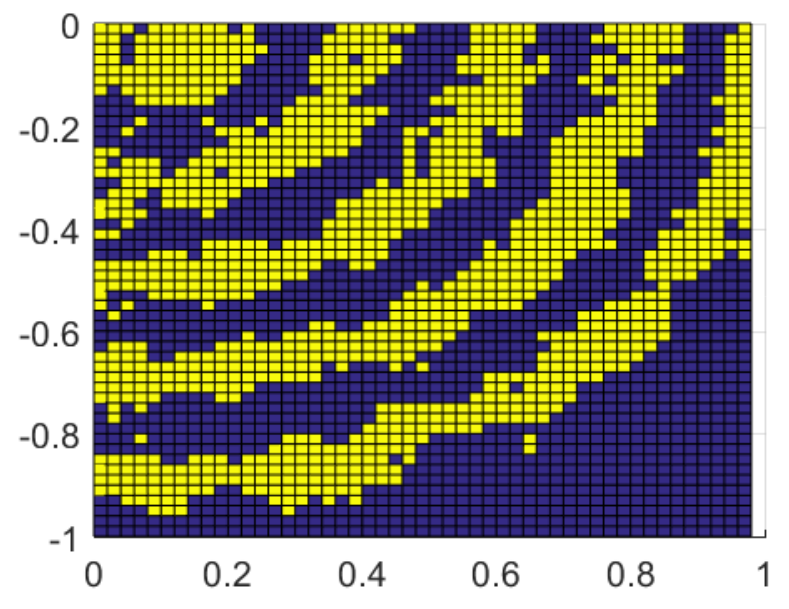}
        \caption{128-6 (A2)}
    \end{subfigure}
    \begin{subfigure}{.2\linewidth}
        \includegraphics[width=1\linewidth]{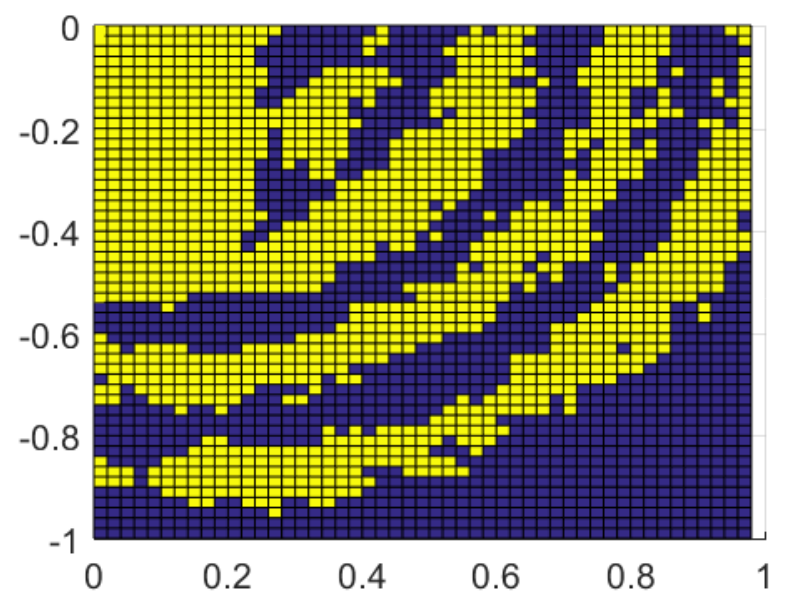}
        \caption{128-8 (A2)}
    \end{subfigure}
    \\
    \begin{subfigure}{.2\linewidth}
        \includegraphics[width=1\linewidth]{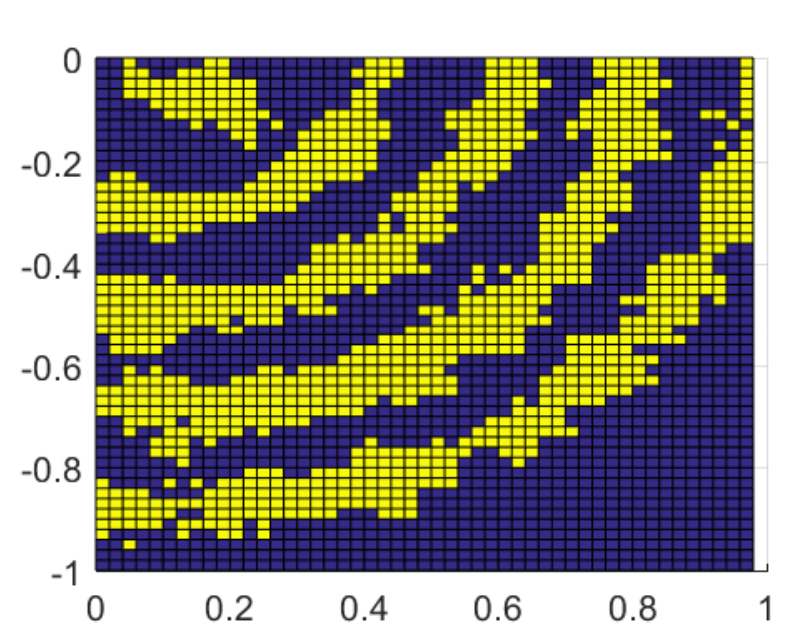}
        \caption{128-5 (W2A2)}
    \end{subfigure}
    \begin{subfigure}{.2\linewidth}
        \includegraphics[width=1\linewidth]{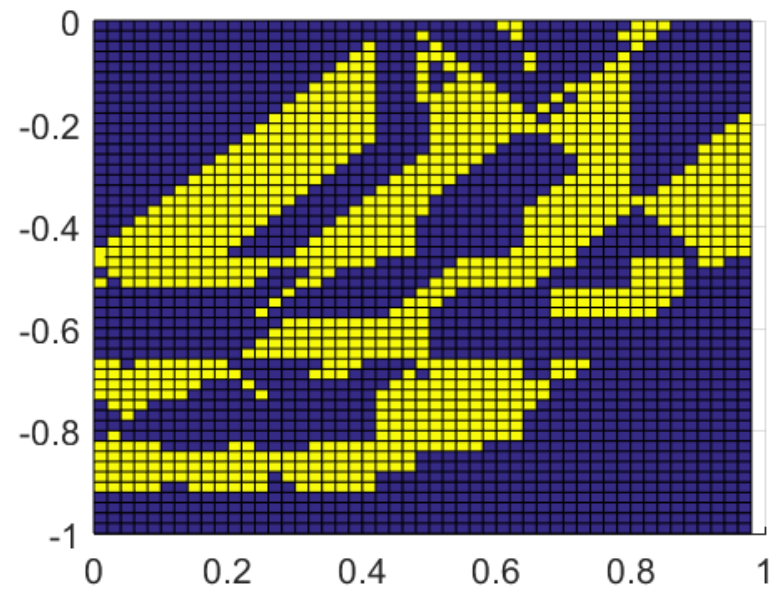}
        \caption{128-6 (W2A2)}
    \end{subfigure}
    \begin{subfigure}{.2\linewidth}
        \includegraphics[width=1\linewidth]{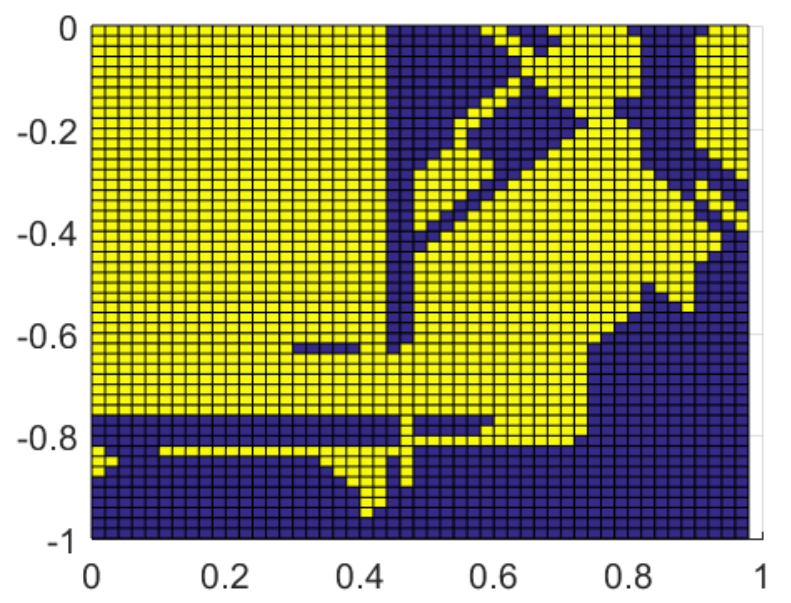}
        \caption{128-8 (W2A2)}
    \end{subfigure}
    \caption{Synthetic data visualization results on FCDNNs with quantized residual connections.}
    \label{app_fig:toy_example_residual}
\end{figure}

\section*{B. Details on Cyclic Learning Rate Scheduling}

\begin{figure}[h]
\centering
\tikzset{mark options={mark size=1}}
\begin{tikzpicture}
    \begin{axis}[
	width=0.8\linewidth,
	height = 0.25\linewidth,
	compat=1.12,
	xmin=0,
	xmax=600,
	ymode=log,
	xticklabels=\empty,
    xlabel=Iterations,
    ylabel=LR scale factor,
	xlabel shift=-3pt,
	ylabel shift=-3pt,
    tick label style={font=\footnotesize},
    label style={font=\footnotesize}
    ]]
	\addplot[color=black] file{data/clr.txt}; 
	\draw[color=red,dashed] (160, 0.008) rectangle (319, 1.2);
    \end{axis}
   \end{tikzpicture}
   \caption{Learning rate scale factor along training iterations. The red dashed box indicates a single cycle of CLR scheduling.}
   \label{app_fig:clr}
   \end{figure}

The cyclic learning rate (CLR) scheduling is shown in~\figurename~\ref{app_fig:clr}. The scale factor is multiplied to the base LR and changes 8 times in a single cycle; $\{1.0$, $\sqrt{0.1}$, $0.1$, $0.1\sqrt{0.1}$, $0.01$, 0$.1\sqrt{0.1}$, $0.1$, $\sqrt{0.1}$ $\}$. The cycle period and base LR for each dataset are as follows:

\noindent \textbf{CIFAR-10:} The base LR is 1e-4, which is 10 times larger than the last LR of the retraining. The cycle period is 8 epochs, thus the LR scaling factor changes at every epoch. 


\noindent \textbf{PASCAL VOC segmentation:} The last LR of the retraining becomes 0 with the polynomial policy for this task. We set the base LR to 1e-5, which is 10 times smaller than the initial LR of the retraining. The cycle period is 3K iterations.

\section*{C. Additional Visualization Examples on the PASCAL VOC Segmentation Benchmark}

\begin{figure}[h]
    \centering
    \begin{subfigure}{.18\linewidth}
        \caption{Image}
        \includegraphics[width=1\linewidth]{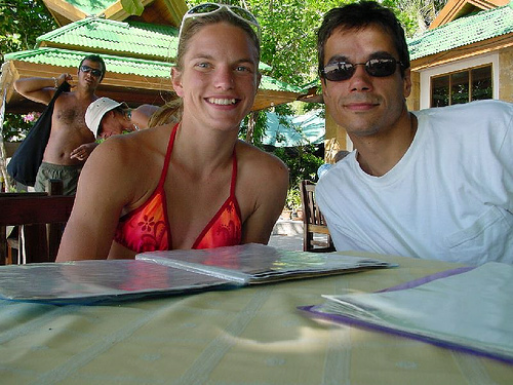}
    \end{subfigure}
    \begin{subfigure}{.18\linewidth}
        \caption{Ground truth}
        \includegraphics[width=1\linewidth]{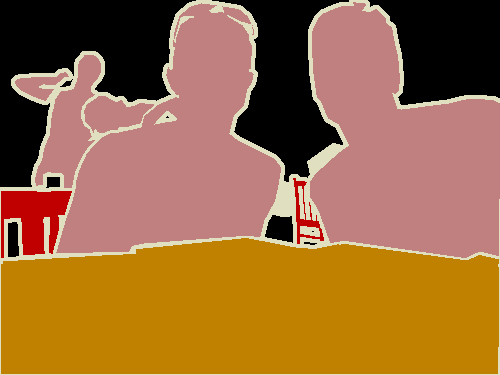}
    \end{subfigure}
    \begin{subfigure}{.18\linewidth}
        \caption{Float}
        \includegraphics[width=1\linewidth]{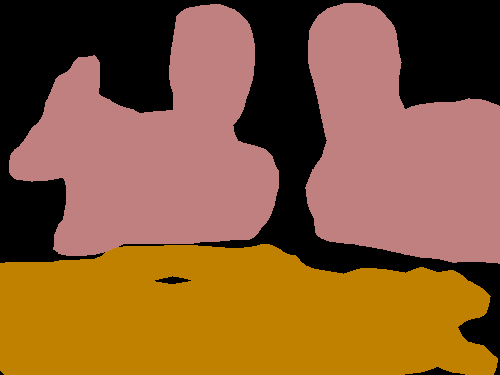}
    \end{subfigure}
    \begin{subfigure}{.18\linewidth}
        \caption{2-bit W}
        \includegraphics[width=1\linewidth]{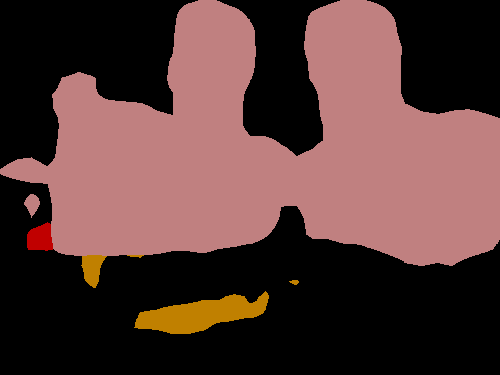}
    \end{subfigure}
    \begin{subfigure}{.18\linewidth}
        \caption{2-bit A}
        \includegraphics[width=1\linewidth]{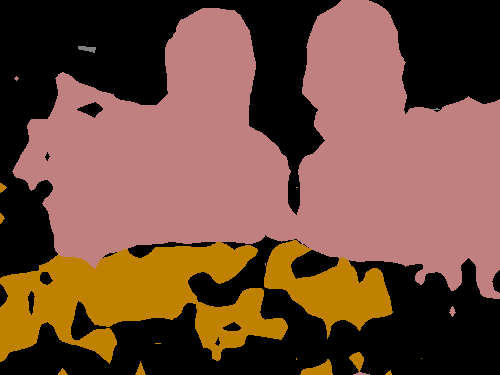}
    \end{subfigure}
    \vskip 0.03in
    \begin{subfigure}{.18\linewidth}
        \includegraphics[width=1\linewidth]{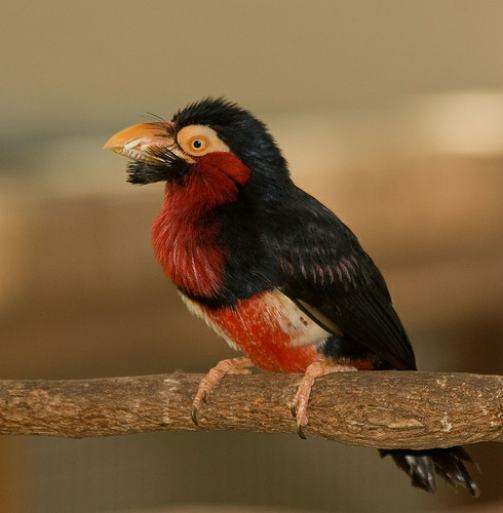}
    \end{subfigure}
    \begin{subfigure}{.18\linewidth}
        \includegraphics[width=1\linewidth]{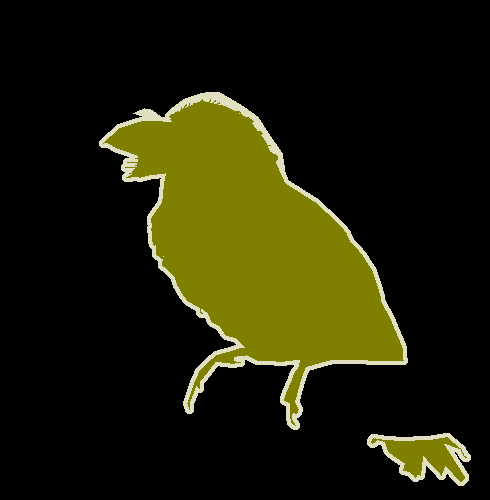}
    \end{subfigure}
    \begin{subfigure}{.18\linewidth}
        \includegraphics[width=1\linewidth]{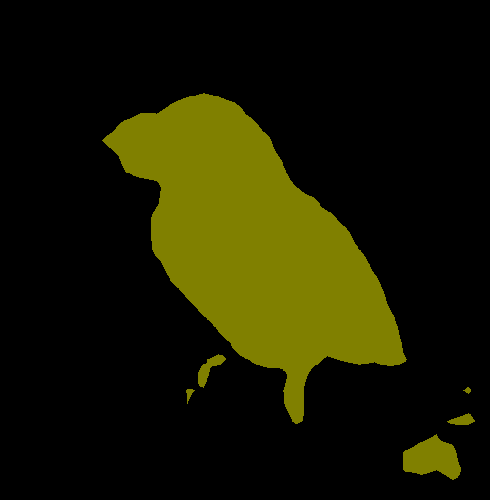}
    \end{subfigure}
    \begin{subfigure}{.18\linewidth}
        \includegraphics[width=1\linewidth]{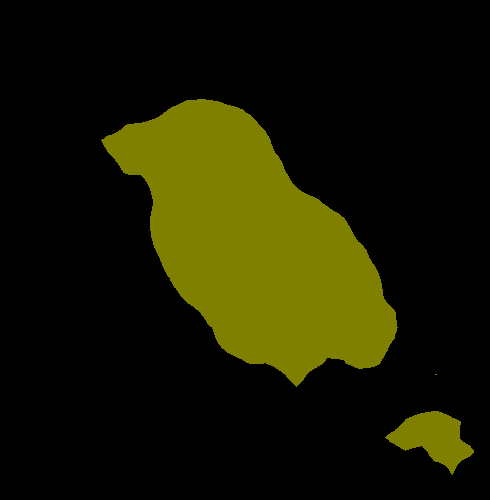}
    \end{subfigure}
    \begin{subfigure}{.18\linewidth}
        \includegraphics[width=1\linewidth]{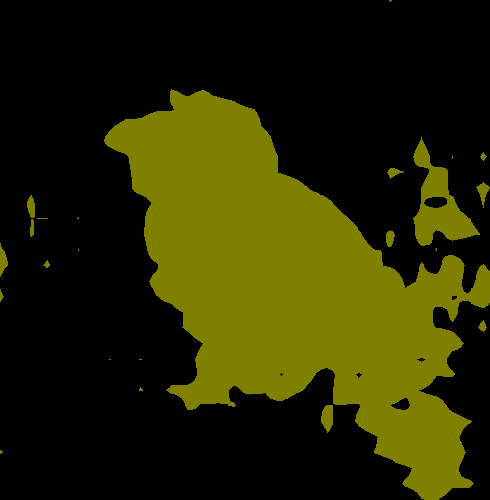}
    \end{subfigure}
   \vskip 0.03in
    \begin{subfigure}{.18\linewidth}
        \includegraphics[width=1\linewidth]{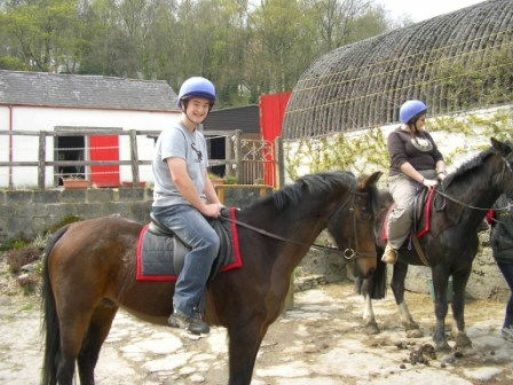}
    \end{subfigure}
    \begin{subfigure}{.18\linewidth}
        \includegraphics[width=1\linewidth]{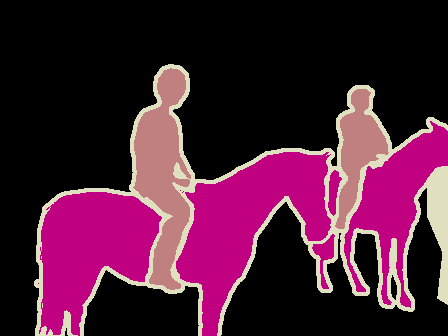}
    \end{subfigure}
    \begin{subfigure}{.18\linewidth}
        \includegraphics[width=1\linewidth]{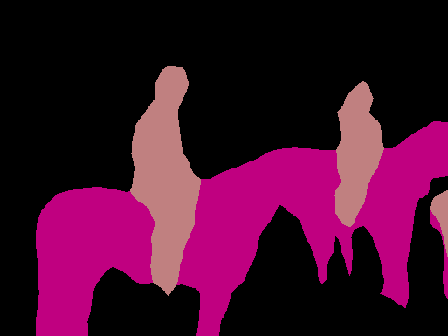}
    \end{subfigure}
    \begin{subfigure}{.18\linewidth}
        \includegraphics[width=1\linewidth]{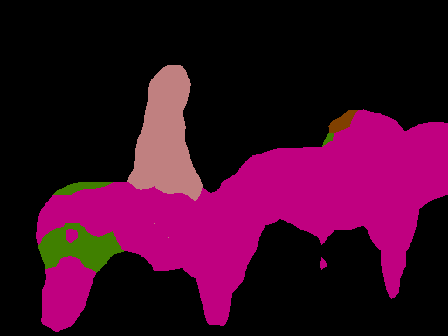}
    \end{subfigure}
    \begin{subfigure}{.18\linewidth}
        \includegraphics[width=1\linewidth]{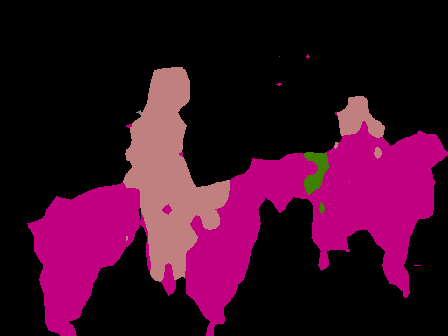}
    \end{subfigure}
    \vskip 0.03in
    \begin{subfigure}{.18\linewidth}
        \includegraphics[width=1\linewidth]{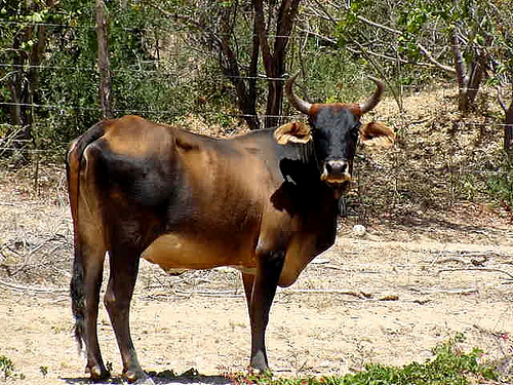}
    \end{subfigure}
    \begin{subfigure}{.18\linewidth}
        \includegraphics[width=1\linewidth]{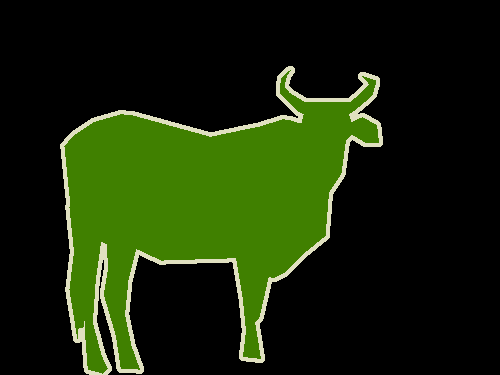}
    \end{subfigure}
    \begin{subfigure}{.18\linewidth}
        \includegraphics[width=1\linewidth]{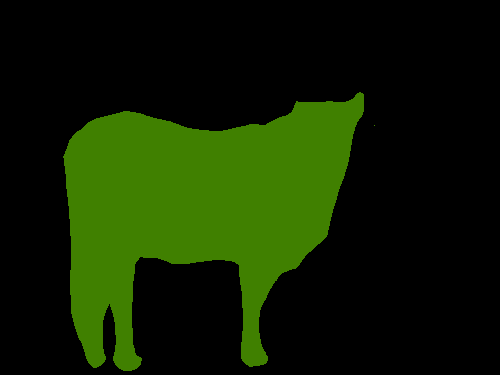}
    \end{subfigure}
    \begin{subfigure}{.18\linewidth}
        \includegraphics[width=1\linewidth]{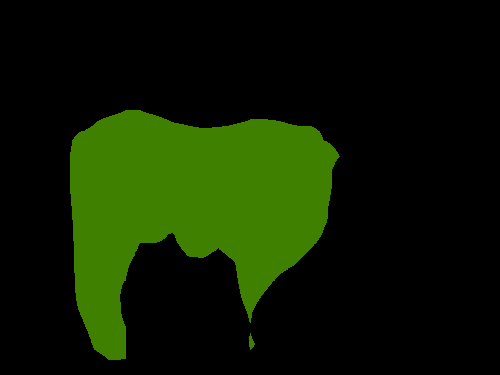}
    \end{subfigure}
    \begin{subfigure}{.18\linewidth}
        \includegraphics[width=1\linewidth]{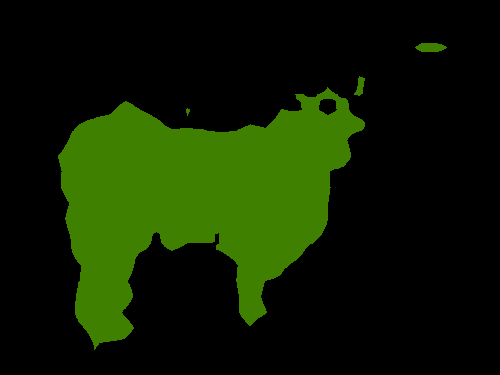}
    \end{subfigure}
    \vskip 0.03in
    \begin{subfigure}{.18\linewidth}
        \includegraphics[width=1\linewidth]{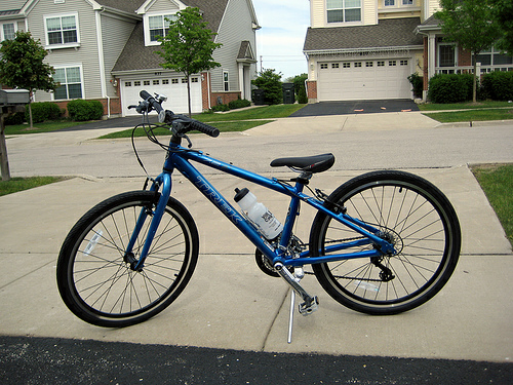}
    \end{subfigure}
    \begin{subfigure}{.18\linewidth}
        \includegraphics[width=1\linewidth]{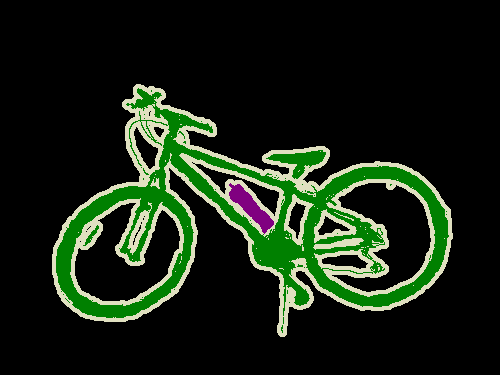}
    \end{subfigure}
    \begin{subfigure}{.18\linewidth}
        \includegraphics[width=1\linewidth]{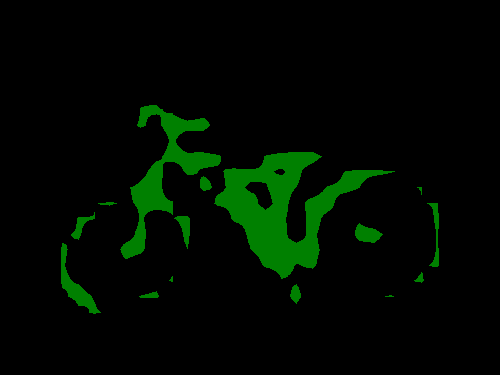}
    \end{subfigure}
    \begin{subfigure}{.18\linewidth}
        \includegraphics[width=1\linewidth]{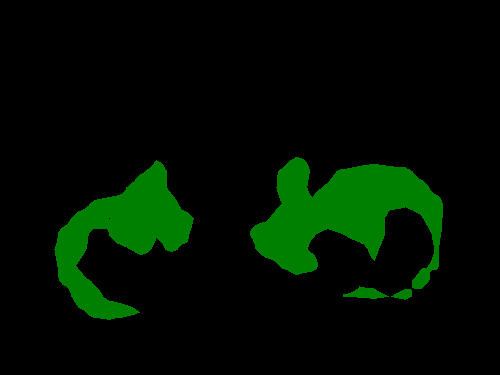}
    \end{subfigure}
    \begin{subfigure}{.18\linewidth}
        \includegraphics[width=1\linewidth]{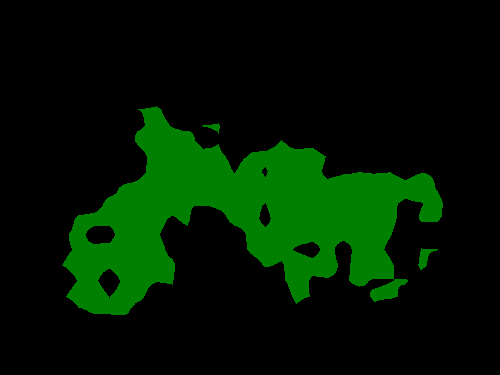}
    \end{subfigure}
    \vskip 0.03in
    \begin{subfigure}{.18\linewidth}
        \includegraphics[width=1\linewidth]{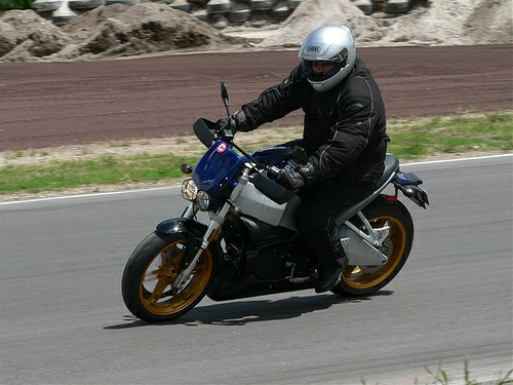}
    \end{subfigure}
    \begin{subfigure}{.18\linewidth}
        \includegraphics[width=1\linewidth]{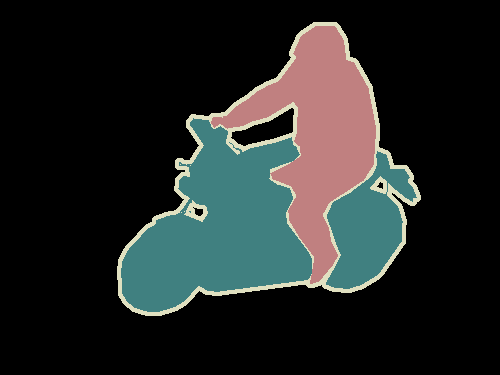}
    \end{subfigure}
    \begin{subfigure}{.18\linewidth}
        \includegraphics[width=1\linewidth]{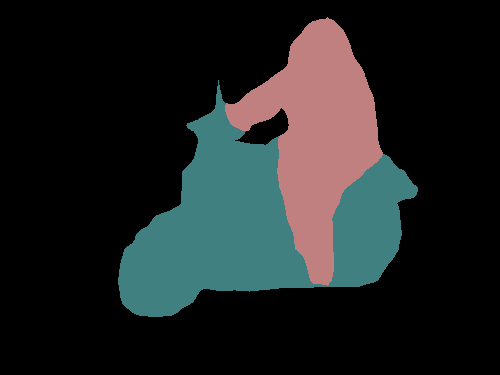}
    \end{subfigure}
    \begin{subfigure}{.18\linewidth}
        \includegraphics[width=1\linewidth]{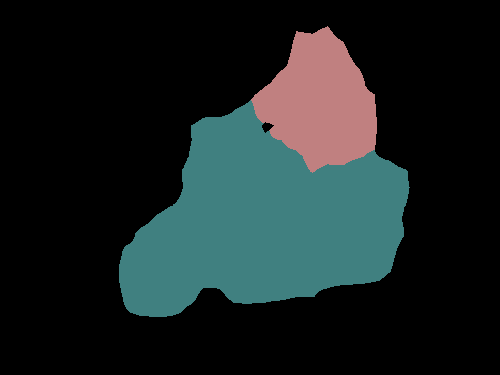}
    \end{subfigure}
    \begin{subfigure}{.18\linewidth}
        \includegraphics[width=1\linewidth]{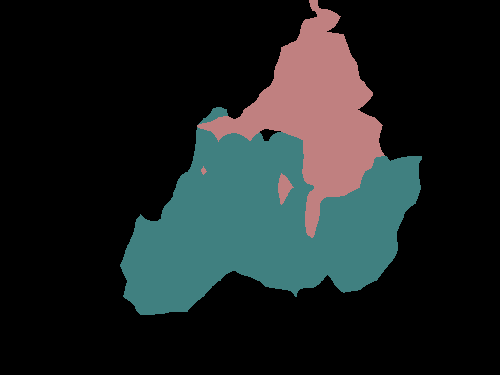}
    \end{subfigure}
    \caption{Visualization examples on the PASCAL VOC segmentation validation set. The first and second columns represent the input images and the ground truth labels. Third to fifth columns show the visualization results of the floating-point, 2-bit weight quantized, and 2-bit activation quantized models, respectively.}
    \label{app_fig:segmentation}
\end{figure}
\clearpage

\section*{D. Details of Experimented Model Structures}

\begin{table}[h]
\centering
\caption{Quantization performance of various model structures on CIFAR-10 testset. `$L$', `$D_{init}$', and `$B$' represent the number of layers, initial channel dimension, and number of blocks, respectively.}
\setlength\tabcolsep{5pt}
\begin{tabular}{ccccc|cccc}
\toprule
\multicolumn{5}{c|}{Model}                                                                                                       & \multicolumn{4}{c}{Test Acc(\%)} \\
Block type                                                                      & $L$ & $D_{init}$ &  $B$ &  Params. & Float   & W2AF   & WFA2  & W2A2  \\ \midrule
\multirow{6}{*}{Basic block}                                                    & 8        & 59         & $1\times3$         & 1.05 M    & 93.26  & 92.83  & 92.58  & 92.32 \\
                                                                                & 14       & 40         & $2\times3$         & 1.10 M    & 93.62  & 93.51  & \textbf{93.30}  & \textbf{93.11} \\
                                                                                & 20       & 31         & $3\times3$         & 1.03 M    & 94.14  & 93.57  & 92.94  & 92.65 \\
                                                                                & 32       & 24         & $5\times3$         & 1.06 M    & 94.24  & 93.71  & 92.42  & 91.84 \\
                                                                                & 44       & 20         & $7\times3$         & 1.04 M    & \textbf{94.27}  & \textbf{93.91}  & 91.40  & 90.69 \\
                                                                                & 56       & 18         & $9\times3$         & 1.09 M    & 93.58  & 93.46  & 90.58  & 89.77 \\ \midrule
\multirow{6}{*}{\begin{tabular}[c]{@{}c@{}}Pre-activation\\ block\end{tabular}} & 8        & 59         & $1\times3$         & 1.05 M    & 93.10  & 92.75  & 91.89  & 91.03 \\
                                                                                & 14       & 40         & $2\times3$         & 1.10 M    & 94.14  & 93.85  & \textbf{92.44}  & \textbf{91.81} \\
                                                                                & 20       & 31         & $3\times3$         & 1.03 M    & 94.25  & 93.82  & 92.42  & \textbf{91.81} \\
                                                                                & 32       & 24         & $5\times3$         & 1.06 M    & 94.07  & 93.89  & 91.51  & 89.03 \\
                                                                                & 44       & 20         & $7\times3$         & 1.04 M    & \textbf{94.31}  & \textbf{94.04}  & 90.58  & 86.26 \\
                                                                                & 56       & 18         & $9\times3$         & 1.09 M    & 93.96  & 93.83  & 88.45  & 85.45 \\ \midrule
\multirow{5}{*}{\begin{tabular}[c]{@{}c@{}}Depthwise\\ block\end{tabular}}      & 29       & 38         & $3\times3$         & 1.11 M    & 93.64  & 93.28  & \textbf{93.38}  & 92.33 \\
                                                                                & 38       & 32         & $4\times3$         & 1.08 M    & 93.65  & 93.47  & 92.93  & \textbf{92.57} \\
                                                                                & 56       & 26         & $6\times3$         & 1.11 M    & 94.27  & 94.02  & 91.70  & 90.61 \\
                                                                                & 74       & 22         & $8\times3$         & 1.08 M    & 93.94  & 93.71  & 90.50  & 89.88 \\
                                                                                & 110      & 18         & $12\times3$         & 1.13 M    & \textbf{94.43}  & \textbf{94.39}  & 88.75  & 87.47 \\ \bottomrule
\end{tabular}
\label{app_table:structure}
\end{table}

The details of the model structure for Fig. 8 in the main contents are shown in Table~\ref{app_table:structure}. The number of parameters in all the experimented models is approximately one million. All CNNs consist of three block groups. `$B$' indicates the block structure; (number of blocks in each group) $\times$ (number of groups). The first block in second and third block groups have a stride of 2. ‘$D$’ represents the channel width of the first layer and the channel width increases by 2 times at the beginning of second and third block groups. Floating-point and weight quantized networks perform better on deeper models, while shallow and wider models shows better performance when activations are quantized.

\bibliographystyle{splncs04}
\bibliography{egbib}